\documentclass[twoside]{article}

\usepackage[accepted]{aistats2023}
\usepackage{multibib}
\newcites{online}{Supplementary Materials References}
\bibliographystyleonline{apalike}

\usepackage[round]{natbib}

\usepackage[dvipsnames]{xcolor}
\usepackage{threeparttable}
\usepackage{booktabs}
\usepackage{arydshln}
\usepackage{courier}  
\usepackage[hyphens]{url}  
\usepackage{graphicx} 
\usepackage{algorithm}
\usepackage{algorithmic}
\usepackage{multirow}
\usepackage{newtxmath}
\usepackage{newfloat}
\usepackage{listings}
\usepackage{caption}
\usepackage[subpreambles=true]{standalone}
\usepackage{import}

\makeatletter
\def\adl@drawiv#1#2#3{%
        \hskip.5\tabcolsep
        \xleaders#3{#2.5\@tempdimb #1{1}#2.5\@tempdimb}%
                #2\z@ plus1fil minus1fil\relax
        \hskip.5\tabcolsep}
\newcommand{\cdashlinelr}[1]{%
  \noalign{\vskip\aboverulesep
           \global\let\@dashdrawstore\adl@draw
           \global\let\adl@draw\adl@drawiv}
  \cdashline{#1}
  \noalign{\global\let\adl@draw\@dashdrawstore
           \vskip\belowrulesep}}
\makeatother

\begin{document}

%

%

\twocolumn[

\aistatstitle{TabLLM: Few-shot Classification of Tabular Data with Large Language Models}

\title{TabLLM: Few-shot Classification of Tabular Data with Large Language Models}

\aistatsauthor{
Stefan Hegselmann\textsuperscript{\rm 1,2} \hspace{0.12cm}
Alejandro Buendia\textsuperscript{\rm 1} \hspace{0.12cm}
Hunter Lang\textsuperscript{\rm 1} \hspace{0.12cm}
Monica Agrawal\textsuperscript{\rm 1} \hspace{0.12cm}
Xiaoyi Jiang\textsuperscript{\rm 2} \hspace{0.12cm}
David Sontag\textsuperscript{\rm 1}
}

\aistatsaddress{
\textsuperscript{\rm 1} MIT CSAIL \hspace{0.2cm}
\textsuperscript{\rm 2} University of Münster}
 ]

 \runningauthor{Stefan Hegselmann, Alejandro Buendia, Hunter Lang, Monica Agrawal, Xiaoyi Jiang, David Sontag}

\begin{abstract}
\looseness=-1 We study the application of large language models to zero-shot and few-shot classification of \emph{tabular data}.
We prompt the large language model with a serialization of the tabular data to a natural-language string, together with a short description of the classification problem.
In the few-shot setting, we fine-tune the large language model using some labeled examples.
We evaluate several serialization methods including templates, table-to-text models, and large language models.
Despite its simplicity, we find that this technique outperforms prior deep-learning-based tabular classification methods on several benchmark datasets.
In most cases, even \emph{zero}-shot classification obtains non-trivial performance, illustrating the method's ability to exploit prior knowledge encoded in large language models.
Unlike many deep learning methods for tabular datasets, this approach is also competitive with strong traditional baselines like gradient-boosted trees, especially in the very-few-shot setting.

\end{abstract}

\section{INTRODUCTION}

Many real world applications generate \emph{tabular data} as a natural byproduct of relational databases \citep{shwartz-ziv_tabular_2021}.
It is ubiquitous in domains ranging from healthcare to climate and finance  \citep{sahakyan2021explainable}.
Obtaining enough labeled data to train supervised learning algorithms for classification can be difficult.
For example, in healthcare, there are 10,000 rare diseases \citep{haendel2020many} affecting very few patients, which hampers the development of risk stratification models.  
Thus, we seek to develop methods that can exploit prior knowledge (e.g., from medical articles) to improve predictive performance in settings with a small number of training examples, i.e. the \emph{few-shot} setting.

While deep learning has led to breakthroughs in computer vision and natural language processing, this success has not yet been extended to the tabular domain.
For example, self-supervised deep learning methods have been introduced for tabular data \citep{yin_tabert_2020, arik_tabnet_2021}, but \citet{grinsztajn_why_2022} showed that these deep techniques still underperform ensembles of gradient boosted trees in the fully supervised setting.
This disparity in performance can be attributed to the differences between tabular data and text or images; tabular data lacks locality, contains mixed data types, and the number of columns is usually fairly small compared to the number of features in text or image data \citep{borisov_deep_2022}.

Recently, large language models (LLMs) such as \mbox{GPT-3}, which are pre-trained on enormous corpora of text, have shown incredible performance on few-shot text classification and generation tasks \citep{brown_language_2020, sanh_multitask_2021, ouyang_training_2022}. 
These LLMs perform well on a variety of tasks and domains, including fact retrieval \citep{liu_gpt_2021}, mathematical reasoning \citep{wei_chain_2022}, medical information extraction \citep{agrawal_large_2022}, and tabular data cleaning tasks  \citep{narayan_can_2022}.
Most importantly, because of all the knowledge encoded in their parameters, LLMs require little or no \emph{labeled} training data to obtain this good performance.

\begin{figure*}[ht!]
  \centering
  \includegraphics[width=\textwidth]{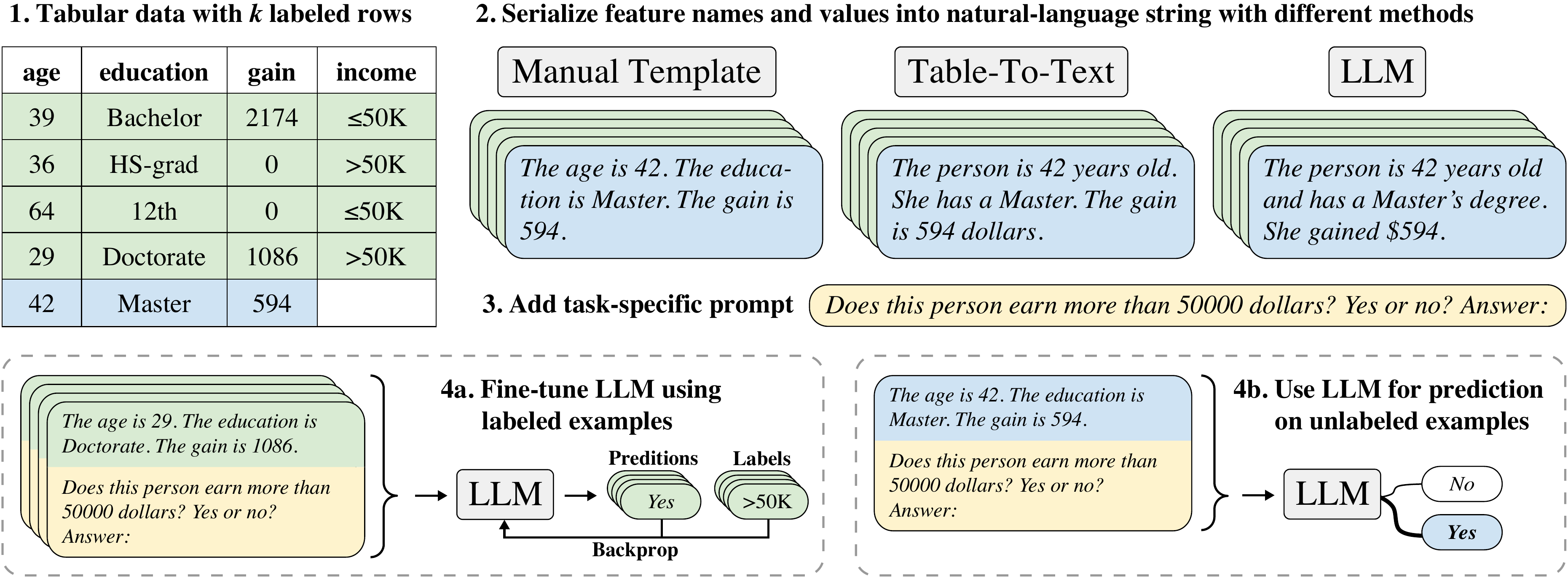}
  \vspace{-0.5cm}
  \caption{Overview of TabLLM. We first serialize the feature names and values into a natural language string.
  We evaluate different strategies.
  This string is then combined with a task-specific prompt. 
  To get predictions, we obtain output probabilities from the LLM for each of a pre-specified set of verbalizer tokens (e.g., ``Yes'', ``No''), which map to class labels (e.g., $1$, $-1$).
  If $k>0$, we use the $k$ labeled examples to fine-tune the large language model using T-Few \citep{liu_few-shot_2022}.
  Finally, we use the (possibly tuned) large language model to obtain predictions on unlabeled examples.
 }
  \vspace{-0.3cm}
  \label{fig:overview}
\end{figure*}

In this work we introduce \textit{TabLLM}, which is a general framework to leverage LLMs for few-shot \emph{classification} of tabular data. 
We prompt the LLM with a serialization of a row to a natural-language representation and a short description of the classification problem. 
For risk stratification, for instance, this serialization could list relevant patient attributes and combine it with, ``Will this patient be hospitalized?''.
We experiment with nine different serializations and the T0 language model of different sizes \citep{sanh_multitask_2021}.
We use the parameter-efficient fine-tuning method T-Few \citep{liu_few-shot_2022} to update the LLM's parameters using some labeled examples.
We also evaluate GPT-3 in the zero-shot setting \citep{brown_language_2020}.
To the best of our knowledge, this is one of the widest evaluations of LLMs for zero- and few-shot tabular classification.

Despite its simplicity, we find that TabLLM outperforms prior deep-learning-based tabular classification methods on several benchmark datasets.
By using information from the natural-language column names and feature values, it often enables effective \emph{zero}-shot classification of tabular data.
Unlike many deep learning methods on tabular data, this approach is also competitive with gradient-boosted tree baselines and outperforms them or is on par until $256$ shots.
In the very-few-shot setting it outperforms them by a considerable margin.
The main contributions of this work are:
\begin{itemize}
\setlength{\itemsep}{1.1pt}
    \item We introduce TabLLM, a novel framework leveraging LLMs for data-efficient tabular classification
    \item We study nine serialization techniques and explore their performance across ten different datasets
    \item We show that TabLLM instantiated with a simple text serialization and the T0 LLM can outperform state-of-the-art neural models and tree ensembles in the zero- and few-shot setting
    \item We investigate the application of TabLLM to a large real-world healthcare claims dataset and introduce serialization methods that deal with many input features
\end{itemize}

\section{RELATED WORK}
\subsection{Machine Learning on Tabular Data}
Due to the success of deep learning in  other domains, there have been many recent attempts at representation learning for tabular data.
Self-supervised objectives have largely revolved around the prediction of masked cells, the identification or correction of corrupted cells, and contrastive losses over augmentations \citep{bahri_scarf_2022, somepalli_saint_2021, yoon_vime_2020, arik_tabnet_2021, huang_tabtransformer_2020}.
Additional efforts have included differentiable trees, which combine advantages of tree ensembles with gradient based optimization of neural networks \citep{kontschieder_deep_2015, popov_neural_2019}.
However, several recent comprehensive reviews \citep{shwartz-ziv_tabular_2021, borisov_deep_2022, grinsztajn_why_2022} found that gradient-boosted tree ensembles like XGBoost \citep{chen_xgboost_2016} and LightGBM \citep{ke_lightgbm_2017} systematically outperform  these novel deep learning architectures, even with proper fine-tuning and regularization \citep{kadra2021well}.
\citet{levin_transfer_2022} found utility in transfer learning in the semi-supervised setting, but required a set of additional supervised tasks on the same table, which can be a nontrivial limitation.
They investigate few-shot classification for medical diagnosis using 4 to 200 labeled examples, but do not exploit the power of large pre-trained models, as we do in this work.
\citet{hollmann2022tabpfn} recently introduced TabPFN, a Bayesian neural network pre-trained on synthetic tabular data, outperforming gradient boosted trees in a comprehensive evaluation.

\subsection{Large Language Models for Tabular Data}
Another approach has been to leverage the natural language capabilities of language models.
\citet{yin_tabert_2020} use a language model for semantic parsing of natural language queries over tabular data.
\citet{li_deep_2020} investigate the ability of language models to perform entity matching on tabular data, i.e. determining if two rows refer to the same object.
\citet{harari2022few} study data enrichment by linking each table row with additional unstructured text (e.g., from Wikipedia) from which they generated additional features using a language model. However, this setup requires named entities (e.g., celebrities, universities, etc.), which is quite limiting. 
\citet{bertsimas2022tabtext} studied two healthcare datasets and used a language model to generate feature embeddings, which they fed into classifiers like gradient boosted trees.
All these studies use a BERT-style language model \citep{devlin_bert_2019}.
\citet{narayan_can_2022} recently assessed in-context learning with the autoregressive language model GPT-3 for tabular data cleaning tasks.
They found that it often outperforms state-of-the-art approaches with ten labeled examples.
\citet{borisov2022language} introduced an LLM-agnostic method to generate realistic tabular data and found that it achieved better results than existing approaches.
In contrast, here we study classification tasks of tabular data and investigate parameter-efficient fine-tuning of LLMs.

To use an LLM for tabular data, the table must be serialized into a natural text representation.
All aforementioned works relied on simple list or sentence serializations; \citet{yin_tabert_2020} also included the column data type in the serialized string.
Only \citet{bertsimas2022tabtext} studied different serialization variants, but this was in a different context of deriving feature embeddings from BERT-style language models.
The LIFT method introduced by \citet{dinh2022lift} comes closest to our work.
The authors evaluated the capabilities of fine-tuned GPT-3 and GPT-J models for regression and classification on synthetic, tabular, and vision data.
They also studied the sample efficiency and considered different static serialization templates assessing the effect of including column names in the input.
In this work, we focus on the publicly available T0 model and perform a broader analysis of nine serialization techniques including automatic approaches and ablations evaluating the importance of feature values.
Particularly, we are interested in leveraging prior knowledge encoded in LLMs and we do a more fine-grained analysis of the sample efficiency including zero-shot experiments on ten different datasets.

\section{METHODS}
\subsection{TabLLM for Tabular Data Classification}

\paragraph{Problem Formalization.}
Suppose we have a tabular dataset with $n$ rows and $d$ columns or features.
We can formalize this as $D = \{(\mathbf{x}_i, y_i)\}_{i=1}^n$, where each $\mathbf{x}_i$ is a $d$-dimensional feature vector.
Since we consider classification, $y_i \in C$ for a set of classes $C$.
We define the column names or feature names as $F = \{f_1, ..., f_d\}$.
We assume the $f_i$'s are natural-language strings such as ``age'' or ``education'' (see Figure \ref{fig:overview}).
For our $k$-shot classification experiments, we only use a subset $D_k$ of size $k$---sampled from $D$ with replacement---for fine-tuning or training.

\paragraph{Serialization of Tabular Data.}
To use an LLM for tabular data, the table must be transformed into a natural text representation.
Typically, when prompting an LLM, there is a template used to both serialize the inputs into one natural-language string, and to provide the prompt itself (e.g., the string \textit{``Does this person make more than 50,000 dollars? Yes or no?''}), which is usually located after the serialized input.
In this work, we break these pieces up into a \emph{serialization} and a \emph{prompt}.
We define a function $\texttt{serialize}(F, \mathbf{x})$ that takes the column names $F$ and feature values $\mathbf{x}$ for a row as inputs and creates a textual representation of the input.
Combining this serialization with a task-specific prompt $p$ will then form the LLM input $(\texttt{serialize}(F, \mathbf{x}), p)$.
This is illustrated in Figure \ref{fig:overview}.
We primarily study the serialization, since that is the biggest difference compared to existing applications of prompting.
Previous work has usually considered a simple concatenation of feature names and values as a serialization of tabular data \citep{li_deep_2020, narayan_can_2022}.
In our work, this function can be arbitrarily complex.
For instance, we explore serializations that include (i) incorporating another LLM and (ii) employing feature selection as a substep.

\paragraph{Large Language Models For Classification}
TabLLM can be used with different LLMs that generate text based on a natural-language input.
Let $\texttt{LLM}$ be an LLM with vocabulary $V$.
Then, $\texttt{LLM}((\texttt{serialize}(F, \mathbf{x}), p)) \in V^*$ is the prompted output of the LLM.
In our few-shot setting, $\{(\texttt{serialize}(F, \mathbf{x}), p) \mid  (\mathbf{x}, y) \in D_k\}$ can be used as training examples for fine-tuning the LLM.
The LLM generates text in the vocabulary space $V^*$ that has to be mapped to a valid class in $C$.
Several approaches already exist for this problem.
For example, the verbalizer \citep{schick_exploiting_2021} defines a mapping between LLM output tokens and the discrete label space. 
Verbalizers can be manually specified or automatically learned; see \citet{cui-etal-2022-prototypical} for an overview of different verbalizer-learning approaches.
In this work, we assume for simplicity that the verbalizer mapping is manually specified (see  \texttt{answer} \texttt{\_choices} in the templates in Sec.  \ref{sec:task_templates} in the Supplement).

\subsection{Our Instantiation of TabLLM}

\paragraph{Serialization Approaches for TabLLM.}
The performance of LLMs is very sensitive to the precise details of the natural-language input \citep{zhao_calibrate_2021, webson_prompt-based_2022}.
In this work, we focus on the serialization of the tabular data.
For the prompt, we use a simple description of the classification task and perform no further prompt engineering.
We study nine different serialization formats varying in complexity.
All serialization methods require minimal human effort to apply to new classification tasks.
We evaluate several methods that generate natural text to create inputs that are closer to the training distribution of the LLM, thereby improving zero and very-few-shot performance.
Additional details and examples for the serializations are given in Sec. \ref{subsec:more_details} and \ref{sec:example_serializations} in the Supplement.
\begin{itemize}
    \item \textbf{List Template}: A list of column names and feature values. We fixed an arbitrary ordering of the columns.
    \item \textbf{Text Template}: An textual enumeration of all features as ``The \emph{column name}  is \emph{value}.'' (see Figure \ref{fig:overview}).
    \item \textbf{Table-To-Text}: We use an LLM fine-tuned on a table-to-text generation task from HuggingFace (\texttt{\small Narrativaai/bloom-560m-finetuned-totto -table-to-text}). To ensure that the serialization includes all data we hand each column-value tuple to the model separately and concatenate the outputs.
    \item \textbf{Text T0}: We use the LLM T0 with 11B parameters (\texttt{\small bigscience/T0pp}) \citep{sanh_multitask_2021}. We split up a row into pairs of two column-value tuples. We send them to LLM separately with the prompt ``Write this information as a sentence:'' and combine the outputs.
    \item \textbf{Text GPT-3}: We use GPT-3 (engine \textit{text-davinci-002}) accessible through an API \citep{ouyang_training_2022}. GPT-3 was able to serialize all features at once, so we use a list of all features with the prompt ``Rewrite all list items in the input as a natural text.'' as input. We guide the output with ``The \{person, car, patient\} is''.
    \end{itemize}

We consider the following serializations as ablations:

\begin{itemize}
    \item \textbf{List Only Values}: \emph{List Template} for feature values only. We want to evaluate whether column names aid the classification performance.
    \item \textbf{List Permuted Names}: \emph{List Template} with permuted column names. Hence, the wrong column name is associated with each feature value. The permutation is the same across all examples. We perform this ablation to study the relevance of the correct association between column names and feature values.
    \item \textbf{List Permuted Values}: \emph{List Template} with consistently permuted values across all examples.
    We generate one permutation for each column and apply this mapping to all column values.
    For continuous values, we use ten uniform bins.
    This tests whether the LLM uses the fine-grained information encoded by the feature \emph{values} for zero-shot and few-shot classification.
    \item \textbf{List Short}: \emph{List Template} with at most ten features. We only consider this for the healthcare dataset where the number of features exceeds the input limit of the LLM. We want to study the effect of less information.
\end{itemize}
    
\paragraph{Large Language Models for TabLLM}
Another crucial component of TabLLM is the LLM.
TabLLM is both agnostic to the LLM and the specific fine-tuning method that is used.
We only consider a single LLM for most of our experiments.
We employ the T0 encoder-decoder model with 11 billion parameters as the LLM for TabLLM \citep{sanh_multitask_2021}.
It was trained on a large variety of task-specific prompts, making it a suitable candidate for our experiments \citep{sanh_multitask_2021}.
This model has a token limit of 1024, which roughly corresponds to 400 words.
We also evaluate the effect of a smaller version of the T0 model (T0 3B).
We fine-tuned on the few-shot data $\mathcal{D}_k$ using the recent T-Few recipe, which outperforms other parameter-efficient tuning methods such as soft prompt tuning \citep{liu_few-shot_2022}.
In addition, we perform zero-shot experiments with the LLM GPT-3 (engine \textit{text-davinci-002})  \citep{ouyang_training_2022}.

\section{EXPERIMENTAL SETUP}
\subsection{Datasets}

We studied TabLLM in two experimental settings.
First, we considered nine medium-sized tabular datasets for binary and multi-class classification.
We systematically identified datasets from \cite{kadra2021well}, \cite{grinsztajn_why_2022}, and \cite{borisov_deep_2022}.
We included datasets with at most 50,000 rows to keep the fine-tuning costs manageable and at most 30 columns to stay within T0's token limit.
We also required textual feature names to make the serializations more meaningful and we excluded datasets with derived feature values (e.g., mean pixel values).
This lead to inclusion of \textbf{Bank} (45,211 rows, 16 feats), \textbf{Blood} (748, 4), \textbf{California} (20,640, 8), \textbf{Car} (1,728, 8), \textbf{Credit-g} (1,000, 20), \textbf{Income} (48,842, 14), and \textbf{Jungle} (44,819, 6).
We added two additional datasets from Kaggle that fulfilled our inclusion criteria: \textbf{Diabetes} (768, 8) and \textbf{Heart} (918, 11).
Second, we evaluated TabLLM for risk stratification on three binary classification tasks, following prior work by \citet{kodialam_deep_2020} and similarly using a de-identified health claims dataset from a U.S. health insurer.
We predicted the end-of-life (\textbf{EoL}) of all patients older than 70 years, which can be used to inform care in a palliative setting \citep{avati2018improving}. 
We also considered the need for any surgical procedure (\textbf{Surgery}) and the likelihood of hospitalization (\textbf{LoH}), which can help with determining health care needs and estimating future costs. 
Additional details on all datasets can be found in Sec. \ref{sec:additional_dataset} in the Supplement.
We release the code for our experiments on Github.\footnote{\url{https://github.com/clinicalml/TabLLM}}

\subsection{LLM and Fine-tuning}
We used the HuggingFace implementation of the T0 model (\texttt{\small bigscience/\{T0pp,T0\_3B\}}).
Prompts for the LLM were designed following \citet{sanh_multitask_2021} using the PromptSource framework \citep{bach_promptsource_2022}.
Each class in our classification tasks was manually encoded in a textual response, e.g., ``Yes'' and ``No'' for true and false \citep{sanh_multitask_2021}.
The prediction probability for each class corresponds to the probability of the LLM generating its token sequence normalized across all classes.
All templates used in this work are given in Sec. \ref{sec:task_templates} in the Supplement.

For fine-tuning, we adopted the default hyperparameters of the T-Few method without any additional parameter tuning \citep{liu_few-shot_2022}.
The authors used a setup of $k=32$ shots and 1,000 training steps for most of their experiments, which corresponds to 31.25 epochs.
Hence, we fixed 30 training epochs for all few-shot experiments on the public tabular datasets.
We used 20\% of the data as a test set.
For the large healthcare claims dataset, we used 10 epochs for up to 256 shots and 3 epochs for 1,024, 4,096 and 16,384 to reduce the runtime and prevent overfitting for many training examples.
We used a test set of 10,000 examples for the three healthcare tasks.
All experiments were evaluated with the area under the receiver operating characteristic curve (AUC).
We used macro-AUC one-versus-rest for the multiclass setting.
Estimates for the runtime are given in Sec. \ref{sec:runtime_estimates} in the Supplement.

\subsection{Baseline Models}

We compared TabLLM to several baselines.
For the simplest baseline, we used a logistic regression (LR) model.
Since previous work showed the superiority of gradient boosted tree ensembles \citep{borisov_deep_2022}, we included the most common models XGBoost \citep{chen_xgboost_2016} and LightGBM \citep{ke_lightgbm_2017}.
We also evaluated several state-of-the-art deep learning baselines.
TabNet is a widely used neural model for tabular data that uses attention over columns \citep{arik_tabnet_2021}.
SAINT is a more recent approach that uses attention over rows and columns \citep{somepalli_saint_2021}.
SAINT performed best in a comprehensive review on tabular data \citep{borisov_deep_2022}.
NODE is a differentiable tree ensemble method that performed best in the evaluation of \citet{shwartz-ziv_tabular_2021}.
Lastly, we include TabPFN, a Bayesian neural network that was pre-trained on synthetic tabular data \citep{hollmann2022tabpfn}.
In contrast to TabLLM, we performed hyperparameter tuning for all baselines except TabPFN (see Sec. \ref{sec:parameter_tuning} in the Supplement), which requires no tuning by design.
We adopted the parameter ranges from previous reviews \citep{borisov_deep_2022, grinsztajn_why_2022}.
Since no validation set exists in the few-shot setting, we used 4-fold cross validation on the $k$-shots.
In particular, we \emph{did not} use a large validation set for hyperparameter tuning, unlike some few-shot learning works as highlighted by \citet{perez2021true}.
We encoded categorical values as one-hot vectors.
We also tested ordinal encoding for LR, XGBoost, LightGBM, and TabPFN, but it showed worse results (see Table \ref{table:results_public_dataset_full1}, \ref{table:results_public_dataset_full2}, and \ref{table:results_public_dataset_full3} in the Supplement).
In addition, we give results for GPT-3 (\texttt{text-davinci-002}) without fine-tuning, i.e. in the zero-shot setting using the \textit{Text Template} serialization.

For the three health claims tasks, we used the same experimental setup for the baselines.
However, we only included LR and LightGBM due to runtime limitations.
Following \citet{kodialam_deep_2020}, each patient's input was a one-hot encoded vector.
For each medical concept, there were three indicator variables of whether that concept occurred within 30 days, 1 year, and anytime before prediction time.

\subsection{Serializations}
For the public datasets, some column names and feature values were manually mapped to human-readable forms, based on the provided documentation. 
For instance, for the \textbf{Income} dataset, the feature name \emph{hours\_per\_week} was mapped to \emph{work hours per week} and the feature value \emph{private} for working class was mapped to \emph{private sector employee}.
Numerical values were not changed.

Serialization was more complex for the healthcare claims data.
Each patient record is a time series of visits, with each visit consisting of a list of medical conditions and procedures.
We only considered the manual serializations \emph{List Template} and \emph{Text Template}.
We tried to mimic the style of a medical professional to tap potential prior knowledge of the LLM.
To this end, the serialization starts with an intro sentence containing the patient's gender, age, and race.
It then describes each visit, stating its date, the type of doctor the patient saw (e.g., dermatology) if an outpatient visit or length of hospitalization if an inpatient visit, the primary complaint of the associated visit, and procedures performed.
Since there are no feature values in this dataset, we omit \emph{List Only Values} and \emph{List Permuted Values}.
We also performed experiments for concept selection and different names for the medical concepts.
Details for these additional experiments and examples of the serializations are given in Sec. \ref{subsubsec:concept_selection}, \ref{subsubsec:alternative_concept}, and \ref{sec:example_serializations} in the Supplement.

\section{RESULTS}
\subsection{Effects of serialization}

\begin{figure}[t!]
  \centering
  \vspace{-0.25cm}
  \centerline{\includegraphics[width=0.50\textwidth]{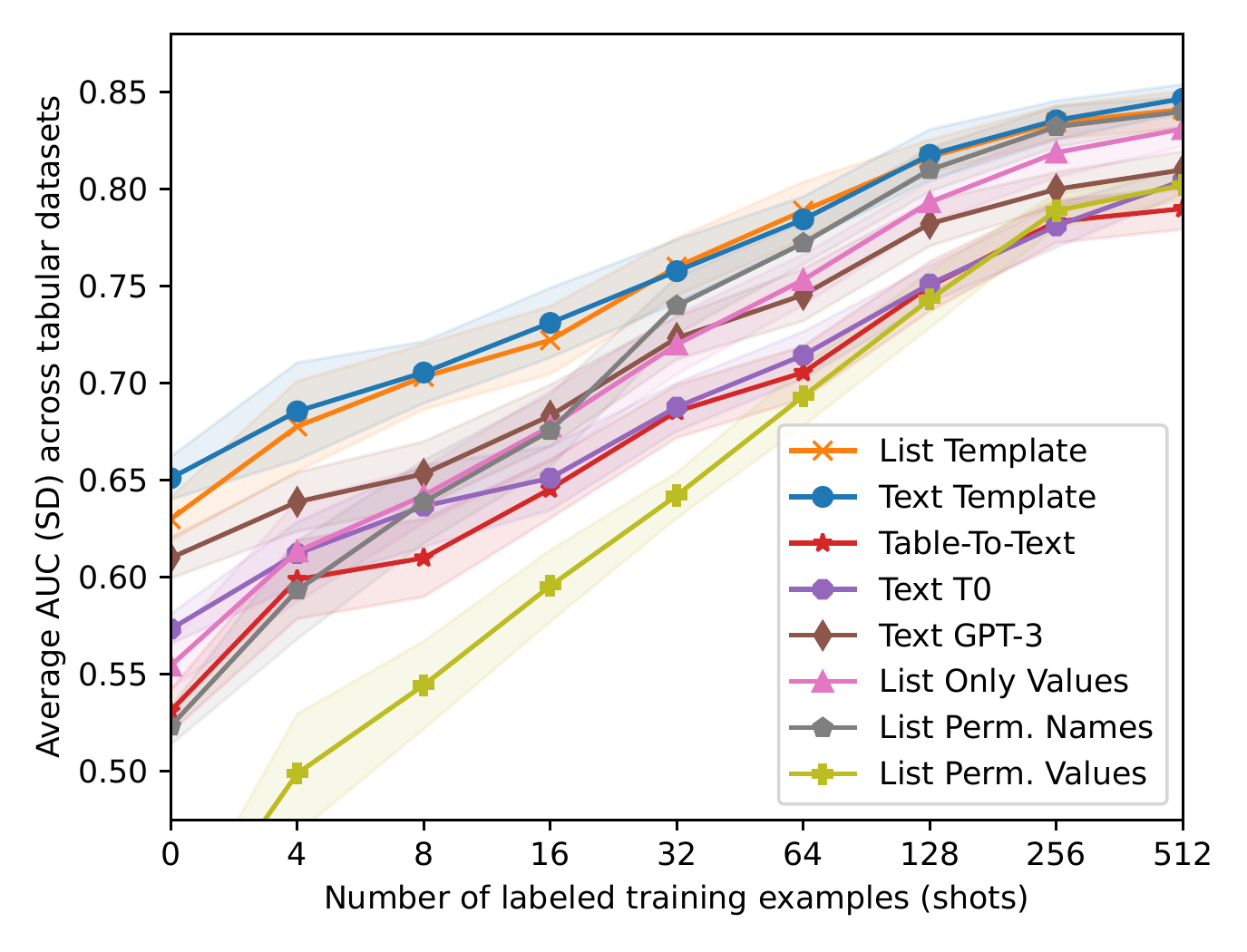}}
  \vspace{-0.2cm}
  \caption{Average AUC and SD of different serializations across nine public datasets. \textit{Text Template} performs best for zero and few training examples. For many examples, the performance of different serializations converges.
  }
  \vspace{-0.4cm}
  \label{fig:serialization-performance}
\end{figure}

Figure \ref{fig:serialization-performance} shows the performance of different serialization methods for TabLLM averaged over the nine public datasets.
The \emph{Text Template} serialization performed very well across all experiments.
In the zero-shot setting, the \emph{Text Template} showed improvements over \emph{List Template}, indicating the benefit of a serialization that is closer to the training distribution of T0.
However, these differences already vanished for 8 training examples.
Hence, very few training examples might already suffice to adjust for different templates. 
This suggests that sophisticated serializations might be unnecessary when some training data exists.

Using LLMs for serialization showed mixed results.
The ordering is according to the complexity of the LLM used for serialization.
GPT-3 has 175B, T0 11B, and the BLOOM table-to-text model 0.56B parameters.
Different reasons might be responsible for the worse performance overall.
The models tended to hallucinate information for some examples, leading to biased predictions of TabLLM.
For instance, GPT-3 added ``this car is a good choice'' or added entirely new data to some examples (see Sec. \ref{sec:example_serializations} in the Supplement). 
Also, the LLMs are not completely faithful at including all features, even though we tried to enforce it in our experiments.
This could explain that none of the LLM serializations reaches the same performance as the template serializations, even for many training examples.

Using only feature \textit{values} had a poor performance for zero and very few shots, but the performance equalized with more training examples.
The same applies to the list serialization with permuted feature names.
This indicates that if enough training examples are available, the serialization approach does not matter, but that TabLLM relies on information from the feature names in the zero-shot and few-shot regime, and also relies on the association of the names with the correct values.
The discrepancy for zero and very few shots was even stronger for \textit{List Permuted Values}, which suggests that TabLLM relies more on the correct values than feature names.
Again, the performance equalized for more examples showing the ability of TabLLM to learn new associations if enough training data is available.
Using the smaller T0 3B model showed a slightly decreased performance (see Table \ref{table:results_public_dataset_full1}, \ref{table:results_public_dataset_full2}, and \ref{table:results_public_dataset_full3} in the Supplement).

\begin{figure}[t!]
  \centering
  \vspace{-0.25cm}
  \centerline{\includegraphics[width=0.50\textwidth]{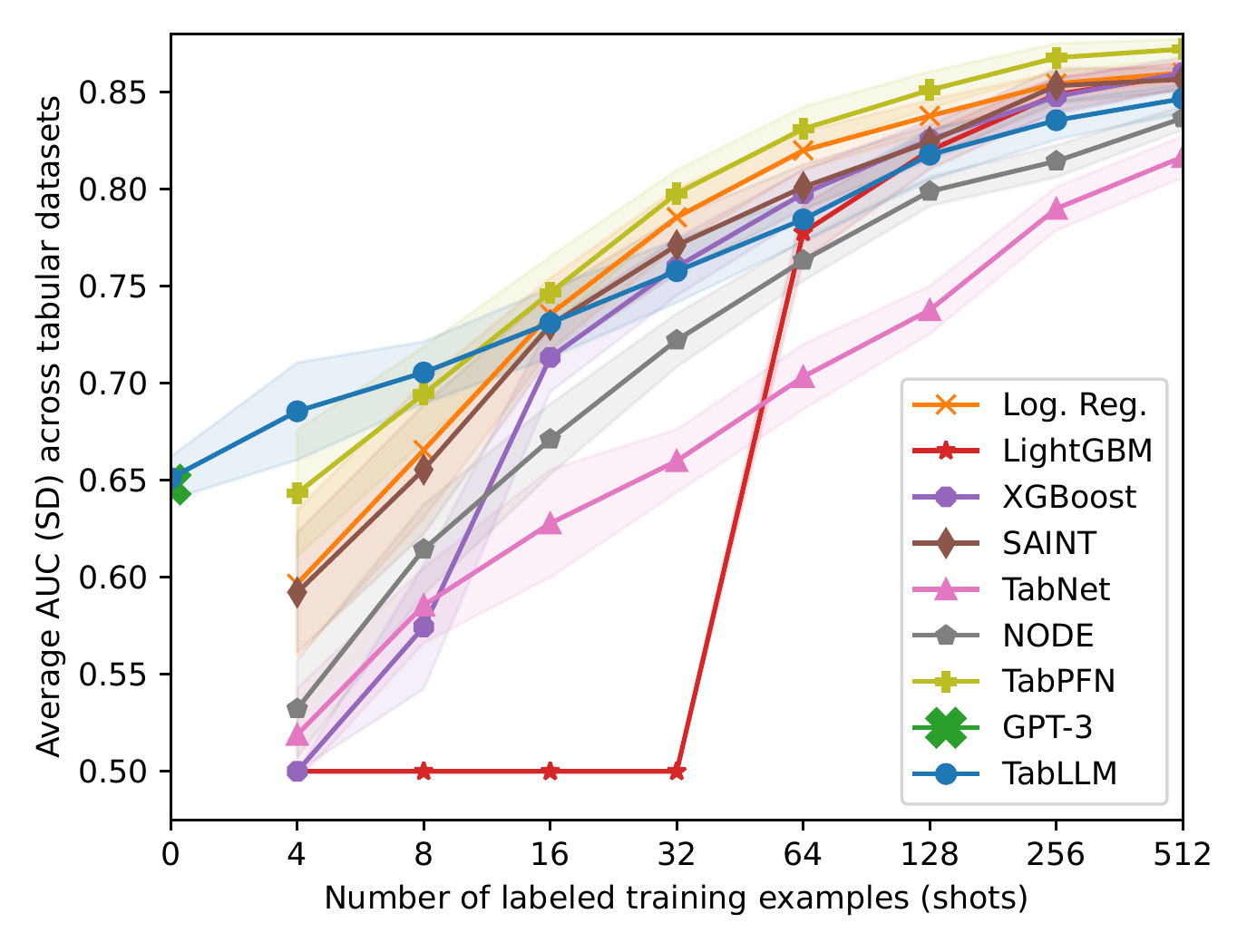}}
  \vspace{-0.2cm}
  \caption{Average AUC and SD of TabLLM versus all baseline models across nine public datasets. TabLLM outperforms all baselines for zero and very few training examples. TabPFN is the strongest baseline.
  }
    \vspace{-0.4cm}
  \label{fig:baselines-performance}
\end{figure}

\begin{table*}[t!]
\caption{Test AUC performance of TabLLM, the best tree ensemble model (XGBoost), and the best baseline (TabPFN) on the public tabular datasets. Each column reports the performance for $k$ training examples. TabLLM (T0 + \textit{Text Template}) outperforms XGBoost and TabPFN in the very-few-shot regime. Standard deviations are given across five random seeds.}
\centering
\setlength{\tabcolsep}{8pt}
\resizebox{\textwidth}{!}{\begin{tabular}{llcccccccccc}
\toprule
& &\multicolumn{10}{c}{\textbf{Number of Shots}}\\
\addlinespace[1.5mm]
\textbf{Dataset} & \textbf{Method} & \textbf{0} & \textbf{4} & \textbf{8} & \textbf{16} & \textbf{32} & \textbf{64} & \textbf{128} & \textbf{256} & \textbf{512} & \textbf{all}\\
\midrule
\multirow{3}{*}{Bank} 
& XGBoost                        & ---             & $0.50_{.00}$   & $0.56_{.09}$   & $0.68_{.04}$   & $0.76_{.03}$   & $\mathbf{0.83}_{.02}$   & $0.85_{.03}$   & $0.88_{.01}$   & $0.90_{.01}$   & $\mathbf{0.94}_{.00}$  \\
& TabPFN                         & ---             & $0.59_{.14}$   & $\mathbf{0.66}_{.08}$   & $\mathbf{0.69}_{.02}$   & $0.76_{.03}$   & $0.82_{.03}$   & $\mathbf{0.86}_{.02}$   & $\mathbf{0.89}_{.00}$   & $0.90_{.00}$   & $0.91_{.00}$  \\
& TabLLM                         & $\mathbf{0.63}_{.01}$    & $0.59_{.10}$    & $0.64_{.05}$    & $0.65_{.05}$    & $0.64_{.06}$    & $0.69_{.03}$    & $0.82_{.05}$    & $0.87_{.01}$    & $0.88_{.01}$    & $0.92_{\phantom{..}}$\dag\\
\midrule
\multirow{3}{*}{Blood} 
& XGBoost                        & ---             & $0.50_{.00}$   & $0.58_{.07}$   & $0.66_{.04}$   & $0.67_{.06}$   & $0.68_{.05}$   & $0.71_{.06}$   & $0.70_{.07}$   & $0.67_{.06}$   & $0.71_{.04}$  \\
& TabPFN                         & ---             & $0.52_{.08}$   & $0.64_{.04}$   & $\mathbf{0.67}_{.01}$   & $\mathbf{0.70}_{.04}$   & $\mathbf{0.73}_{.04}$   & $\mathbf{0.75}_{.04}$   & $\mathbf{0.76}_{.04}$   & $\mathbf{0.76}_{.03}$   & $\mathbf{0.74}_{.03}$  \\
& TabLLM                         & $\mathbf{0.61}_{.04}$    & $\mathbf{0.58}_{.09}$    & $\mathbf{0.66}_{.03}$    & $0.66_{.07}$    & $0.68_{.04}$    & $0.68_{.04}$    & $0.68_{.06}$    & $0.70_{.08}$    & $0.68_{.04}$    & $0.70_{.04}$  \\
\midrule
\multirow{3}{*}{Calhousing} 
& XGBoost                        & ---             & $0.50_{.00}$   & $0.62_{.10}$   & $0.74_{.03}$   & $0.79_{.04}$   & $0.82_{.04}$   & $0.87_{.01}$   & $0.90_{.01}$   & $0.92_{.01}$   & $\mathbf{0.97}_{.00}$  \\
& TabPFN                         & ---             & $0.63_{.13}$   & $\mathbf{0.63}_{.11}$   & $\mathbf{0.80}_{.03}$   & $\mathbf{0.85}_{.03}$   & $\mathbf{0.89}_{.01}$   & $\mathbf{0.91}_{.01}$   & $\mathbf{0.92}_{.00}$   & $\mathbf{0.93}_{.00}$   & $0.94_{.00}$  \\
& TabLLM                         & $\mathbf{0.61}_{.01}$    & $0.63_{.05}$    & $0.60_{.07}$    & $0.70_{.08}$    & $0.77_{.08}$    & $0.77_{.04}$    & $0.81_{.02}$    & $0.83_{.01}$    & $0.86_{.02}$    & $0.95_{.00}$  \\
\midrule
\multirow{3}{*}{Car} 
& XGBoost                        & ---            & $0.50_{.00}$   & $0.59_{.04}$   & $0.70_{.08}$   & $0.82_{.03}$   & $0.91_{.02}$   & $0.95_{.01}$   & $0.98_{.01}$   & $0.99_{.01}$   & $1.00_{.00}$            \\
& TabPFN                         & ---             & $0.64_{.06}$   & $0.75_{.05}$   & $\mathbf{0.87}_{.04}$   & $\mathbf{0.92}_{.02}$   & $\mathbf{0.97}_{.00}$   & $\mathbf{0.99}_{.01}$   & $\mathbf{1.00}_{.00}$   & $1.00_{.00}$   & $1.00_{.00}$  \\
& TabLLM                         & $\mathbf{0.82}_{.02}$    & $\mathbf{0.83}_{.03}$    & $\mathbf{0.85}_{.03}$    & $0.86_{.03}$    & $0.91_{.02}$    & $0.96_{.02}$    & $0.98_{.01}$    & $0.99_{.00}$    & $1.00_{.00}$    & $1.00_{.00}$            \\
\midrule
\multirow{3}{*}{Credit-g} 
& XGBoost                        & ---             & $0.50_{.00}$   & $0.51_{.07}$   & $0.59_{.05}$   & $0.66_{.03}$   & $0.67_{.06}$   & $0.68_{.02}$   & $0.73_{.02}$   & $0.75_{.03}$   & $\mathbf{0.78}_{.04}$  \\
& TabPFN                         & ---             & $0.58_{.08}$   & $0.59_{.03}$   & $0.64_{.06}$   & $0.69_{.07}$   & $0.70_{.07}$   & $\mathbf{0.72}_{.06}$   & $\mathbf{0.75}_{.04}$   & $0.75_{.02}$   & $0.75_{.03}$  \\
& TabLLM                         & $\mathbf{0.53}_{.05}$    & $\mathbf{0.69}_{.04}$    & $\mathbf{0.66}_{.04}$    & $\mathbf{0.66}_{.05}$    & $\mathbf{0.72}_{.06}$    & $0.70_{.07}$    & $0.71_{.07}$    & $0.72_{.03}$    & $0.72_{.02}$    & $0.70_{.02}$  \\
\midrule
\multirow{3}{*}{Diabetes} 
& XGBoost                        & ---            & $0.50_{.00}$   & $0.59_{.16}$   & $\mathbf{0.72}_{.07}$   & $0.69_{.08}$   & $0.73_{.05}$   & $0.78_{.05}$   & $0.80_{.03}$   & $0.80_{.01}$   & $\mathbf{0.84}_{.03}$           \\
& TabPFN                         & ---             & $0.61_{.13}$   & $\mathbf{0.67}_{.11}$   & $0.71_{.07}$   & $\mathbf{0.77}_{.03}$   & $\mathbf{0.82}_{.03}$   & $\mathbf{0.83}_{.03}$   & $\mathbf{0.83}_{.03}$   & $\mathbf{0.81}_{.02}$   & $0.81_{.03}$  \\
& TabLLM                         & $\mathbf{0.68}_{.06}$    & $0.61_{.09}$    & $0.63_{.08}$    & $0.69_{.07}$    & $0.68_{.04}$    & $0.73_{.03}$    & $0.79_{.04}$    & $0.78_{.02}$    & $0.78_{.04}$    & $0.80_{.04}$            \\
\midrule
\multirow{3}{*}{Heart} 
& XGBoost                        & ---            & $0.50_{.00}$   & $0.55_{.14}$   & $0.84_{.07}$   & $0.88_{.04}$   & $0.91_{.01}$   & $0.91_{.01}$   & $0.90_{.01}$   & $0.92_{.01}$   & $0.94_{.01}$            \\
& TabPFN                         & ---             & $\mathbf{0.84}_{.06}$   & $\mathbf{0.88}_{.05}$   & $0.87_{.06}$   & $\mathbf{0.91}_{.02}$   & $\mathbf{0.92}_{.02}$   & $\mathbf{0.92}_{.02}$   & $0.92_{.01}$   & $0.92_{.02}$   & $0.92_{.02}$  \\
& TabLLM                         & $\mathbf{0.54}_{.04}$    & $0.76_{.14}$    & $0.83_{.05}$    & $0.87_{.04}$    & $0.87_{.06}$    & $0.91_{.01}$    & $0.90_{.01}$    & $0.92_{.01}$    & $0.92_{.01}$    & $0.94_{.01}$            \\
\midrule
\multirow{3}{*}{Income} 
& XGBoost                        & ---             & $0.50_{.00}$   & $0.59_{.06}$   & $0.77_{.02}$   & $0.79_{.03}$   & $0.82_{.02}$   & $0.84_{.01}$   & $0.87_{.01}$   & $0.88_{.00}$   & $\mathbf{0.93}_{.00}$  \\
& TabPFN                         & ---             & $0.73_{.08}$   & $0.71_{.09}$   & $0.76_{.09}$   & $0.80_{.04}$   & $0.82_{.04}$   & $0.84_{.01}$   & $0.86_{.01}$   & $0.87_{.01}$   & $0.89_{.00}$  \\
& TabLLM                         & $\mathbf{0.84}_{.00}$    & $\mathbf{0.84}_{.01}$    & $\mathbf{0.84}_{.02}$    & $\mathbf{0.84}_{.04}$    & $\mathbf{0.84}_{.01}$    & $\mathbf{0.84}_{.02}$    & $\mathbf{0.86}_{.01}$    & $0.87_{.00}$    & $\mathbf{0.89}_{.01}$    & $0.92_{.00}$            \\
\midrule
\multirow{3}{*}{Jungle} 
& XGBoost                        & ---             & $0.50_{.00}$   & $0.58_{.07}$   & $\mathbf{0.72}_{.05}$   & $0.78_{.03}$   & $0.81_{.02}$   & $0.84_{.02}$   & $0.87_{.01}$   & $0.91_{.01}$   & $0.98_{.00}$  \\
& TabPFN                         & ---             & $\mathbf{0.65}_{.08}$   & $\mathbf{0.72}_{.04}$   & $0.71_{.07}$   & $0.78_{.02}$   & $0.81_{.01}$   & $0.84_{.01}$   & $\mathbf{0.88}_{.01}$   & $0.91_{.00}$   & $0.93_{.00}$  \\
& TabLLM                         & $\mathbf{0.60}_{.00}$    & $0.64_{.01}$    & $0.64_{.02}$    & $0.65_{.03}$    & $0.71_{.02}$    & $0.78_{.02} $    & $0.81_{.02}$    & $0.84_{.01}$    & $0.89_{.01}$    & $\mathbf{1.00}_{\phantom{..}}$\dag  \\  
\bottomrule
\end{tabular}}
\begin{tablenotes}[flushleft]
    \footnotesize 
    \item[]\dag ~These experiments were only performed for a single run due to runtime limitations of TabLLM on the full dataset.
\end{tablenotes}
\vspace{-0.4cm}
\label{table:results_public_dataset}
\end{table*}

\begin{table}[t!]
\vspace{0.1cm}
\caption{Five highest and lowest weighted features for zero-shot TabLLM and logistic regression (LR) trained on all data for \textbf{Income}. Both models show very similar trends for important features.}
\vspace{-0.2cm}
\setlength{\tabcolsep}{5.5pt}
\small
\begin{tabular}{lrlrl}
\toprule
\textbf{Feature}    & \multicolumn{2}{c}{\textbf{TabLLM}} & \multicolumn{2}{c}{\textbf{LR}} \\
                    &  rank & weight &  rank & weight \\
\midrule
                             capital\_gain &         1 &     5.310 &         2 &     2.393 \\
                        education\_Masters &         2 &     4.623 &         6 &     1.455 \\
                      education\_Doctorate &         3 &     3.410 &         4 &     2.066 \\
                      education\_Bachelors &         4 &     2.995 &         7 &     1.135 \\
                    education\_Prof-school &         5 &     2.949 &         5 &     1.900 \\
\cdashlinelr{1-5}
               occupation\_Priv-house-serv &       102 &    -2.840 &       105 &    -1.909 \\
                           education\_12th &       103 &    -3.178 &        79 &    -0.480 \\
                      education\_Preschool &       104 &    -3.520 &       106 &    -2.385 \\
               occupation\_Farming-fishing &       105 &    -3.853 &        98 &    -0.982 \\
                    workclass\_Without-pay &       106 &    -4.423 &        69 &    -0.174 \\
\bottomrule
\end{tabular}
\vspace{-0.35cm}
\label{table:feature_importance}
\end{table}

For the healthcare claims dataset, we found that the \textit{List Template} slightly outperformed the \textit{Text Template} serialization (see Table \ref{table:results_ibc_dataset_full} in the Supplement).
This was consistent across tasks.
The \textit{List Short} serialization only performed slightly worse.
The evaluation of different concept selection strategies showed that choosing the most frequent conditions per patient performed best.
We found no considerable performance difference for different concept names.

From here onwards, we show results for TabLLM using the \textit{Text Template} serialization for the public datasets.
For the healthcare claims dataset, we use the \textit{List Template} serialization and select the most frequent conditions.
Results for all (dataset, serialization) combinations (Table \ref{table:results_public_dataset_full1}, \ref{table:results_public_dataset_full2}, and \ref{table:results_public_dataset_full3}) and the additional experiments on the healthcare dataset (Table \ref{table:zero_shot_concept_selection} and 
\ref{table:zero_shot_concept_serialization}) can be found in the Supplement.

\begin{table*}
\caption{Test AUC on the healthcare claims dataset. TabLLM outperforms logistic regression (LR) for up to $64$ and LightGBM for up $256$ training examples on End of Life (\textbf{EoL}). Standard deviations are given across five random seeds.}
\centering
\setlength{\tabcolsep}{12.5pt}
\renewcommand{\arraystretch}{0.95}
\resizebox{\textwidth}{!}{\begin{tabular}{llccccccccccc}
\toprule
& & \multicolumn{8}{c}{\textbf{Number of Shots}}\\
\addlinespace[1.5mm]
\textbf{Dataset} & \textbf{Method} & \textbf{0} & \textbf{16} & \textbf{64} & \textbf{256} & \textbf{1,024} & \textbf{4,096} & \textbf{16,384} & \textbf{all}\\
\midrule
\multirow{2}{*}{EoL} &
LR                & ---       & $0.65_{.07}$    & $0.77_{.02}$    & $\mathbf{0.80}_{.02}$    & $\mathbf{0.83}_{.01}$    & $\mathbf{0.83}_{.01}$    &$\mathbf{0.84}_{.01}$   & $\mathbf{0.84}_{.01}$    \\
& LightGBM        & ---       & $0.50_{.00}$    & $0.71_{.01}$    & $0.76_{.02}$    & $0.80_{.01}$    & $0.82_{.01}$    & $0.83_{.01}$    & $0.82_{\phantom{..}}$\dag   \\
& TabLLM   & $\mathbf{0.70}$    & $\mathbf{0.74}_{\phantom{.00}}$    & $\mathbf{0.78}_{\phantom{.00}}$    & $0.78_{\phantom{.00}}$    & $0.79_{\phantom{.00}}$    & $0.81_{\phantom{.00}}$    & $0.81_{\phantom{.00}}$    & ---  \\
\midrule
\multirow{2}{*}{Surgery} &
LR               & ---     & $0.72_{.04}$    & $\mathbf{0.75}_{.05}$    & $0.77_{.01}$    & $0.79_{.01}$    & $0.80_{.01}$    & $0.80_{.00}$    & $0.81_{.00}$  \\
& LightGBM       & ---     & $0.50_{.00}$    & $0.73_{.02}$    & $0.77_{.01}$    & $0.79_{.01}$    & $0.80_{.00}$    & $0.81_{.01}$    & $\mathbf{0.82}_{\phantom{..}}$\dag   \\
& TabLLM         & $\mathbf{0.67}$    & $\mathbf{0.73}_{\phantom{.00}}$    & $0.72_{\phantom{.00}}$    & $0.73_{\phantom{.00}}$    & $0.75_{\phantom{.00}}$    & $0.78_{\phantom{.00}}$    & $0.79_{\phantom{.00}}$    & ---  \\
\midrule
\multirow{2}{*}{LoH} &
LR               & ---     & $0.72_{.04}$    & $\mathbf{0.76}_{.03}$    & $\mathbf{0.80}_{.01}$    & $\mathbf{0.82}_{.01}$    & $0.83_{.01}$    & $0.83_{.01}$    & $0.84_{.01}$  \\
& LightGBM       & ---      & $0.50_{.00}$    & $0.72_{.02}$    & $0.76_{.03}$    & $0.81_{.01}$    & $0.83_{.00}$    & $0.83_{.01}$    & $\mathbf{0.85}_{\phantom{..}}$\dag   \\
& TabLLM             & $\mathbf{0.71}$    & $\mathbf{0.73}_{\phantom{.00}}$    & $0.73_{\phantom{.00}}$    & $0.76_{\phantom{.00}}$    & $0.78_{\phantom{.00}}$    & $0.81_{\phantom{.00}}$    & $0.82_{\phantom{.00}}$    & ---  \\
\bottomrule
\end{tabular}}
\begin{tablenotes}[flushleft]
    \footnotesize 
    \item[]\dag ~These experiments were only performed for a single run due to runtime limitations on the full dataset.
\end{tablenotes}
\vspace{-0.4cm}
\label{table:results_ibc_dataset_maintext}
\end{table*}

\subsection{Public Tabular Datasets}
Figure \ref{fig:baselines-performance} shows the averaged results for TabLLM using the best serialization (\textit{Text Template}) versus all baseline models.
Table \ref{table:results_public_dataset} contains the detailed results for TabLLM, TabPFN, and XGBoost.
TabLLM showed a similar behavior across datasets.
It achieved nontrivial zero-shot performance for all tasks except on \textbf{Credit-g} and \textbf{Heart}.
For \textbf{Heart} this might be due to the dataset's inclusion criteria requiring eligibility for a heart procedure biasing the prediction.
In all cases, TabLLM's performance improved with a higher number of shots.
In the zero-shot setting, TabLLM was on par with GPT-3 even though GPT-3 is a much larger model than T0 (175B vs. 11B parameters).
TabPFN consistently outperformed the other baseline models across all numbers of training examples.
TabPFN reached TabLLM's performance with 4 to 256 (\textbf{Income}) training examples.
LR was the second-best baseline often beating the tree models, which might be due to our extensive parameter tuning (see Sec. \ref{sec:comparing_baseline} in the Supplement).
TabLLM outperformed or was on par with the tree ensemble baselines until 256 training examples for all datasets except \textbf{Calhousing} and \textbf{Jungle}.
For fewer shots, it often outperformed them by a large margin.
XGBoost performed relatively poorly for few shots, which was probably due to overfitting on the small training and validation sets (as described in the previous section, we do \emph{not} use large validation sets for hyperparameter tuning to ensure the results are truly few-shot).
TabLLM outperformed the neural baselines SAINT, NODE, and TabNet in many settings.
It also was on par or very close to the best baseline models on the full datasets, indicating that there is little performance lost due to the serialization and the choice of model family.

\begin{table}[!t]
\vspace{0.1cm}
\caption{Five highest and lowest weighted features for zero-shot TabLLM for \textbf{EoL} and their relative risk (RR) with confidence intervals (CI). The top five features show a significant increase of the relative risk.}
\vspace{-0.2cm}
\setlength{\tabcolsep}{3.4pt}
\small
\begin{tabular}{lcc}
\toprule
\textbf{Feature}    & \multicolumn{1}{c}{\textbf{TabLLM}} & \multicolumn{1}{c}{\textbf{RR (95\% CI)}} \\
\midrule
               atrial fibrillation &  0.633 & 2.72 (2.51-2.95) \\
atherosclerosis of coronary art... &  0.530 & 2.10 (1.94-2.27) \\
          atherosclerosis of aorta &  0.473 & 1.99 (1.81-2.19) \\
exudative age-related macular d... &  0.452 & 2.38 (2.06-2.75) \\
                          sex\_male & 0.442 & 1.23 (1.14-1.33) \\
\cdashlinelr{1-3}
open angle with borderline intr... &  -0.338 & 1.20 (1.03-1.40) \\
primary localized osteoarthrosi... &  -0.366 & 1.08 (0.82-1.43) \\
 localized, primary osteoarthritis &  -0.393 & 1.23 (1.07-1.40) \\
                       sex\_female &  -0.441 & 0.81 (0.75-0.88) \\
  open-angle glaucoma - borderline &  -0.495 & 0.97 (0.85-1.10) \\
\bottomrule
\end{tabular}
\vspace{-0.35cm}
\label{table:feature_importance_ibc}
\end{table}

\paragraph{Introspecting TabLLM---What Prior Knowledge Does it Use?}
Given the strong zero-shot performance of TabLLM on the \textbf{Income} dataset, we next sought to understand which features it based its predictions on in order to shed light on the prior knowledge used by the LLM. 
To determine the feature importance for TabLLM, we fit a LR model to the zero-shot prediction using the original features as covariates as described in Sec. \ref{sec:feature_importance} in the Supplement.
Highly weighted features (see Table \ref{table:feature_importance}) for zero-shot TabLLM include the individual's occupation  (with e.g., `Farming-fishing' having a large negative weight), highest education level (`Masters' and `Doctorate' have positive weights; `Preschool' grade has a negative weight), and workclass (`Without-pay' has a negative weight).
TabLLM also seems to be able to correctly interpret the numerically encoded capital gain value.
For comparison, we also show the feature weights for a LR model trained on all data.
We see a strong concordance between both models; TabLLM's top five features are all among the top seven of the LR model.
However, TabLLM scores the highest education degrees in the opposite order.
Table \ref{table:feature_importance_full} in the Supplement shows the importance of all 106 features.

\subsection{Large Healthcare Claims Dataset}
Table \ref{table:results_ibc_dataset_maintext} shows the results for TabLLM with the \textit{List Template} serialization on \textbf{EoL}, \textbf{Surgery}, and \textbf{LoH}, the three prediction tasks for the healthcare claims dataset.
TabLLM showed very considerable zero-shot performance, ranging from 0.67 AUC for \textbf{Surgery} to 0.71 for \textbf{LoH}.
The performance improves with higher number of training examples.
However, the performance jumps happen at different steps and to a different extent.
TabLLM outperformed LR for up to 16 (\textbf{Surgery} and \textbf{LoH}) to 64 (\textbf{EoL}) training examples and LightGBM for up to 64 (\textbf{LoH}) and 256 (\textbf{EoL}) examples.
For more examples, LR and LightGBM performed slightly better.
This could suggest that the information lost from our concept selection procedure, needed because of the token limits of the LLM, eventually starts costing TabLLM performance.
We also evaluated TabLLM and LR in an unbalanced setting (see Table \ref{table:results_ibc_dataset_full} in the Supplement). 
In this case, TabLLM outperforms LR up to 64 training examples on all datasets emphasizing its utility in a real world setting with limited access to labeled data.

\paragraph{Introspecting TabLLM---What Prior Knowledge Does it Use?}
We also performed a feature analysis to study the strong zero-shot performance on \textbf{EoL}.
However, we did not compare to a LR model trained on all data due to the vast amount of features and potential colinearites in the data.
Instead, we compared to the relative risk (RR) with a 95\% confidence interval (CI).
Table \ref{table:feature_importance_ibc} shows the five highest and lowest weighted features of zero-shot TabLLM and their relative risk for \textbf{EoL}.
All top five features have a significantly increased relative risk demonstrating the capabilities of TabLLM to identify relevant features even without any training examples.
For the five lowest weighted features, only `sex\_female' has a significantly decreased risk.
A list of 100 features is given in Table \ref{table:feature_importance_ibc_full} in the Supplement.

\section{DISCUSSION}
For all datasets except \textbf{Credit-g} and \textbf{Heart}, the \textit{List Template} and \textit{Text Template} serializations showed nontrivial zero-shot performance, indicating that TabLLM is able to effectively utilize prior knowledge in the LLM for classification.
Serializations with LLMs proved suboptimal due to their noisy outputs suggesting that simple templates are preferable for TabLLM.
The performance drops observed when we removed or permuted the column names indicate that the LLM actually makes use of feature names and their relationships to the correct values, especially in the few-shot setting.
These findings are partly consistent with \citet{dinh2022lift} who used GPT-3 and tested serializations with removed or permuted column names.
When using all training examples, they showed that using the correct column names led to the best performance on four classification tasks.
In contrast to our results, however, they could not confirm these findings when using only a fraction (0.2, 0.4, 0.6, 0.8) of the training data.
A reason for this could be that we tested much fewer number of training examples.
In addition to that, we found a very strong drop in performance for permuted values showing that the LLM relies more on the correct values than feature names.
Surprisingly, however, all serializations with less information came close to the best serialization for 256 (tabular datasets) to 1024 training examples (insurance dataset).
Hence, when hundreds of training examples are available, the input format proved less relevant, and the LLM was able to adapt  \citep{jin2022a}.
Like our results, \citet{bertsimas2022tabtext} found that natural language representation of healthcare data gave little-to-no improvement (in their different setup) compared to a more straightforward serialization in the medium-shot setting.
Our findings also support prior work showing that irrelevant and even misleading inputs can lead to similar few-shot performance \citep{min2022rethinking, webson_prompt-based_2022, reynolds2021prompt}.
For instance, permuting the column names only showed a difference for up to 16 training examples (see Figure \ref{fig:serialization-performance}).

We found clear performance improvements for TabLLM when using additional training examples.
It often outperformed strong baseline models in the very-few-shot setting.
This emphasizes the value of leveraging LLMs when only little labeled data is available.
Surprisingly, \citet{dinh2022lift} could not confirm these findings for GPT-3.
On two binary classification tasks a fine-tuned GPT-3 model performed worse than LR for up to 250 training examples.
Our results indicate that the sample efficiency of TabLLM is highly task-dependent.
The performance on \textbf{Blood}, \textbf{Credit-g}, \textbf{Diabetes}, and \textbf{Heart} is worse than the performance on \textbf{Income} and \textbf{Car}.
Most features of the latter datasets have semantically meaningful textual values likely boosting TabLLM's performance.
However, TabLLM also achieved reasonable results on numerical datasets (\textbf{Blood}, \textbf{California}, \textbf{Diabetes}, and \textbf{Jungle}).
In addition, \textbf{Diabetes} and \textbf{Heart} have somewhat specialized feature names and values, such as ``ventricular hypertrophy'' and ``Plasma glucose concentration,'' whereas \textbf{Income} and \textbf{Car} are more general-domain knowledge.
This indicates that T0, the language model we used in TabLLM, seems to have less prior knowledge about medicine than about general-domain concepts.
Indeed, the training tasks for T0 do not contain any tasks with medical data \citep{sanh_multitask_2021}.

Our findings on the three insurance claims datasets partly reinforce this hypothesis.
Zero-shot performance depends on the concept selection strategy and the LLM seems to have little knowledge about medical procedures. 
Prior work has shown that medical-domain-specific language models, such as PubMedBERT, and general-domain models with medical data in their training sets, such as GPT-3, perform well at downstream prediction tasks on medical data even with fairly few samples \citep{gu2021domain, agrawal_large_2022}.
Substituting T0 with one of these models in TabLLM to study medical predictions tasks is an interesting direction for future work.

Our results on the public \textbf{Blood}, \textbf{Diabetes}, and \textbf{Heart} datasets are very similar to our results for \textbf{EoL}, \textbf{Surgery}, and \textbf{LoH}, which are practically relevant but rely on private data.
Except for the zero-shot and very few-shot regime, other baselines tend to outperform TabLLM on these datasets.
This suggests that \textbf{Blood}, \textbf{Diabetes}, and \textbf{Heart} datasets could be good proxies for the community to further study medical-domain tabular classification with LLMs without needing access to large private datasets.

\section{LIMITATIONS AND CONCLUSION}
TabLLM has a much larger computational footprint compared to traditional algorithms.
It still requires fairly large GPUs to fine-tune the LLM, and inference with T0 requires far more FLOPs than inference with XGBoost or LR.
Our results indicate that TabLLM trades off this computational efficiency for improved sample efficiency.
Further, as we saw with the three healthcare claims tasks, performance may suffer if the dense feature set for a given row cannot fit within the token limit for a given LLM. 
Since the gains from TabLLM stem from its ability to use existing domain knowledge, the semantics of the column names and feature values need to have been observed during the LLM's original pre-training.
For example, if the columns represent genes, we may not expect a vanilla LLM to have strong representations for gene names. 
Finally, due to dataset shift, the pre-training data for a given LLM may not necessarily reflect the settings under which a given table was aggregated, e.g., due to inflation and a changing value of money (see Sec. \ref{sec:adjusting_income} in the Supplement).

Despite these limitations, our empirical results show that TabLLM enjoys strong performance at tabular classification, outperforming state-of-the-art baseline algorithms like XGBoost and SAINT by over 5 AUC points in the very-few-shot regime, all while staying competitive with these methods when a large number of samples is available.

Currently, TabLLM does not use any \textit{unlabeled} data; a fruitful direction could involve leveraging unlabeled data, e.g., using the techniques from \citet{lang_co-training_2022} to combine the few-shot performance of TabLLM with the ultimate performance of tree-based baselines by co-training the models together. Other improvements could include more faithful LLM serializations as well as numeric-specific encoding methods \citep{gorishniy2022embeddings}.

\section{SOCIETAL IMPACT}
Similar to other ML systems that were trained on historic data, LLMs are prone to replicate existing biases and stereotypes.
Hence, when applying TabLLM for sensitive tasks such as income or a health trajectory, predictions should be considered with great care and further analyses (e.g., for subgroups) are mandatory.
In addition, LLMs require a lot of computing resources.
This bears the risk of creating an exclusive research environment.
Also, the environmental impact of LLMs can be significant.

\section{ACKNOWLEDGEMENTS}
SH was supported by the German Academic Exchange Service, HL by NSF AiTF award CCF-1723344, MA by a Takeda Fellowship, and DS, HL, AB, and SH in part by Independence Blue Cross.
Thanks to Dr. Steven Horng for generously donating GPU-time on the BIDMC computing cluster \citep{horng2022} and to NVIDIA Corporation for their donation of two NVIDIA A100 GPUs used in this work.

\bibliography{references.bib}

\setcounter{section}{0}

%

%

\onecolumn
\aistatstitle{Supplementary Materials:

TabLLM: Few-shot Classification of Tabular Data with Large Language Models
}

\section{ADDITIONAL DATASET DETAILS}
\label{sec:additional_dataset}

\subsection{Public Tabular Datasets}

We systematically identified datasets for classification  from \cite{kadra2021well}, \cite{grinsztajn_why_2022}, \cite{borisov_deep_2022}, and from Kaggle.
Each dataset was separated into $80/20$ train-test splits.
The $k$ labeled examples $\mathcal{D}_k$ were sampled in a class-balanced manner from the training set.
We performed experiments for different numbers of trainings examples (shots) ranging from $0$ to $512$ and the entire dataset (all).
To characterize the sensitivity of models to the choice of $k$ labeled examples, we repeated the dataset splitting and sampling procedures for five different seeds and report the mean AUC and standard deviation (SD) across seeds.
No hyperparameter tuning was conducted for TabLLM; for baselines, internal cross validation was conducted to choose optimal hyperparameters, and the model was then retrained on all data.
We analyzed the following datasets:

\begin{itemize}
    \item \textbf{Bank} \citeponline{kadra2021well} contains information of a direct marketing campaign from a Portugese banking institution \citeponline{moro2014data}. The goal is to predict whether a customer subscribed to a term deposit or not. It consists of 45,211 rows and 16 features; 5,289 labels are positive.
    \item \textbf{Blood} \citeponline{kadra2021well} consists of data of a blood transfusion service from Taiwan  \citeponline{yeh2009knowledge}. It contains 4 attributes of 748 donors and the label is representing whether they returned for another donation (178 positive).
    \item \textbf{California} \citeponline{grinsztajn_why_2022} contains eight attributes of 20,640 districts in California and the goal is to predict the median house value in each district \citeponline{pace1997sparse}. Analogously to \cite{grinsztajn_why_2022}, we created a balanced classification task by predicting whether the house value is below or above the median (10,317 positive). 
    \item \textbf{Car} \citeponline{kadra2021well} has entries for different cars that are characterized by six attributes; the task is a multiclass classification problem evaluating the state of each car. The dataset contains 1,728 rows, and the four classes have a distribution of 1210, 384, 65, and 69 examples.
    \item \textbf{Credit-g} \citeponline{kadra2021well} describes 1,000 people from Germany that want to receive a credit using 20 attributes. The label is to predict whether they have good or bad risk; 700 are classified as good. 
    \item \textbf{Diabetes} (from Kaggle\footnote{https://www.kaggle.com/datasets/uciml/pima-indians-diabetes-database \allowbreak(06/28/2022)}) was collected by the National Institute of Diabetes and Digestive and Kidney Diseases \citeponline{smith_using_1988} and contains 768 rows, each corresponding to women of Pima Indian heritage with eight clinical variables. The task is binary classification of whether a person has diabetes; 268 cases are positive.
    \item \textbf{Heart} (from Kaggle\footnote{ https://www.kaggle.com/fedesoriano/heart-failure-prediction\allowbreak (06/28/2022)}) contains data of four different hospitals \citeponline{detrano1989international}. Each row contains 11 clinical variables of a patient. The task is binary classification of coronary artery disease. Of the 918 patients, 508 are positive.
    \item \textbf{Income} \citeponline{kadra2021well, borisov_deep_2022} also called Adult contains rows for 48,842 individuals with twelve attributes collected in the 1994 U.S. Census \citeponline{kohavi1996scaling, Dua_2019}. The task is to predict whether each person has an annual income over \$50,000. The dataset has 11,687 positive labels.
    \item \textbf{Jungle} \citeponline{kadra2021well} is a collection of 44,819 end game positions of Jungle Chess \citeponline{van2014endgame}. Each game is described with 6 attributes and the goal is to predict whether the white player will win (23,062 positive). 
\end{itemize}

\vspace{0.8cm}

\subsection{Large Healthcare Claims Dataset}

The de-identified health claims data set was provided by a large U.S. health insurer.
The data is stored in the  Observational Medical Outcomes Partnership (OMOP) Common Data Model version 6.0 \citeponline{hripcsak_observational_2015}.
It contains an entry for every encounter a patient has with the health system.
Each entry is associated with a date, a visit type (5 total), a medical specialty (216 total), present conditions (14,095 total), and performed procedures (21,184 total).
We additionally used the static concepts age, sex, and race at time of prediction.

We studied three different tasks on this dataset with distinct cohorts.
For all tasks, we used a six month outcome period and a gap of three months between time of prediction and the outcome window to prevent data leakage.
We required patients to have at least one medical visit and to have been actively enrolled in an insurance plan for at least 95\% of the last year and the six month outcome window.
We used 10\% of the data as a holdout set and sampled the $k$ balanced shots with replacement from the remaining data.
We chose larger shot sizes, as the tasks are more complex.
We only ran the experiments for a single seed due to runtime limitations.
We considered the following tasks:

\begin{itemize}
    \item \textbf{End of Life (EoL)}: We predicted the mortality of all patients older than 70 years. This is often used as a surrogate task. For instance, it can improve initiation of palliative care \citeponline{avati2018improving} and can help to inform close relatives to reduce family distress \citeponline{curtis2016randomized}. The final cohort contained 94,972 individuals; 2,424 were positive.
    \item \textbf{Surgical Procedure (Surgery)}: We predicted the need for any surgical procedure. The task is important in determining health care needs and estimating costs. 
    The cohort included 620,382 people of which 243,349 were positive.
    \item \textbf{Likelihood of Hospitalization (LoH)}: We also predicted the likelihood of being hospitalized. Again, this information can help identify needs and estimate costs. The cohort included 612,656 individuals; 22,427 were positive.
\end{itemize}

\subsubsection{More Details on the Serialization}
\label{subsec:more_details}

Each serialization begins with the patient's age, sex, and race.
For each concept entry that we included, we also added information of the associated visit.
This included its date, the type of doctor the patient saw (e.g., dermatology), if an outpatient visit or length of hospitalization if an inpatient visit, and the primary complaint of the associated visit.
If a visit was already added to the serialization, we just added the concept to the existing visit entry.
For the \emph{List Template} and \emph{Text Template} serializations approximately 40 medical concepts could be added until the token limit of T0 was reached.
To explore the effect of fewer information in the input, we also tested the \emph{List Short} serializations were we added only 10 medical concepts to the serialization.
Hence, not the entire token limit of the LLM was used.
Examples of the \emph{List Template}, \emph{Text Template} and \emph{List Permuted Names} serializations illustrating this structure are given in Sec. \ref{subsec:example_serializations_ibc} at the end of the Supplement.

\begin{table}[!t]
\caption{Evaluation of different concept selection methods for the healthcare claims dataset in the zero-shot setting. The last two rows show the performance when concepts where selected based on the lasso path of logistic regression weights, which violates the zero-shot assumption (*).}
\vspace{-0.2cm}
\small
\begin{tabular}{lccc}
\toprule
\textbf{Method}                         & \textbf{EoL}  & \textbf{Surgery} & \textbf{LoH} \\
\midrule
Age, sex, and race               & $0.59$          & $0.57$          & $0.65$  \\
\midrule
Least frequent conditions                & $0.57$          & $0.64$          & $0.67$  \\
Least frequent procedures                & $0.59$          & $0.59$          & $0.65$  \\
Least frequent concepts (cond. + proc.)  & $0.55$          & $0.55$          & $0.66$  \\
Most frequent conditions                 & $0.67$          & $0.66$          & $0.69$  \\
Most frequent procedures                 & $0.59$          & $0.58$          & $0.65$  \\
Most frequent concepts (cond. + proc.)   & $0.62$          & $0.61$          & $0.65$  \\
Oldest conditions                        & $0.65$          & $0.66$          & $0.69$  \\
Oldest procedures                        & $0.59$          & $0.58$          & $0.65$  \\
Oldest concepts (cond. + proc.)          & $0.60$          & $0.60$          & $0.67$  \\
Most recent conditions                   & $0.65$          & $0.66$          & $0.69$  \\
Most recent procedures                   & $0.55$          & $0.59$          & $0.65$  \\
Most recent concepts (cond. + proc.)     & $0.59$          & $0.60$          & $0.66$  \\
\midrule
Most relevant concepts based on 256 shots*  & $0.60$          & $0.58$          & $0.69$  \\
Most relevant concepts based on 4096 shots* & $0.65$          & $0.57$          & $0.68$  \\
\bottomrule
\end{tabular}
\label{table:zero_shot_concept_selection}
\end{table}

\subsubsection{Concept Selection}
\label{subsubsec:concept_selection}

For the healthcare claims dataset, the number of recorded medical concepts per patients usually exceeded T0's token limit.
Hence, we had to determine which concepts of a patient should be included during the serialization.
We evaluated four different concept selection strategies in the zero-shot setting for the \emph{List Template} serialization.
Choosing the least frequent, most frequent, oldest, or most recent concepts per patient.
We tested these for all concepts (conditions and procedures), only conditions, or only procedures.
For each patient, we ranked all concepts according to one of the above methods and added concepts until the token limit of the LLM was reached.
For least frequent and most frequent, we used the earliest visits associated with the selected medical concepts.
We used a simple serialization that only contained the patient's age, sex, and race as a baseline for our experiments.
We also tested concept selection based on the lasso path of a logistic regression model determined on 256 and 4,096 shots.
This violates the few-shot assumption, but we considered it an interesting comparison with the other strategies that select concepts per patient.

\begin{table}[t!]
\caption{Five examples of different concept names for conditions. The first column shows the original name in the healthcare claims dataset using SNOMED codes. A dash illustrates that no mapping was available.}
\vspace{-0.2cm}
\setlength{\tabcolsep}{4pt}
{\renewcommand{\arraystretch}{1.2}
\small
\begin{tabular}{p{0.15\textwidth}p{0.15\textwidth}p{0.15\textwidth}p{0.15\textwidth}p{0.15\textwidth}p{0.15\textwidth}}
\midrule
\textbf{Original name}            & \textbf{ICD}                                      & \textbf{MEDCIN}                      & \textbf{CHV}                      & \textbf{Simplify (GPT-3)}                       & \textbf{Jargon (GPT-3)}             \tabularnewline \midrule
\raggedright{Seasonal allergic rhinitis}       & \raggedright{Allergic rhinitis due to pollen}          & hay fever                   & hay fever                & Allergies                                & \raggedright{Seasonal allergic rhinitis}           \tabularnewline \midrule
\raggedright{Disturbance in speech}            & \raggedright{Unspecified speech disturbances}          & \raggedright{speech difficulties}         & \raggedright{speech impairment}        & \raggedright{Speech problems}                          & \raggedright{Dysarthria}                           \tabularnewline \midrule
\raggedright{Congenital duplication of cervix} & ---                                      & ---                         & \raggedright{double cervix}            & \raggedright{Double cervix}                            & \raggedright{Congenital duplication of the cervix} \tabularnewline \midrule
\raggedright{Hypertensive retinopathy}         & \raggedright{Hypertensive retinopathy}                 & \raggedright{hypertensive retinopathy}    & \raggedright{hypertensive retinopathy} & \raggedright{High blood pressure affecting the retina} & \raggedright{Retinopathy h-tensa}                  \tabularnewline \midrule
\raggedright{Malignant neoplasm of liver}      & \raggedright{Malignant neoplasm of liver, unspecified} & \raggedright{malignant neoplasm of liver} & \raggedright{liver cancer}             & \raggedright{Liver cancer}                         & \raggedright{Hepato-ca}                            \tabularnewline \bottomrule
\end{tabular}}
\label{table:alternative_concept_names}
\end{table}

\begin{table}[!t]
\caption{Evaluation of alternative condition concepts names. International Classification of Diseases (ICD), MEDCIN and the Consumer Health Vocabulary (CHV) are alternative medical terminologies. We also tested shortening, simplifying, and rewriting concepts as medical jargon via GPT-3. None of the alternative concept names showed consistent performance improvement.}
\vspace{-0.2cm}
\small
\begin{tabular}{lccc}
\toprule
\textbf{Method}             & \textbf{EoL}  & \textbf{Surgery} & \textbf{LoH} \\
\midrule
Original concept names (SNOMED)      & $0.67$          & $0.66$          & $0.69$  \\
\midrule
Map to ICD concept names             & $0.67$          & $0.67$          & $0.68$  \\
Map to MEDCIN concept names          & $0.67$          & $0.66$          & $0.69$  \\
Map to CHV concept names             & $0.66$          & $0.66$      & $0.69$  \\
Shorten longs concepts with GPT-3    & $0.67$          & $0.66$          & $0.69$  \\
Simplify concepts with GPT-3         & $0.67$          & $0.66$          & $0.70$  \\
Medical jargon with GPT-3            & $0.68$          & $0.67$          & $0.70$  \\
\bottomrule
\end{tabular}
\label{table:zero_shot_concept_serialization}
\end{table}

The results are given in Table \ref{table:zero_shot_concept_selection}.
Using the most frequent conditions per patient consistently outperformed all other selection strategies.
Frequent conditions might be useful since they reveal the most relevant condition of a patient.
Also, they are usually more common allowing more prior knowledge of the LLM.
Across all strategies conditions were usually more useful than procedures.
This suggests more prior knowledge of conditions.
Interestingly, selecting the most frequent conditions is even better than using the concept weights of a LR model trained on 256 or 4,096 shots.

\subsubsection{Alternative Concept Names}
\label{subsubsec:alternative_concept}

The healthcare claims dataset used SNOMED concept names for conditions and SNOMED, Healthcare Common Procedure Coding System (HCPCS), International Classification of Diseases (ICD), and Current Procedural Terminology (CPT) concept names for procedures.
We tested different concept names to assess their effect on the performance.
We used a zero-shot setting with the \emph{List Template} serialization and the most frequent conditions per patient as the best selection strategy determined as described above.
Since the selection method only considered conditions, we only used different condition names.
We considered three alternative vocabularies in the Unified Medical Language System (UMLS) that covered at least 20\% of the condition concepts and offered different names.
ICD is a very common medical terminology offering alternative names for conditions.
MEDCIN and the Consumer Health Vocabulary (CHV) offer concept names specifically targeted at clinicians or consumers.
We mapped the concept via their UMLS identifier.
For ICD we were able to map 7,372, for MEDCIN 9,370 and for CHV 3,700 of the 14,095 condition concepts.
Alternatively, we explored concept names generated by GPT-3 \citeponline{brown_language_2020}.
To do so, we used the publicly accessible GPT-3 API (engine \textit{ text-davinci-002}) \citeponline{ouyang_training_2022}.
We considered shortened names for concepts with more than sixty character (``Rewrite this medical condition with at most six words.''), simplified concept names (``Write this medical condition in a short form in lay language.'') and medical jargon (``Write this medical condition in medical jargon.'').
For the simplified names and the medical jargon, we provided GPT-3 with a single example for in-context learning.
Examples for all alternative concept names except the shortening are given in Table \ref{table:alternative_concept_names}.

The results of this experiment are given in Table \ref{table:zero_shot_concept_serialization}.
We used the most frequent concept as a concept selection methods.
Based on the best concept selection, we performed additional experiments for alternative concept names.
We found no consistent performance difference even though there were considerable differences in the concept names (see Table \ref{table:alternative_concept_names}).
Surprisingly, TabLLM performs better for \textbf{EoL} and \textbf{Surgery} using medical jargon to encode concepts.

\begin{table}[t!]
    \caption{Hyperparameters for LR model.}
\vspace{-0.2cm}
    \small
    \begin{tabular}{ll}
    \toprule
    \textbf{Parameter} & \textbf{Values}                                   \\
    \midrule
    penalty   & `l1', `l2'                               \\
    C         & 100, 10, 1, 1e-1, 1e-2, 1e-3, 1e-4, 1e-5 \\
    \bottomrule
    \end{tabular}
    \label{table:hyperparamater_lr}
\end{table}

\begin{table}[t!]
    \caption{Hyperparameters for LightGBM model.}
\vspace{-0.2cm}
    \small
    \begin{tabular}{ll}
    \toprule
    \textbf{Parameter}           & \textbf{Values}                            \\
    \midrule
    num\_leaves     & 2, 4, 8, 16, 32, 64, 128, 256, 512, 1024, 2048, 4096     \\
    lambda\_l1      & 1e-8, 1e-7, 1e-6, 1e-5, 1e-4, 1e-3, 1e-2, 1e-1, 1., 10.  \\
    lambda\_l2      & 1e-8, 1e-7, 1e-6, 1e-5, 1e-4, 1e-3, 1e-2, 1e-1, 1., 10.  \\
    learning\_rate  & 0.01, 0.03, 0.1, 0.3 \\
    \bottomrule
    \end{tabular}
    \label{table:hyperparamater_lightgbm}
\end{table}

\begin{table}[t!]
    \caption{Hyperparameters for XGBoost model.}
\vspace{-0.2cm}
    \small
    \begin{tabular}{ll}
    \toprule
    \textbf{Parameter}           & \textbf{Values}                          \\
    \midrule
    max\_depth      & 2, 4, 6, 8, 10, 12     \\
    lambda\_l1      & 1e-8, 1e-7, 1e-6, 1e-5, 1e-4, 1e-3, 1e-2, 1e-1, 1.  \\
    lambda\_l2      & 1e-8, 1e-7, 1e-6, 1e-5, 1e-4, 1e-3, 1e-2, 1e-1, 1.  \\
    eta  & 0.01, 0.03, 0.1, 0.3 \\
    \bottomrule
    \end{tabular}
    \label{table:hyperparamater_xgboost}
\end{table}

\section{RUNTIME ESTIMATES FOR TABLLM}
\label{sec:runtime_estimates}

The TabLLM training time on the \textbf{Income} dataset for 64 training examples and 30 epochs with a batch size of 8 was less than 3 minutes.
The average inference time for the test set of 10,000 examples with a batch size of 16 was 2 minutes, around 12 ms per example.
The training and inference times for the other public datasets were comparable.
Due to the larger size of the healthcare claims dataset, it took nearly 4 minutes to train for 64 examples and 10 epochs for \textbf{EoL} and was similar for the other two tasks.
Inference took approximately 14 minutes for 10,000 examples with a batch size of 16, i.e. around 84 ms per example.
The training times scaled linearly in the shot size.

\section{PARAMETER TUNING FOR BASELINES}
\label{sec:parameter_tuning}

We used the scikit-learn framework to perform cross-validation and parameter tuning for the LR and the tree-based models \citeponline{scikit-learn}. 
For LR we tried common parameters for the penalty term and regularization strength (see Table \ref{table:hyperparamater_lr}).
We used the same LR parameters for the public tabular datasets and the healthcare claims dataset.
For the tree-based models we adopted the hyperparameter ranges from \citetonline{borisov_deep_2022} and \citetonline{grinsztajn_why_2022}.
We discretized the parameter ranges and performed a complete grid search (see Tables \ref{table:hyperparamater_lightgbm} and \ref{table:hyperparamater_xgboost}).

For the neural baselines SAINT, TabNet, and NODE, we used the setup and suggested hyperparameter ranges in \citetonline{borisov_deep_2022}.
We modified the open-source implementation of these methods\footnote{https://github.com/kathrinse/TabSurvey} to support ingestion of the nine public tabular datasets.
We used the hyperparameter-tuning framework Optuna\footnote{https://github.com/optuna/optuna} and selected parameters that maximize AUC-ROC across folds.
Note that for the 4-shot setting of the \textbf{Car} dataset, AUC may not be defined if the selected validation set includes only one label; in this case we used accuracy as our validation metric but report AUC-ROC on the holdout test set.
Each neural baseline model was run for 20 trials with Optuna and trained for 100 epochs per hyperparameter settings.

\section{COMPARING BASELINE RESULTS TO THE LITERATURE}
\label{sec:comparing_baseline}

To assess whether our baseline results match results reported in the literature, we report studies that used the same models.

\paragraph{Bank Dataset.}
\citetonline{kadra2021well} trained a XGBoost, TabNet, and NODE baseline on this dataset and achieved a balanced accuracy of 72.7, 70.6, and 74.6.
Our experiments for a set of 512 balanced training examples (512 shots) show a better performance for XGBoost than NODE.

\paragraph{Blood Dataset.}
The XGBoost, TabNet, and NODE baselines trained in \citetonline{kadra2021well}  achieved a balanced accuracy of 62.3, 64.3, 50.
Our results for a set of 512 balanced training examples (512 shots) also show a better performance for TabNet than XGBoost.
However, in our experiments NODE performs better than XGBoost and not worse.

\paragraph{California Dataset.}
\citetonline{borisov_deep_2022} trained a Linear Model, XGBoost, LightGBM, TabNet, NODE, and SAINT baseline on a regression version of the dataset.
They achieved a mean squared error of 0.53, 0.21, 0.20, 0.35, 0.28, and 0.23.
Our experiments for a set of 512 balanced training examples (512 shots) show a better performance for XGBoost than LightGBM and the same performance for TabNet and NODE.
Also, our linear model performs much better which is probably due to more extensive hyperparameter tuning.

\paragraph{Car Dataset.}
The XGBoost, TabNet, and NODE models in \citetonline{kadra2021well} showed a balanced accuracy of 92.4, 98.7, and 46.1.
In our experiments, XGBoost and TabNet performed very similar for many training examples and NODE was only slightly inferior.

\paragraph{Credit-g Dataset.}
The XGBoost, TabNet, and NODE baselines trained in \citetonline{kadra2021well} achieved a balanced accuracy of 68.9, 61.2, and 73.1.
Our AUC results cannot easily be compared but our experiments for 512 balanced training examples (512 shots) follow the same trend.

\paragraph{Diabetes Dataset.}
\citetonline{hasan_diabetes_2020} reported an AUC of 0.828 (0.030) for XGBoost on the diabetes dataset, which matches our findings.
With additional feature selection and preprocessing methods they reached an AUC of 0.946 (0.020) with XGBoost, but this was out of the scope of our work.
XGBoost was the most performant model that they included in their experiments.

\paragraph{Heart Dataset.}
\citetonline{muhammad_early_2020} used only the 303 instances from the Cleveland cohort, while we combined all four sub-cohorts. 
They achieved an AUC of 0.923 with LR, which is close to our results on all sub-cohorts.
They also tested several models that outperformed LR.

\paragraph{Income Dataset.}
Many studies used the Income or Adult dataset.
The review \citetonline{borisov_deep_2022} included several of our baselines.
They reported an AUC of 0.854 (0.002) for a linear model, 0.928 (0.001) for XGBoost, 0.928 (0.001) for LightGBM, 0.916 (0.002) for SAINT, 0.911 (0.001) for TabNet,  and  0.911 (0.002) for NODE.
These are in accordance with our results.
We reckon the better performance of our LR model is due to more extensive parameter tuning.

\paragraph{Jungle Dataset.}   
The XGBoost and TabNet  baselines trained in \citetonline{kadra2021well} achieved a balanced accuracy of 87.3 and 73.4.
They did not train a NODE moel for this dataset.
The results follows the same trend as our experiments for a set of 512 balanced training examples (512 shots).

\section{ADJUSTING INCOME DATASET FOR INFLATION}
\label{sec:adjusting_income}

We wanted to investigate how a distribution shift caused by inflation affects the zero-shot performance of TabLLM.
The \textbf{Income} dataset was collected in 1994, and the label and two features (capital gain/loss in last year) contain dollar values.
T0 was trained in 2021 \citeponline{sanh_multitask_2021}, and we assumed that the training data is much more recent than the \textbf{Income} dataset.
The inflation rate from 1994 to 2021 is 1.79\footnote{U.S. Bureau of Labor Statistics, CPI Inflation Calculator:  https://www.bls.gov/data/inflation\_calculator.htm}.
Without inflation correction the zero-shot results were 0.80 (0.01).
Correcting the two features, correcting only the prompt, and correcting both all yielded the same performance as the uncorrected one.
The accuracy values also remained the same with the inflation correction.

\section{FEATURE IMPORTANCE ANALYSIS OF TABLLM}
\label{sec:feature_importance}

We wanted to understand which features were most important for the zero-shot performance of TabLLM on \textbf{Income} and \textbf{EoL}.
To this end, we used zero-shot TabLLM with the \textit{List Template} serialization to predict the label probability of all examples in the dataset.
We then used 4-fold cross validation to fit a L2-regularized LR model to the predicted label using the features in the serialization as covariates.
For \textbf{EoL}, we used age, sex, race, and the conditions as inputs, which summed up to 14,105 features.

For \textbf{Income} we compared these approximated importance scores to the feature coefficients of a LR model trained on all data for a single seed (Table \ref{table:feature_importance_full}).
We used the same setup for the LR model as for our main experiments.
We did 4-fold cross validation on an $80\%$ training split to choose hyperparameters, and then refit the model using all training data.
The best parameters of the LR model for \textbf{Income} were a `l1' penalty and a regularization constant of $1$.
For \textbf{EoL}, we decided that the LR model coefficients did not provide a good estimate of the ground truth due to the vast amount of features and possible collinearities in the data.
Instead, we provide the relative risk (RR) with 95\% confidence intervals (CI) treating the occurrence of a feature as an intervention.
We report the 50 most and least important features of TabLLM in Table \ref{table:feature_importance_ibc_full}.

\section{EFFECT OF USING DIFFERENT PROMPTS}
\label{sec:effect_prompts}

\begin{table}[]
\caption{The mean performance for one prompt (ours, SD over five seed omitted) and the mean performance and SD across five different prompts (each again over five seeds).}
\vspace{-0.25cm}
\resizebox{\columnwidth}{!}{
\begin{tabular}{lccccccccc}
\toprule
\textbf{Dataset}                                & \textbf{Bank} & \textbf{Blood} & \textbf{California} & \textbf{Car} & \textbf{Credit-g} & \textbf{Diabetes} & \textbf{Heart} & \textbf{Income} & \textbf{Jungle} \\ \midrule
TabLLM 0-shot: 1 prompt (ours)    & $0.63_{\phantom{.00}}$     & $0.61_{\phantom{.00}}$      & $0.61_{\phantom{.00}}$           & $0.81_{\phantom{.00}}$    & $0.53_{\phantom{.00}}$       & $0.68_{\phantom{.00}}$         & $0.54_{\phantom{.00}}$      & $0.84_{\phantom{.00}}$       & $0.60_{\phantom{.00}}$       \\ \midrule
TabLLM 0-shot: avg. 5 prompts & $0.64_{.01}$     & $0.60_{.02}$      & $0.59_{.01}$           & $0.80_{.01}$    & $0.52_{.01}$       & $0.67_{.01}$         & $0.55_{.04}$      & $0.84_{.01}$       & $0.60_{.00}$         \\ \bottomrule 
\end{tabular}}
\label{tab:effect_prompts}
\end{table}

To evaluate the effect of using a different prompt we considered the zero-shot setting, since even few training examples mostly cancel the effect.
For all datasets we constructed five different prompts that contained the same question, e.g., ``Does this person earn a lot of money?'' instead of ``Does this person earn more than 50000 dollars per year?'' for the \textbf{Income} dataset.
The results are summarized in Table \ref{tab:effect_prompts}.
The effects were relative small ranging from a standard deviation of 0.00 for \textbf{Jungle} to 0.04 for \textbf{Heart} across the five prompts.
This suggests that TabLLM is not very sensitive to using different prompts.

\newpage

\begin{table*}[h]
\caption{Test AUC performance of competing methods on public tabular datasets. Each column reports the $k$-shot performance for different values of $k$. Standard deviations across five random seeds are shown as subscripts.}
\vspace{-0.4cm}
\centering
\renewcommand{\arraystretch}{.95}
\setlength{\tabcolsep}{9.5pt}
\resizebox{\textwidth}{!}{
\begin{threeparttable}
\begin{tabular}{llcccccccccc}
\toprule
& \multicolumn{10}{c}{\textbf{Number of Shots}}\\
\addlinespace[1.5mm]
\textbf{Method} & \textbf{0} & \textbf{4} & \textbf{8} & \textbf{16} & \textbf{32} & \textbf{64} & \textbf{128} & \textbf{256} & \textbf{512} & \textbf{all}\\
\midrule
\midrule
\multicolumn{11}{l}{\textbf{Bank Dataset}}\\
\midrule
Logistic regression            & ---             & $0.55_{.09}$   & $0.66_{.09}$   & $0.75_{.06}$   & $0.81_{.02}$   & $0.84_{.02}$   & $0.86_{.02}$   & $0.88_{.01}$   & $0.89_{.00}$   & $0.91_{.00}$  \\
Logistic regression (ordinal)  & ---             & $0.51_{.02}$   & $0.60_{.12}$   & $0.68_{.09}$   & $0.78_{.04}$   & $0.82_{.01}$   & $0.84_{.03}$   & $0.86_{.01}$   & $0.87_{.00}$   & $0.88_{.00}$    \\
LightGBM                       & ---             & $0.50_{.00}$   & $0.50_{.00}$   & $0.50_{.00}$   & $0.50_{.00}$   & $0.77_{.03}$   & $0.84_{.03}$   & $0.88_{.01}$   & $0.89_{.00}$   & $0.94_{.00}$  \\
LightGBM (ordinal)             & ---             & $0.50_{.00}$   & $0.50_{.00}$   & $0.50_{.00}$   & $0.50_{.00}$   & $0.78_{.03}$   & $0.84_{.02}$   & $0.87_{.01}$   & $0.89_{.00}$   & $0.94_{.00}$  \\
XGBoost                        & ---             & $0.50_{.00}$   & $0.56_{.09}$   & $0.68_{.04}$   & $0.76_{.03}$   & $0.83_{.02}$   & $0.85_{.03}$   & $0.88_{.01}$   & $0.90_{.01}$   & $0.94_{.00}$  \\
XGBoost (ordinal)              & ---             & $0.50_{.00}$   & $0.56_{.09}$   & $0.69_{.05}$   & $0.75_{.04}$   & $0.82_{.02}$   & $0.84_{.03}$   & $0.87_{.01}$   & $0.89_{.00}$   & $0.93_{.00}$  \\
\midrule
SAINT                          & ---             & $0.51_{.10}$   & $0.61_{.11}$   & $0.70_{.04}$   & $0.77_{.03}$   & $0.81_{.03}$   & $0.85_{.02}$   & $0.88_{.01}$   & $0.88_{.01}$   & $0.93_{.00}$  \\
TabNet                         & ---             & $0.51_{.06}$   & $0.58_{.05}$   & $0.64_{.10}$   & $0.62_{.04}$   & $0.71_{.06}$   & $0.73_{.03}$   & $0.80_{.04}$   & $0.83_{.03}$   & $0.93_{.00}$  \\
NODE                           & ---             & $0.52_{.02}$   & $0.55_{.06}$   & $0.64_{.06}$   & $0.73_{.06}$   & $0.78_{.02}$   & $0.83_{.03}$   & $0.85_{.01}$   & $0.86_{.01}$   & $0.76_{.02}$  \\
TabPFN                         & ---             & $0.59_{.14}$   & $0.66_{.08}$   & $0.69_{.02}$   & $0.76_{.03}$   & $0.82_{.03}$   & $0.86_{.02}$   & $0.89_{.00}$   & $0.90_{.00}$   & $0.91_{.00}$  \\
TabPFN (ordinal)               & ---             & $0.57_{.10}$   & $0.67_{.05}$   & $0.71_{.05}$   & $0.78_{.04}$   & $0.83_{.01}$   & $0.86_{.02}$   & $0.87_{.00}$   & $0.88_{.00}$   & $0.89_{.00}$   \\
\midrule
TabLLM (T0  + Text GPT-3)      & $0.63_{.01}$    & $0.61_{.04}$    & $0.62_{.02}$    & $0.63_{.03}$    & $0.64_{.02}$    & $0.66_{.04}$    & $0.76_{.04}$    & $0.81_{.02}$    & $0.82_{.01}$    & *  \\
TabLLM (T0 + Text T0)          & $0.54_{.01}$    & $0.56_{.08}$    & $0.60_{.06}$    & $0.59_{.06}$    & $0.60_{.04}$    & $0.62_{.04}$    & $0.67_{.04}$    & $0.79_{.03}$    & $0.85_{.01}$    & *  \\
TabLLM (T0 + Table-To-Text)    & $0.42_{.01}$    & $0.48_{.07}$    & $0.50_{.05}$    & $0.56_{.03}$    & $0.57_{.04}$    & $0.59_{.05}$    & $0.63_{.03}$    & $0.68_{.02}$    & $0.74_{.01}$    & *  \\
TabLLM (T0 + Text Template)    & $0.63_{.01}$    & $0.59_{.10}$    & $0.64_{.05}$    & $0.65_{.05}$    & $0.64_{.06}$    & $0.69_{.03}$    & $0.82_{.05}$    & $0.87_{.01}$    & $0.88_{.01}$    & $0.92_{\phantom{..}}$\dag\\
TabLLM (T0 + List Template)    & $0.60_{.01}$    & $0.59_{.10}$    & $0.66_{.02}$    & $0.65_{.04}$    & $0.66_{.05}$    & $0.74_{.07}$    & $0.85_{.02}$    & $0.87_{.01}$    & $0.87_{.01}$    & *  \\
TabLLM (T0 + List Only Values) & $0.56_{.01}$    & $0.58_{.09}$    & $0.60_{.04}$    & $0.63_{.03}$    & $0.67_{.03}$    & $0.71_{.05}$    & $0.79_{.03}$    & $0.84_{.01}$    & $0.86_{.01}$    & *  \\
TabLLM (T0 + List Perm. Names) & $0.64_{.00}$    & $0.55_{.10}$    & $0.62_{.07}$    & $0.63_{.04}$    & $0.63_{.05}$    & $0.68_{.04}$    & $0.82_{.02}$    & $0.86_{.01}$    & $0.88_{.00}$    & *  \\
TabLLM (T0 + List Perm. Values)& $0.38_{.01}$    & $0.47_{.11}$    & $0.53_{.06}$    & $0.55_{.07}$    & $0.57_{.05}$    & $0.65_{.04}$    & $0.75_{.07}$    & $0.84_{.01}$    & $0.85_{.01}$    & *  \\
TabLLM (T0 3B + Text Template) & $0.61_{.01}$    & $0.60_{.10}$    & $0.65_{.05}$    & $0.64_{.07}$    & $0.65_{.05}$    & $0.70_{.02}$    & $0.77_{.05}$    & $0.88_{.01}$    & $0.89_{.01}$    & *  \\
\midrule
\midrule
\multicolumn{11}{l}{\textbf{Blood Dataset}}\\
\midrule
Logistic regression            & ---             & $0.54_{.09}$   & $0.59_{.08}$   & $0.72_{.03}$   & $0.70_{.06}$   & $0.74_{.02}$   & $0.76_{.02}$   & $0.76_{.02}$   & $0.76_{.03}$   & $0.76_{.03}$  \\
Logistic regression (ordinal)  & ---             & $0.54_{.09}$   & $0.59_{.08}$   & $0.72_{.03}$   & $0.70_{.06}$   & $0.74_{.02}$   & $0.76_{.02}$   & $0.76_{.02}$   & $0.76_{.03}$   & $0.76_{.03}$   \\
LightGBM                       & ---             & $0.50_{.00}$   & $0.50_{.00}$   & $0.50_{.00}$   & $0.50_{.00}$   & $0.69_{.04}$   & $0.71_{.05}$   & $0.71_{.07}$   & $0.67_{.05}$   & $0.74_{.04}$  \\
LightGBM (ordinal)             & ---             & $0.50_{.00}$   & $0.50_{.00}$   & $0.50_{.00}$   & $0.50_{.00}$   & $0.69_{.04}$   & $0.71_{.05}$   & $0.71_{.07}$   & $0.67_{.05}$   & $0.74_{.04}$  \\
XGBoost                        & ---             & $0.50_{.00}$   & $0.58_{.07}$   & $0.66_{.04}$   & $0.67_{.06}$   & $0.68_{.05}$   & $0.71_{.06}$   & $0.70_{.07}$   & $0.67_{.06}$   & $0.71_{.04}$  \\
XGBoost (ordinal)              & ---             & $0.50_{.00}$   & $0.58_{.07}$   & $0.66_{.04}$   & $0.67_{.06}$   & $0.68_{.05}$   & $0.71_{.06}$   & $0.70_{.07}$   & $0.67_{.06}$   & $0.71_{.04}$  \\
\midrule
SAINT                          & ---             & $0.47_{.12}$   & $0.66_{.08}$   & $0.66_{.03}$   & $0.67_{.06}$   & $0.67_{.05}$   & $0.71_{.03}$   & $0.76_{.05}$   & $0.73_{.02}$   & $0.74_{.03}$  \\
TabNet                         & ---             & $0.47_{.09}$   & $0.61_{.06}$   & $0.60_{.09}$   & $0.66_{.06}$   & $0.63_{.06}$   & $0.66_{.04}$   & $0.72_{.06}$   & $0.72_{.02}$   & $0.71_{.03}$  \\
NODE                           & ---             & $0.49_{.04}$   & $0.60_{.07}$   & $0.62_{.04}$   & $0.67_{.03}$   & $0.71_{.05}$   & $0.76_{.03}$   & $0.74_{.03}$   & $0.76_{.03}$   & $0.74_{.03}$  \\
TabPFN                         & ---             & $0.52_{.08}$   & $0.64_{.04}$   & $0.67_{.01}$   & $0.70_{.04}$   & $0.73_{.04}$   & $0.75_{.04}$   & $0.76_{.04}$   & $0.76_{.03}$   & $0.74_{.03}$  \\
TabPFN (ordinal)               & ---             & $0.52_{.08}$   & $0.64_{.04}$   & $0.67_{.01}$   & $0.70_{.04}$   & $0.73_{.04}$   & $0.75_{.04}$   & $0.76_{.04}$   & $0.76_{.03}$   & $0.74_{.03}$  \\
\midrule
TabLLM (T0  + Text GPT-3)      & $0.63_{.04}$    & $0.61_{.07}$    & $0.65_{.04}$    & $0.63_{.02}$    & $0.64_{.03}$    & $0.62_{.05}$    & $0.67_{.06}$    & $0.68_{.05}$    & $0.66_{.05}$    & *  \\
TabLLM (T0 + Text T0)          & $0.49_{.04}$    & $0.51_{.03}$    & $0.59_{.08}$    & $0.59_{.06}$    & $0.64_{.04}$    & $0.65_{.06}$    & $0.66_{.05}$    & $0.68_{.06}$    & $0.66_{.03}$    & *  \\
TabLLM (T0 + Table-To-Text)    & $0.61_{.04}$    & $0.59_{.04}$    & $0.59_{.03}$    & $0.57_{.03}$    & $0.62_{.07}$    & $0.56_{.07}$    & $0.57_{.07}$    & $0.64_{.07}$    & $0.61_{.05}$    & *  \\
TabLLM (T0 + Text Template)    & $0.61_{.04}$    & $0.58_{.09}$    & $0.66_{.03}$    & $0.66_{.07}$    & $0.68_{.04}$    & $0.68_{.04}$    & $0.68_{.06}$    & $0.70_{.08}$    & $0.68_{.04}$    & $0.70_{.04}$  \\
TabLLM (T0 + List Template)    & $0.56_{.05}$    & $0.54_{.08}$    & $0.64_{.02}$    & $0.64_{.08}$    & $0.67_{.05}$    & $0.66_{.06}$    & $0.67_{.05}$    & $0.70_{.06}$    & $0.67_{.06}$    & *  \\
TabLLM (T0 + List Only Values) & $0.45_{.05}$    & $0.49_{.07}$    & $0.57_{.03}$    & $0.57_{.06}$    & $0.62_{.06}$    & $0.61_{.04}$    & $0.64_{.04}$    & $0.68_{.07}$    & $0.67_{.05}$    & *  \\
TabLLM (T0 + List Perm. Names) & $0.52_{.04}$    & $0.49_{.07}$    & $0.62_{.03}$    & $0.62_{.06}$    & $0.65_{.05}$    & $0.65_{.04}$    & $0.68_{.06}$    & $0.72_{.06}$    & $0.68_{.04}$    & *  \\
TabLLM (T0 + List Perm. Values)& $0.51_{.03}$    & $0.51_{.06}$    & $0.54_{.04}$    & $0.52_{.07}$    & $0.55_{.03}$    & $0.59_{.06}$    & $0.59_{.02}$    & $0.62_{.06}$    & $0.62_{.05}$    & *  \\
TabLLM (T0 3B + Text Template) & $0.42_{.05}$    & $0.47_{.04}$    & $0.62_{.04}$    & $0.62_{.09}$    & $0.65_{.07}$    & $0.67_{.04}$    & $0.69_{.04}$    & $0.71_{.06}$    & $0.67_{.04}$    & *  \\
\midrule
\midrule
\multicolumn{11}{l}{\textbf{California Dataset}}\\
\midrule
Logistic regression            & ---             & $0.58_{.11}$   & $0.69_{.13}$   & $0.80_{.06}$   & $0.84_{.03}$   & $0.88_{.01}$   & $0.90_{.00}$   & $0.91_{.00}$   & $0.91_{.00}$   & $0.92_{.00}$  \\
Logistic regression (ordinal)  & ---             & $0.58_{.11}$   & $0.69_{.13}$   & $0.80_{.06}$   & $0.84_{.03}$   & $0.88_{.01}$   & $0.90_{.00}$   & $0.91_{.00}$   & $0.91_{.00}$   & $0.92_{.00}$   \\
LightGBM                       & ---             & $0.50_{.00}$   & $0.50_{.00}$   & $0.50_{.00}$   & $0.50_{.00}$   & $0.81_{.02}$   & $0.87_{.01}$   & $0.90_{.01}$   & $0.92_{.00}$   & $0.97_{.00}$  \\
LightGBM (ordinal)             & ---             & $0.50_{.00}$   & $0.50_{.00}$   & $0.50_{.00}$   & $0.50_{.00}$   & $0.81_{.02}$   & $0.87_{.01}$   & $0.90_{.01}$   & $0.92_{.00}$   & $0.97_{.00}$  \\
XGBoost                        & ---             & $0.50_{.00}$   & $0.62_{.10}$   & $0.74_{.03}$   & $0.79_{.04}$   & $0.82_{.04}$   & $0.87_{.01}$   & $0.90_{.01}$   & $0.92_{.01}$   & $0.97_{.00}$  \\
XGBoost (ordinal)              & ---             & $0.50_{.00}$   & $0.62_{.10}$   & $0.74_{.03}$   & $0.79_{.04}$   & $0.82_{.04}$   & $0.87_{.01}$   & $0.90_{.01}$   & $0.92_{.01}$   & $0.97_{.00}$  \\
\midrule
SAINT                          & ---             & $0.59_{.09}$   & $0.64_{.12}$   & $0.73_{.06}$   & $0.76_{.06}$   & $0.81_{.02}$   & $0.84_{.01}$   & $0.88_{.02}$   & $0.91_{.02}$   & $0.95_{.00}$  \\
TabNet                         & ---             & $0.50_{.08}$   & $0.57_{.06}$   & $0.67_{.02}$   & $0.69_{.05}$   & $0.72_{.03}$   & $0.79_{.02}$   & $0.84_{.02}$   & $0.87_{.01}$   & $0.96_{.00}$  \\
NODE                           & ---             & $0.58_{.06}$   & $0.57_{.07}$   & $0.70_{.05}$   & $0.77_{.03}$   & $0.80_{.01}$   & $0.86_{.02}$   & $0.86_{.02}$   & $0.87_{.01}$ & $0.87_{.01}$ \\
TabPFN                         & ---             & $0.63_{.13}$   & $0.63_{.11}$   & $0.80_{.03}$   & $0.85_{.03}$   & $0.89_{.01}$   & $0.91_{.01}$   & $0.92_{.00}$   & $0.93_{.00}$   & $0.94_{.00}$  \\
TabPFN (ordinal)               & ---             & $0.63_{.13}$   & $0.63_{.11}$   & $0.80_{.03}$   & $0.85_{.03}$   & $0.89_{.01}$   & $0.91_{.01}$   & $0.92_{.00}$   & $0.93_{.00}$   & $0.94_{.00}$  \\
\midrule
TabLLM (T0  + Text GPT-3)      & $0.56_{.00}$    & $0.55_{.03}$    & $0.57_{.05}$    & $0.61_{.06}$    & $0.73_{.05}$    & $0.73_{.04}$    & $0.82_{.01}$    & $0.84_{.01}$    & $0.85_{.01}$    & *  \\
TabLLM (T0 + Text T0)          & $0.49_{.01}$    & $0.52_{.02}$    & $0.51_{.02}$    & $0.52_{.02}$    & $0.54_{.04}$    & $0.56_{.04}$    & $0.69_{.02}$    & $0.73_{.03}$    & $0.80_{.02}$    & *  \\
TabLLM (T0 + Table-To-Text)    & $0.49_{.01}$    & $0.50_{.01}$    & $0.51_{.01}$    & $0.52_{.02}$    & $0.57_{.04}$    & $0.58_{.04}$    & $0.74_{.03}$    & $0.79_{.02}$    & $0.82_{.01}$    & *  \\
TabLLM (T0 + Text Template)    & $0.61_{.01}$    & $0.63_{.05}$    & $0.60_{.07}$    & $0.70_{.08}$    & $0.77_{.08}$    & $0.77_{.04}$    & $0.81_{.02}$    & $0.83_{.01}$    & $0.86_{.02}$    & $0.95_{.00}$  \\
TabLLM (T0 + List Template)    & $0.61_{.01}$    & $0.64_{.05}$    & $0.62_{.06}$    & $0.68_{.07}$    & $0.77_{.07}$    & $0.79_{.02}$    & $0.82_{.02}$    & $0.84_{.01}$    & $0.87_{.01}$    & *  \\
TabLLM (T0 + List Only Values) & $0.58_{.01}$    & $0.57_{.08}$    & $0.55_{.03}$    & $0.65_{.09}$    & $0.74_{.08}$    & $0.77_{.03}$    & $0.83_{.01}$    & $0.84_{.02}$    & $0.86_{.02}$    & *  \\
TabLLM (T0 + List Perm. Names) & $0.54_{.01}$    & $0.52_{.03}$    & $0.52_{.04}$    & $0.52_{.03}$    & $0.66_{.06}$    & $0.74_{.01}$    & $0.81_{.02}$    & $0.84_{.02}$    & $0.86_{.02}$    & *  \\
TabLLM (T0 + List Perm. Values)& $0.47_{.01}$    & $0.48_{.02}$    & $0.50_{.01}$    & $0.52_{.02}$    & $0.57_{.03}$    & $0.64_{.04}$    & $0.71_{.04}$    & $0.76_{.01}$    & $0.78_{.02}$    & *  \\
TabLLM (T0 3B + Text Template) & $0.57_{.01}$    & $0.59_{.03}$    & $0.57_{.04}$    & $0.66_{.07}$    & $0.77_{.06}$    & $0.79_{.02}$    & $0.81_{.01}$    & $0.83_{.01}$    & $0.85_{.01}$    & *  \\
\bottomrule
\end{tabular}
\begin{tablenotes}[flushleft]
    \footnotesize 
    \item[*] Result omitted due to runtime limitations of TabLLM on the full dataset.
    \item[\dag] Only a single run performed due to runtime limitations of TabLLM on the full dataset.
\end{tablenotes}
\end{threeparttable}
}
\label{table:results_public_dataset_full1}
\end{table*}

\begin{table*}[h]
\caption{Test AUC performance of competing methods on public tabular datasets. Each column reports the $k$-shot performance for different values of $k$. Standard deviations across five random seeds are shown as subscripts.}
\vspace{-0.4cm}
\centering
\renewcommand{\arraystretch}{.95}
\setlength{\tabcolsep}{9.5pt}
\resizebox{\textwidth}{!}{
\begin{threeparttable}
\begin{tabular}{llcccccccccc}
\toprule
& \multicolumn{10}{c}{\textbf{Number of Shots}}\\
\addlinespace[1.5mm]
\textbf{Method} & \textbf{0} & \textbf{4} & \textbf{8} & \textbf{16} & \textbf{32} & \textbf{64} & \textbf{128} & \textbf{256} & \textbf{512} & \textbf{all}\\
\midrule
\midrule
\multicolumn{11}{l}{\textbf{Car Dataset}}\\
\midrule
Logistic regression            & ---            & $0.61_{.02}$   & $0.65_{.10}$   & $0.74_{.07}$   & $0.83_{.02}$   & $0.93_{.02}$   & $0.96_{.01}$   & $0.97_{.01}$   & $0.98_{.00}$   & $0.98_{.00}$  \\
Logistic regression (ordinal)  & ---            & $0.62_{.06}$   & $0.63_{.05}$   & $0.64_{.07}$   & $0.75_{.04}$   & $0.73_{.03}$   & $0.73_{.03}$   & $0.74_{.03}$   & $0.76_{.02}$   & $0.78_{.03}$  \\
LightGBM                       & ---            & $0.50_{.00}$   & $0.50_{.00}$   & $0.50_{.00}$   & $0.50_{.00}$   & $0.85_{.06}$   & $0.93_{.01}$   & $0.98_{.01}$   & $0.99_{.01}$   & $1.00_{.00}$            \\
LightGBM (ordinal)             & ---             & $0.50_{.00}$   & $0.50_{.00}$   & $0.50_{.00}$   & $0.50_{.00}$   & $0.75_{.04}$   & $0.91_{.05}$   & $0.98_{.01}$   & $0.99_{.00}$   & $1.00_{.00}$  \\
XGBoost                        & ---            & $0.50_{.00}$   & $0.59_{.04}$   & $0.70_{.08}$   & $0.82_{.03}$   & $0.91_{.02}$   & $0.95_{.01}$   & $0.98_{.01}$   & $0.99_{.01}$   & $1.00_{.00}$            \\
XGBoost (ordinal)              & ---             & $0.50_{.00}$   & $0.55_{.03}$   & $0.70_{.04}$   & $0.78_{.03}$   & $0.90_{.03}$   & $0.94_{.01}$   & $0.98_{.01}$   & $0.99_{.01}$   & $1.00_{.00}$  \\
\midrule
SAINT & --- & $0.56_{.08}$ & $0.64_{.08}$ & $0.76_{.03}$ & $0.85_{.03}$ & $0.92_{.02}$ & $0.96_{.01}$ & $0.98_{.01}$ & $0.99_{.00}$ & $1.00_{.00}$ \\
TabNet                         & ---  &  \dag & $0.54_{.05}$   & $0.64_{.05}$   & $0.66_{.05}$   & $0.73_{.07}$   & $0.81_{.04}$   & $0.93_{.02}$   & $0.98_{.01}$ & $1.00_{.00}$   \\
NODE                           & ---   &  $0.51_{.10}$   & $0.57_{.06}$  & $0.69_{.02}$   & $0.74_{.03}$    & $0.80_{.02}$    & $0.82_{.01}$   & $0.91_{.01}$ & $0.96_{.01}$ & $0.93_{.01}$  \\
TabPFN                         & ---             & $0.64_{.06}$   & $0.75_{.05}$   & $0.87_{.04}$   & $0.92_{.02}$   & $0.97_{.00}$   & $0.99_{.01}$   & $1.00_{.00}$   & $1.00_{.00}$   & $1.00_{.00}$  \\
TabPFN (ordinal)               & ---             & $0.59_{.06}$   & $0.65_{.08}$   & $0.75_{.04}$   & $0.82_{.06}$   & $0.89_{.01}$   & $0.93_{.01}$   & $0.98_{.01}$   & $0.99_{.01}$   & $1.00_{.00}$     \\
\midrule
TabLLM (T0 + Text GPT-3)       & $0.72_{.02}$    & $0.75_{.03}$    & $0.75_{.02}$    & $0.78_{.01}$    & $0.83_{.01}$    & $0.87_{.02}$    & $0.90_{.01}$    & $0.93_{.02}$    & $0.93_{.02}$    & $0.96_{.01}$            \\
TabLLM (T0 + Text T0)          & $0.85_{.01}$    & $0.85_{.02}$    & $0.84_{.03}$    & $0.86_{.02}$    & $0.89_{.02}$    & $0.92_{.02}$    & $0.94_{.01}$    & $0.98_{.01}$    & $0.99_{.00}$    & $1.00_{.00}$            \\
TabLLM (T0 + Table-To-Text)    & $0.61_{.01}$    & $0.69_{.04}$    & $0.74_{.04}$    & $0.79_{.02}$    & $0.88_{.01}$    & $0.91_{.02}$    & $0.94_{.01}$    & $0.96_{.01}$    & $0.95_{.01}$    & $0.96_{.00}$            \\
TabLLM (T0 + Text Template)             & $0.82_{.02}$    & $0.83_{.03}$    & $0.85_{.03}$    & $0.86_{.03}$    & $0.91_{.02}$    & $0.96_{.02}$    & $0.98_{.01}$    & $0.99_{.00}$    & $1.00_{.00}$    & $1.00_{.00}$            \\
TabLLM (T0 + List Template)             & $0.79_{.02}$   & $0.84_{.03}$   & $0.85_{.02}$   & $0.86_{.03}$   & $0.91_{.02}$   & $0.95_{.01}$   & $0.98_{.01}$   & $0.99_{.00}$   & $1.00_{.00}$   & $1.00_{.00}$ \\
TabLLM (T0 + List Only Values) & $0.48_{.03}$   & $0.62_{.04}$   & $0.67_{.03}$   & $0.70_{.03}$   & $0.75_{.02}$   & $0.87_{.02}$   & $0.94_{.01}$   & $0.98_{.01}$   & $0.99_{.01}$   & $1.00_{.00}$ \\
TabLLM (T0 + List Perm. Names) & $0.39_{.02}$   & $0.54_{.10}$   & $0.58_{.06}$   & $0.70_{.03}$   & $0.86_{.02}$   & $0.94_{.01}$   & $0.97_{.02}$   & $0.99_{.01}$   & $0.99_{.00}$   & $1.00_{.00}$ \\
TabLLM (T0 + List Perm. Values)& $0.38_{.02}$    & $0.48_{.08}$    & $0.55_{.05}$    & $0.63_{.04}$    & $0.69_{.03}$    & $0.78_{.02}$    &  $0.90_{.03}$   & $0.98_{.01}$    & $1.00_{.00}$    & $1.00_{.00}$ \\
TabLLM (T0 3B + Text Template)          & $0.78_{.02}$    & $0.80_{.03}$    & $0.84_{.03}$    & $0.84_{.04}$    & $0.89_{.03}$    & $0.91_{.01}$    & $0.96_{.01}$    & $0.98_{.01}$    & $0.99_{.00}$    & $1.00_{.00}$            \\
\midrule
\midrule
\multicolumn{11}{l}{\textbf{Credit-g Dataset}}\\
\midrule
Logistic regression            & ---             & $0.50_{.08}$   & $0.56_{.06}$   & $0.58_{.08}$   & $0.68_{.08}$   & $0.66_{.07}$   & $0.71_{.06}$   & $0.75_{.04}$   & $0.76_{.02}$   & $0.79_{.03}$  \\
Logistic regression (ordinal)  & ---             & $0.56_{.05}$   & $0.54_{.06}$   & $0.55_{.05}$   & $0.61_{.05}$   & $0.68_{.05}$   & $0.66_{.03}$   & $0.68_{.04}$   & $0.71_{.02}$   & $0.72_{.02}$  \\
LightGBM                       & ---             & $0.50_{.00}$   & $0.50_{.00}$   & $0.50_{.00}$   & $0.50_{.00}$   & $0.61_{.09}$   & $0.68_{.03}$   & $0.72_{.02}$   & $0.75_{.02}$   & $0.78_{.02}$  \\
LightGBM (ordinal)             & ---             & $0.50_{.00}$   & $0.50_{.00}$   & $0.50_{.00}$   & $0.50_{.00}$   & $0.68_{.07}$   & $0.66_{.04}$   & $0.72_{.02}$   & $0.75_{.03}$   & $0.76_{.04}$  \\
XGBoost                        & ---             & $0.50_{.00}$   & $0.51_{.07}$   & $0.59_{.05}$   & $0.66_{.03}$   & $0.67_{.06}$   & $0.68_{.02}$   & $0.73_{.02}$   & $0.75_{.03}$   & $0.78_{.04}$  \\
XGBoost (ordinal)              & ---             & $0.50_{.00}$   & $0.54_{.11}$   & $0.57_{.08}$   & $0.64_{.05}$   & $0.66_{.06}$   & $0.68_{.04}$   & $0.74_{.02}$   & $0.76_{.03}$   & $0.76_{.04}$  \\
\midrule
SAINT                          & ---             & $0.56_{.08}$   & $0.53_{.05}$   & $0.60_{.05}$   & $0.66_{.06}$   & $0.66_{.06}$   & $0.68_{.05}$   & $0.72_{.04}$   & $0.73_{.03}$   & $0.77_{.04}$  \\
TabNet                         & ---             & $0.48_{.05}$   & $0.52_{.07}$   & $0.49_{.03}$   & $0.52_{.03}$   & $0.56_{.05}$   & $0.60_{.05}$   & $0.61_{.02}$   & $0.66_{.04}$   & $0.64_{.03}$  \\
NODE                           & ---             & $0.54_{.09}$   & $0.54_{.10}$   & $0.54_{.09}$   & $0.59_{.07}$   & $0.63_{.04}$   & $0.68_{.02}$   & $0.68_{.05}$   & $0.70_{.02}$   & $0.65_{.03}$  \\
TabPFN                         & ---             & $0.58_{.08}$   & $0.59_{.03}$   & $0.64_{.06}$   & $0.69_{.07}$   & $0.70_{.07}$   & $0.72_{.06}$   & $0.75_{.04}$   & $0.75_{.02}$   & $0.75_{.03}$  \\
TabPFN (ordinal)               & ---             & $0.55_{.08}$   & $0.51_{.07}$   & $0.57_{.06}$   & $0.62_{.03}$   & $0.66_{.05}$   & $0.70_{.02}$   & $0.73_{.01}$   & $0.73_{.03}$   & $0.75_{.04}$  \\
\midrule
TabLLM (T0  + Text GPT-3)      & $0.52_{.04}$   & $0.53_{.04}$   & $0.56_{.03}$   & $0.56_{.05}$   & $0.55_{.05}$   & $0.57_{.08}$   & $0.60_{.06}$   & $0.61_{.04}$   & $0.63_{.05}$   & *  \\
TabLLM (T0 + Text T0)          & $0.49_{.02}$   & $0.50_{.06}$   & $0.54_{.06}$   & $0.55_{.04}$   & $0.60_{.06}$   & $0.61_{.02}$   & $0.61_{.02}$   & $0.63_{.03}$   & $0.65_{.02}$   & *  \\
TabLLM (T0 + Table-To-Text)    & $0.50_{.06}$   & $0.65_{.04}$   & $0.60_{.05}$   & $0.60_{.07}$   & $0.65_{.05}$   & $0.67_{.05}$   & $0.65_{.05}$   & $0.68_{.04}$   & $0.64_{.05}$   & *  \\
TabLLM (T0 + Text Template)    & $0.53_{.05}$    & $0.69_{.04}$    & $0.66_{.04}$    & $0.66_{.05}$    & $0.72_{.06}$    & $0.70_{.07}$    & $0.71_{.07}$    & $0.72_{.03}$    & $0.72_{.02}$    & $0.70_{.02}$  \\
TabLLM (T0 + List Template)    & $0.53_{.05}$   & $0.64_{.04}$   & $0.60_{.06}$   & $0.64_{.05}$   & $0.70_{.05}$   & $0.66_{.08}$   & $0.67_{.03}$   & $0.70_{.03}$   & $0.70_{.04}$   & *  \\
TabLLM (T0 + List Only Values) & $0.66_{.06}$   & $0.71_{.03}$   & $0.67_{.06}$   & $0.69_{.06}$   & $0.72_{.06}$   & $0.69_{.05}$   & $0.69_{.07}$   & $0.70_{.06}$   & $0.68_{.04}$   & *  \\
TabLLM (T0 + List Perm. Names) & $0.44_{.01}$   & $0.58_{.09}$   & $0.59_{.08}$   & $0.60_{.07}$   & $0.70_{.06}$   & $0.69_{.06}$   & $0.67_{.05}$   & $0.70_{.05}$   & $0.70_{.03}$   & *  \\
TabLLM (T0 + List Perm. Values)& $0.50_{.05}$   & $0.55_{.06}$   & $0.56_{.07}$   & $0.58_{.04}$   & $0.64_{.03}$   & $0.66_{.08}$   & $0.67_{.09}$   & $0.68_{.03}$   & $0.69_{.03}$   & *  \\
TabLLM (T0 3B + Text Template) & $0.54_{.03}$   & $0.65_{.05}$   & $0.63_{.05}$   & $0.63_{.03}$   & $0.73_{.04}$   & $0.69_{.05}$   & $0.68_{.06}$   & $0.73_{.05}$   & $0.73_{.03}$   & *  \\
\midrule
\midrule
\multicolumn{11}{l}{\textbf{Diabetes Dataset}}\\
\midrule
Logistic regression            & ---            & $0.60_{.15}$   & $0.68_{.11}$   & $0.73_{.05}$   & $0.76_{.05}$   & $0.80_{.02}$   & $0.81_{.02}$   & $0.83_{.02}$   & $0.83_{.02}$   & $0.83_{.02}$   \\
Logistic regression (ordinal)  & ---             & $0.60_{.15}$   & $0.68_{.11}$   & $0.73_{.05}$   & $0.76_{.05}$   & $0.80_{.02}$   & $0.81_{.02}$   & $0.83_{.02}$   & $0.83_{.02}$   & $0.83_{.02}$  \\
LightGBM                       & ---            & $0.50_{.00}$   & $0.50_{.00}$   & $0.50_{.00}$   & $0.50_{.00}$   & $0.79_{.02}$   & $0.79_{.04}$   & $0.79_{.02}$   & $0.79_{.03}$   & $0.83_{.03}$           \\
LightGBM (ordinal)             & ---             & $0.50_{.00}$   & $0.50_{.00}$   & $0.50_{.00}$   & $0.50_{.00}$   & $0.79_{.02}$   & $0.79_{.04}$   & $0.79_{.02}$   & $0.79_{.03}$   & $0.83_{.03}$  \\
XGBoost                        & ---            & $0.50_{.00}$   & $0.59_{.16}$   & $0.72_{.07}$   & $0.69_{.08}$   & $0.73_{.05}$   & $0.78_{.05}$   & $0.80_{.03}$   & $0.80_{.01}$   & $0.84_{.03}$           \\
XGBoost (ordinal)              & ---             & $0.50_{.00}$   & $0.59_{.16}$   & $0.72_{.07}$   & $0.69_{.08}$   & $0.73_{.05}$   & $0.78_{.05}$   & $0.80_{.03}$   & $0.80_{.01}$   & $0.84_{.03}$  \\
\midrule
SAINT & --- & $0.46_{.12}$ & $0.65_{.11}$ & $0.73_{.06}$ & $0.73_{.06}$ & $0.79_{.03}$ & $0.81_{.03}$ & $0.81_{.04}$ & $0.77_{.03}$ & $0.83_{.03}$ \\
TabNet                         & ---              & $0.56_{.04}$    & $0.56_{.06}$   & $0.64_{.09}$   & $0.66_{.06}$    & $0.71_{.04}$   & $0.73_{.04}$   & $0.74_{.05}$   & $0.74_{.07}$ & $0.81_{.03}$ \\
NODE                           & ---    & $0.49_{.13}$   & $0.67_{.09}$   & $0.69_{.08}$   & $0.73_{.05}$   & $0.77_{.04}$    & $0.80_{.04}$   & $0.81_{.03}$    & $0.83_{.02}$ & $0.83_{.03}$    \\ 
TabPFN                         & ---             & $0.61_{.13}$   & $0.67_{.11}$   & $0.71_{.07}$   & $0.77_{.03}$   & $0.82_{.03}$   & $0.83_{.03}$   & $0.83_{.03}$   & $0.81_{.02}$   & $0.81_{.03}$  \\
TabPFN (ordinal)               & ---             & $0.61_{.13}$   & $0.67_{.11}$   & $0.71_{.07}$   & $0.77_{.03}$   & $0.82_{.03}$   & $0.83_{.03}$   & $0.83_{.03}$   & $0.81_{.02}$   & $0.81_{.03}$    \\
\midrule
TabLLM (T0 + Text GPT-3)       & $0.61_{.06}$    & $0.61_{.07}$    & $0.56_{.12}$    & $0.67_{.08}$    & $0.74_{.04}$    & $0.77_{.02}$    & $0.79_{.03}$    & $0.76_{.03}$    & $0.78_{.04}$    & $0.81_{.04}$            \\
TabLLM (T0 + Text T0)          & $0.58_{.04}$    & $0.53_{.05}$    & $0.53_{.06}$    & $0.54_{.09}$    & $0.59_{.05}$    & $0.68_{.02}$    & $0.73_{.04}$    & $0.72_{.05}$    & $0.72_{.03}$    & $0.76_{.01}$            \\
TabLLM (T0 + Table-To-Text)    & $0.58_{.04}$    & $0.51_{.10}$    & $0.53_{.07}$    & $0.56_{.05}$    & $0.57_{.04}$    & $0.59_{.04}$    & $0.72_{.05}$    & $0.74_{.04}$    & $0.75_{.06}$    & $0.77_{.04}$            \\
TabLLM (T0 + Text Template)             & $0.68_{.06}$    & $0.61_{.09}$    & $0.63_{.08}$    & $0.69_{.07}$    & $0.68_{.04}$    & $0.73_{.03}$    & $0.79_{.04}$    & $0.78_{.02}$    & $0.78_{.04}$    & $0.80_{.04}$            \\
TabLLM (T0 + List Template)             & $0.64_{.06}$    & $0.64_{.09}$    & $0.64_{.10}$    & $0.67_{.07}$    & $0.70_{.05}$    & $0.76_{.04}$    & $0.78_{.03}$    & $0.78_{.03}$    & $0.78_{.04}$    & $0.81_{.05}$            \\
TabLLM (T0 + List Only Values) & $0.55_{.05}$   & $0.54_{.07}$   & $0.52_{.05}$   & $0.59_{.08}$   & $0.63_{.04}$   & $0.67_{.07}$   & $0.73_{.03}$   & $0.75_{.06}$   & $0.77_{.04}$   & $0.79_{.03}$           \\
TabLLM (T0 + List Perm. Names) & $0.56_{.07}$    & $0.60_{.09}$    & $0.68_{.12}$    & $0.74_{.05}$    & $0.74_{.03}$    & $0.72_{.04}$    & $0.76_{.04}$    & $0.77_{.04}$    & $0.77_{.04}$    & $0.81_{.04}$            \\
TabLLM (T0 + List Perm. Values)& $0.44_{.03}$    & $0.47_{.09}$    & $0.43_{.06}$    & $0.55_{.07}$    & $0.61_{.05}$    & $0.65_{.05}$    & $0.73_{.03}$    & $0.76_{.03}$    & $0.78_{.02}$    & $0.80_{.03}$            \\
TabLLM (T0 3B + Text Template)          & $0.62_{.05}$    & $0.57_{.07}$    & $0.60_{.08}$    & $0.67_{.05}$    & $0.67_{.06}$    & $0.76_{.03}$    & $0.77_{.04}$    & $0.81_{.05}$    & $0.80_{.04}$    & $0.82_{.04}$            \\
\bottomrule
\end{tabular}
\begin{tablenotes}[flushleft]
    \footnotesize 
    \item[*] Result omitted due to runtime limitations of TabLLM on the full dataset.
    \item[\dag] Result omitted due to TabNet package not supporting unseen labels in validation set during cross validation.
\end{tablenotes}
\end{threeparttable}
}
\label{table:results_public_dataset_full2}
\end{table*}

\begin{table*}[h]
\caption{Test AUC performance of competing methods on public tabular datasets. Each column reports the $k$-shot performance for different values of $k$. Standard deviations across five random seeds are shown as subscripts.}
\vspace{-0.4cm}
\centering
\renewcommand{\arraystretch}{.95}
\setlength{\tabcolsep}{9.5pt}
\resizebox{\textwidth}{!}{
\begin{threeparttable}
\begin{tabular}{llcccccccccc}
\toprule
& \multicolumn{10}{c}{\textbf{Number of Shots}}\\
\addlinespace[1.5mm]
\textbf{Method} & \textbf{0} & \textbf{4} & \textbf{8} & \textbf{16} & \textbf{32} & \textbf{64} & \textbf{128} & \textbf{256} & \textbf{512} & \textbf{all}\\
\midrule
\midrule
\multicolumn{11}{l}{\textbf{Heart Dataset}}\\
\midrule
Logistic regression            & ---            & $0.69_{.17}$   & $0.75_{.13}$   & $0.82_{.06}$   & $0.87_{.05}$   & $0.91_{.01}$   & $0.90_{.02}$   & $0.92_{.01}$   & $0.93_{.01}$   & $0.93_{.01}$  \\
Logistic regression (ordinal)  & ---            & $0.70_{.17}$   & $0.73_{.14}$   & $0.84_{.04}$   & $0.88_{.03}$   & $0.89_{.01}$   & $0.88_{.02}$   & $0.90_{.02}$   & $0.92_{.02}$   & $0.92_{.02}$  \\
LightGBM                       & ---            & $0.50_{.00}$   & $0.50_{.00}$   & $0.50_{.00}$   & $0.50_{.00}$   & $0.91_{.01}$   & $0.91_{.01}$   & $0.91_{.01}$   & $0.93_{.00}$   & $0.94_{.01}$            \\
LightGBM (ordinal)             & ---             & $0.50_{.00}$   & $0.50_{.00}$   & $0.50_{.00}$   & $0.50_{.00}$   & $0.91_{.01}$   & $0.91_{.02}$   & $0.91_{.01}$   & $0.92_{.01}$   & $0.94_{.01}$  \\
XGBoost                        & ---            & $0.50_{.00}$   & $0.55_{.14}$   & $0.84_{.07}$   & $0.88_{.04}$   & $0.91_{.01}$   & $0.91_{.01}$   & $0.90_{.01}$   & $0.92_{.01}$   & $0.94_{.01}$            \\
XGBoost (ordinal)              & ---             & $0.50_{.00}$   & $0.56_{.15}$   & $0.84_{.07}$   & $0.90_{.03}$   & $0.91_{.01}$   & $0.90_{.01}$   & $0.90_{.01}$   & $0.92_{.01}$   & $0.94_{.01}$   \\
\midrule
SAINT & --- & $0.80_{.12}$ & $0.83_{.10}$ & $0.88_{.07}$ & $0.90_{.01}$ & $0.90_{.04}$ & $0.90_{.02}$ & $0.90_{.01}$ & $0.92_{.01}$ & $0.93_{.01}$ \\
TabNet                         & ---   & $0.56_{.12}$   & $0.70_{.05}$    & $0.73_{.14}$   & $0.80_{.04}$   & $0.83_{.05}$   & $0.84_{.03}$  & $0.88_{.02}$    & $0.88_{.03}$ & $0.89_{.03}$  \\
NODE                           & ---   & $0.52_{.10}$   & $0.78_{.08}$   & $0.83_{.03}$   & $0.86_{.02}$   & $0.88_{.02}$   & $0.88_{.01}$   & $0.91_{.02}$   & $0.92_{.03}$ & $0.92_{.03}$   \\
TabPFN                         & ---             & $0.84_{.06}$   & $0.88_{.05}$   & $0.87_{.06}$   & $0.91_{.02}$   & $0.92_{.02}$   & $0.92_{.02}$   & $0.92_{.01}$   & $0.92_{.02}$   & $0.92_{.02}$  \\
TabPFN (ordinal)               & ---             & $0.79_{.08}$   & $0.85_{.07}$   & $0.88_{.05}$   & $0.90_{.02}$   & $0.92_{.01}$   & $0.92_{.01}$   & $0.92_{.00}$   & $0.92_{.02}$   & $0.92_{.02}$   \\
\midrule
TabLLM (T0 + Text GPT-3)       & $0.51_{.04}$    & $0.72_{.05}$    & $0.82_{.03}$    & $0.85_{.05}$    & $0.88_{.03}$    & $0.91_{.02}$    & $0.89_{.02}$    & $0.91_{.01}$    & $0.91_{.01}$    & $0.93_{.01}$            \\
TabLLM (T0 + Text T0)          & $0.44_{.03}$    & $0.74_{.07}$    & $0.82_{.10}$    & $0.87_{.02}$    & $0.88_{.02}$    & $0.89_{.04}$    & $0.90_{.01}$    & $0.89_{.02}$    & $0.89_{.03}$    & $0.93_{.02}$            \\
TabLLM (T0 + Table-To-Text)    & $0.56_{.05}$    & $0.73_{.09}$    & $0.78_{.08}$    & $0.86_{.06}$    & $0.88_{.03}$    & $0.91_{.02}$    & $0.91_{.02}$    & $0.90_{.02}$    & $0.91_{.01}$    & $0.92_{.01}$            \\
TabLLM (T0 + Text Template)             & $0.54_{.04}$    & $0.76_{.14}$    & $0.83_{.05}$    & $0.87_{.04}$    & $0.87_{.06}$    & $0.91_{.01}$    & $0.90_{.01}$    & $0.92_{.01}$    & $0.92_{.01}$    & $0.94_{.01}$            \\
TabLLM (T0 + List Template)             & $0.52_{.03}$    & $0.73_{.12}$    & $0.83_{.05}$    & $0.87_{.04}$    & $0.88_{.04}$    & $0.91_{.02}$    & $0.91_{.01}$    & $0.92_{.01}$    & $0.92_{.01}$    & $0.94_{.01}$            \\ 
TabLLM (T0 + List Only Values) & $0.40_{.04}$   & $0.67_{.16}$   & $0.83_{.06}$   & $0.84_{.05}$   & $0.88_{.03}$   & $0.89_{.03}$   & $0.92_{.02}$   & $0.90_{.00}$   & $0.90_{.01}$   & $0.92_{.01}$            \\
TabLLM (T0 + List Perm. Names) & $0.57_{.02}$    & $0.78_{.07}$    & $0.85_{.02}$    & $0.82_{.06}$    & $0.87_{.05}$    & $0.90_{.02}$    & $0.92_{.02}$    & $0.91_{.01}$    & $0.91_{.01}$    & $0.93_{.02}$            \\
TabLLM (T0 + List Perm. Values)& $0.23_{.02}$    & $0.63_{.20}$    & $0.79_{.12}$    & $0.83_{.07}$    & $0.88_{.04}$    & $0.89_{.04}$    & $0.90_{.02}$    & $0.91_{.01}$    & $0.91_{.01}$    & $ 0.93_{.00}$            \\
TabLLM (T0 3B + Text Template)&$0.56_{.03}$    & $0.68_{.13}$    & $0.82_{.04}$    & $0.85_{.02}$    & $0.86_{.03}$    & $0.90_{.01}$    & $0.91_{.01}$    & $0.93_{.01}$    & $0.93_{.01}$    & $0.94_{.01}$            \\
\midrule
\midrule
\multicolumn{11}{l}{\textbf{Income Dataset}}\\
\midrule
Logistic regression            & ---             & $0.68_{.15}$   & $0.72_{.13}$   & $0.80_{.03}$   & $0.82_{.01}$   & $0.83_{.03}$   & $0.85_{.01}$   & $0.87_{.01}$   & $0.88_{.00}$   & $0.90_{.00}$  \\
Logistic regression (ordinal)  & ---             & $0.55_{.04}$   & $0.56_{.06}$   & $0.58_{.07}$   & $0.70_{.06}$   & $0.76_{.03}$   & $0.79_{.01}$   & $0.80_{.01}$   & $0.80_{.00}$   & $0.81_{.00}$  \\
LightGBM                       & ---             & $0.50_{.00}$   & $0.50_{.00}$   & $0.50_{.00}$   & $0.50_{.00}$   & $0.78_{.03}$   & $0.81_{.03}$   & $0.87_{.01}$   & $0.88_{.00}$   & $0.93_{.00}$  \\
LightGBM (ordinal)             & ---             & $0.50_{.00}$   & $0.50_{.00}$   & $0.50_{.00}$   & $0.50_{.00}$   & $0.78_{.01}$   & $0.81_{.01}$   & $0.86_{.01}$   & $0.89_{.00}$   & $0.93_{.00}$  \\
XGBoost                        & ---             & $0.50_{.00}$   & $0.59_{.06}$   & $0.77_{.02}$   & $0.79_{.03}$   & $0.82_{.02}$   & $0.84_{.01}$   & $0.87_{.01}$   & $0.88_{.00}$   & $0.93_{.00}$  \\
XGBoost (ordinal)              & ---             & $0.50_{.00}$   & $0.63_{.04}$   & $0.74_{.04}$   & $0.76_{.04}$   & $0.79_{.03}$   & $0.84_{.02}$   & $0.86_{.01}$   & $0.88_{.00}$   & $0.93_{.00}$  \\
\midrule
SAINT & --- & $0.74_{.03}$ & $0.65_{.15}$ & $0.79_{.03}$ & $0.81_{.03}$ & $0.84_{.02}$ & $0.84_{.02}$ & $0.87_{.01}$ & $0.88_{.00}$ & $0.91_{.00}$ \\
TabNet                         &  ---             & $0.56_{.04}$ & $0.59_{.07}$ & $0.62_{.11}$ & $0.64_{.06}$    & $0.71_{.04}$    & $0.73_{.05}$    & $0.80_{.02}$   & $0.83_{.02}$   &  $0.92_{.00}$ \\
NODE                           &  ---             & $0.54_{.02}$  & $0.54_{.04}$ & $0.65_{.04}$ & $0.67_{.03}$    & $0.75_{.02}$    & $0.78_{.01}$    & $0.78_{.01}$    & $0.83_{.01}$    &  $0.82_{.00}$  \\
TabPFN                         & ---             & $0.73_{.08}$   & $0.71_{.09}$   & $0.76_{.09}$   & $0.80_{.04}$   & $0.82_{.04}$   & $0.84_{.01}$   & $0.86_{.01}$   & $0.87_{.01}$   & $0.89_{.00}$  \\
TabPFN (ordinal)               & ---             & $0.64_{.11}$   & $0.64_{.06}$   & $0.72_{.04}$   & $0.77_{.02}$   & $0.80_{.02}$   & $0.81_{.01}$   & $0.83_{.01}$   & $0.85_{.01}$   & $0.87_{.00}$  \\
\midrule
TabLLM (T0 + Text GPT-3)       & $0.75_{.01}$    & $0.79_{.03}$    & $0.80_{.03}$    & $0.82_{.02}$    & $0.82_{.01}$    & $0.84_{.02}$    & $0.84_{.02}$    & $0.85_{.01}$    & $0.86_{.00}$    & *            \\
TabLLM (T0 + Text T0)          & $0.65_{.01}$    & $0.67_{.03}$    & $0.66_{.07}$    & $0.72_{.02}$    & $0.75_{.03}$    & $0.79_{.04}$    & $0.82_{.02}$    & $0.83_{.02}$    & $0.86_{.01}$    & *            \\
TabLLM (T0 + Table-To-Text)    & $0.50_{.00}$    & $0.64_{.07}$    & $0.64_{.11}$    & $0.72_{.05}$    & $0.74_{.03}$    & $0.79_{.03}$    & $0.81_{.01}$    & $0.84_{.01}$    & $0.84_{.01}$    & *            \\
TabLLM (T0 + Text Template)             & $0.84_{.00}$    & $0.84_{.01}$    & $0.84_{.02}$    & $0.84_{.04}$    & $0.84_{.01}$    & $0.84_{.02}$    & $0.86_{.01}$    & $0.87_{.00}$    & $0.89_{.01}$    & $0.92_{.00}$            \\
TabLLM (T0 + List Template)             & $0.79_{.01}$    & $0.83_{.01}$    & $0.83_{.03}$    & $0.83_{.02}$    & $0.84_{.01}$    & $0.85_{.01}$    & $0.86_{.01}$    & $0.87_{.01}$    & $0.88_{.01}$    & *            \\
TabLLM (T0 + List Only Values) & $0.73_{.01}$    & $0.74_{.04}$    & $0.75_{.04}$    & $0.80_{.03}$    & $0.82_{.01}$    & $0.84_{.01}$    & $0.84_{.01}$    & $0.86_{.01}$    & $0.87_{.01}$    & *            \\
TabLLM (T0 + List Perm. Names) & $0.65_{.00}$    & $0.75_{.03}$    & $0.74_{.05}$    & $0.82_{.02}$    & $0.83_{.02}$    & $0.84_{.02}$    & $0.86_{.01}$    & $0.86_{.01}$    & $0.88_{.01}$    & *            \\
TabLLM (T0 + List Perm. Values)& $0.26_{.00}$    & $0.40_{.04}$    & $0.48_{.10}$    & $0.65_{.06}$    & $0.72_{.03}$    & $0.79_{.03}$    & $0.81_{.02}$    & $0.83_{.01}$    & $0.84_{.01}$    & *            \\
TabLLM (T0 3B + Text Template)          & $0.76_{.00}$    & $0.77_{.06}$    & $0.80_{.04}$    & $0.83_{.02}$    & $0.83_{.03}$    & $0.85_{.01}$    & $0.86_{.00}$    & $0.86_{.01}$    & $0.88_{.01}$    & *            \\
\midrule
\midrule
\multicolumn{11}{l}{\textbf{Jungle Dataset}}\\
\midrule
Logistic regression            & ---             & $0.62_{.09}$   & $0.69_{.09}$   & $0.68_{.04}$   & $0.76_{.03}$   & $0.79_{.01}$   & $0.79_{.00}$   & $0.80_{.01}$   & $0.80_{.00}$ & $0.81_{.00}$  \\
Logistic regression (ordinal)  & ---             & $0.62_{.09}$   & $0.69_{.09}$   & $0.68_{.04}$   & $0.76_{.03}$   & $0.79_{.01}$   & $0.79_{.00}$   & $0.80_{.01}$   & $0.80_{.00}$   & $0.81_{.00}$   \\
LightGBM                       & ---             & $0.50_{.00}$   & $0.50_{.00}$   & $0.50_{.00}$   & $0.50_{.00}$   & $0.79_{.02}$   & $0.84_{.02}$   & $0.88_{.01}$   & $0.91_{.00}$   & $0.98_{.00}$  \\
LightGBM (ordinal)             & ---             & $0.50_{.00}$   & $0.50_{.00}$   & $0.50_{.00}$   & $0.50_{.00}$   & $0.79_{.02}$   & $0.84_{.02}$   & $0.88_{.01}$   & $0.91_{.00}$   & $0.98_{.00}$  \\
XGBoost                        & ---             & $0.50_{.00}$   & $0.58_{.07}$   & $0.72_{.05}$   & $0.78_{.03}$   & $0.81_{.02}$   & $0.84_{.02}$   & $0.87_{.01}$   & $0.91_{.01}$   & $0.98_{.00}$  \\
XGBoost (ordinal)              & ---             & $0.50_{.00}$   & $0.58_{.07}$   & $0.72_{.05}$   & $0.78_{.03}$   & $0.81_{.02}$   & $0.84_{.02}$   & $0.87_{.01}$   & $0.91_{.01}$   & $0.98_{.00}$  \\
\midrule
SAINT                          & ---             & $0.64_{.05}$   & $0.69_{.06}$   & $0.72_{.05}$   & $0.79_{.02}$   & $0.81_{.01}$   & $0.83_{.01}$   & $0.88_{.01}$   & $0.90_{.00}$   & $1.00_{.00}$  \\
TabNet                         & ---             & $0.53_{.09}$   & $0.60_{.05}$   & $0.62_{.03}$   & $0.69_{.04}$   & $0.73_{.04}$   & $0.75_{.02}$   & $0.79_{.02}$   & $0.84_{.01}$   & $0.99_{.00}$  \\
NODE                           & ---             & $0.60_{.01}$   & $0.71_{.03}$   & $0.68_{.04}$   & $0.74_{.02}$   & $0.75_{.04}$   & $0.78_{.01}$   & $0.79_{.01}$   & $0.80_{.00}$   & $0.81_{.00}$  \\
TabPFN                         & ---             & $0.65_{.08}$   & $0.72_{.04}$   & $0.71_{.07}$   & $0.78_{.02}$   & $0.81_{.01}$   & $0.84_{.01}$   & $0.88_{.01}$   & $0.91_{.00}$   & $0.93_{.00}$  \\
TabPFN (ordinal)               & ---             & $0.65_{.08}$   & $0.72_{.04}$   & $0.71_{.07}$   & $0.78_{.02}$   & $0.81_{.01}$   & $0.84_{.01}$   & $0.88_{.01}$   & $0.91_{.00}$   & $0.93_{.00}$   \\
\midrule
TabLLM (T0  + Text GPT-3)      & $0.56_{.01}$    & $0.58_{.02}$    & $0.55_{.02}$    & $0.60_{.06}$    & $0.68_{.03}$    & $0.74_{.03}$    & $0.77_{.01}$    & $0.81_{.01}$    & $0.85_{.01}$    & *  \\
TabLLM (T0 + Text T0)          & $0.63_{.00}$    & $0.63_{.04}$    & $0.64_{.05}$    & $0.62_{.06}$    & $0.70_{.01}$    & $0.71_{.03}$    & $0.74_{.02}$    & $0.78_{.02}$    & $0.82_{.01}$    & *  \\
TabLLM (T0 + Table-To-Text)    & $0.51_{.01}$    & $0.60_{.02}$    & $0.60_{.04}$    & $0.63_{.05}$    & $0.69_{.03}$    & $0.75_{.01}$    & $0.78_{.03}$    & $0.82_{.01}$    & $0.85_{.01}$    & *  \\
TabLLM (T0 + Text Template)    & $0.60_{.00}$    & $0.64_{.01}$    & $0.64_{.02}$    & $0.65_{.03}$    & $0.71_{.02}$    & $0.78_{.02} $    & $0.81_{.02}$    & $0.84_{.01}$    & $0.89_{.01}$    & $1.00_{\phantom{..}}$\dag   \\
TabLLM (T0 + List Template)    & $0.63_{.00}$    & $0.65_{.01}$    & $0.66_{.03}$    & $0.66_{.04}$    & $0.71_{.03}$    & $0.78_{.02}$    & $0.81_{.03}$    & $0.84_{.01}$    & $0.88_{.01}$    & *  \\
TabLLM (T0 + List Only Values) & $0.58_{.00}$    & $0.60_{.03}$    & $0.62_{.03}$    & $0.63_{.02}$    & $0.65_{.04}$    & $0.73_{.01}$    & $0.76_{.02}$    & $0.82_{.02}$    & $0.88_{.01}$    & *  \\
TabLLM (T0 + List Perm. Names) & $0.40_{.00}$    & $0.53_{.06}$    & $0.55_{.05}$    & $0.63_{.10}$    & $0.72_{.03}$    & $0.79_{.02}$    & $0.80_{.03}$    & $0.84_{.02}$    & $0.89_{.01}$    & *  \\
TabLLM (T0 + List Perm. Values)& $0.48_{.00}$    & $0.50_{.02}$    & $0.52_{.03}$    & $0.53_{.03}$    & $0.55_{.01}$    & $0.59_{.02}$    & $0.63_{.01}$    & $0.72_{.02}$    & $0.75_{.01}$    & *  \\
TabLLM (T0 3B + Text Template) & $0.54_{.00}$    & $0.63_{.02}$    & $0.64_{.04}$    & $0.67_{.03}$    & $0.72_{.03}$    & $0.77_{.02}$    & $0.80_{.02}$    & $0.83_{.01}$    & $0.87_{.01}$    & *  \\
\bottomrule
\end{tabular}
\begin{tablenotes}[flushleft]
    \footnotesize 
    \item[*] Result omitted due to runtime limitations of TabLLM on the full dataset.
    \item[\dag] These experiments were only performed for a single run due to runtime limitations of TabLLM on the full dataset.
\end{tablenotes}
\end{threeparttable}
}
\label{table:results_public_dataset_full3}
\end{table*}

\clearpage

\begin{table*}
\caption{Full results on healthcare claims dataset. The best concept selection method (most frequent concepts) and concept names (original concept names) were used as determined in prior zero-shot experiments. A fix number of 10 epochs was used for up to 256 shots and 3 epochs for more shots to decrease the runtime and prevent overfitting.}
\vspace{-0.4cm}
\centering
\setlength{\tabcolsep}{10pt}
\renewcommand{\arraystretch}{0.95}
\resizebox{\textwidth}{!}{\begin{tabular}{llccccccc}
\toprule
& \multicolumn{7}{c}{\textbf{Number of Shots}}\\
\addlinespace[1.5mm]
\textbf{Method} & \textbf{0} & \textbf{16} & \textbf{64} & \textbf{256} & \textbf{1,024} & \textbf{4,096} & \textbf{16,384} & \textbf{all}\\
\midrule
\midrule
\multicolumn{8}{l}{\textbf{End of Life (EoL)}}\\
\midrule
TabLLM (T0 + List Template)            & $0.70_{\phantom{.00}}$    & $0.74_{\phantom{.00}}$    & $0.78_{\phantom{.00}}$    & $0.78_{\phantom{.00}}$    & $0.79_{\phantom{.00}}$    & $0.81_{\phantom{.00}}$    & $0.81_{\phantom{.00}}$    & ---  \\
TabLLM (T0 + Text Template)    & $0.63_{\phantom{.00}}$    & $0.71_{\phantom{.00}}$    & $0.74_{\phantom{.00}}$    & $0.76_{\phantom{.00}}$    & $0.78_{\phantom{.00}}$    & $0.79_{\phantom{.00}}$    & $0.80_{\phantom{.00}}$    & ---  \\
TabLLM (T0 + List Short)      & $0.68_{\phantom{.00}}$    & $0.71_{\phantom{.00}}$    & $0.76_{\phantom{.00}}$    & $0.79_{\phantom{.00}}$    & $0.80_{\phantom{.00}}$    & $0.81_{\phantom{.00}}$    & $0.82_{\phantom{.00}}$    & ---  \\
TabLLM (T0 + List Perm. Names)& $0.62_{\phantom{.00}}$    & $0.66_{\phantom{.00}}$    & $0.70_{\phantom{.00}}$    & $0.74_{\phantom{.00}}$    & $0.75_{\phantom{.00}}$    & $0.77_{\phantom{.00}}$    & $0.79_{\phantom{.00}}$    & ---  \\
\midrule
Logistic Regression           & ---   & $0.65_{.07}$    & $0.77_{.02}$    & $0.80_{.02}$    & $0.83_{.01}$    & $0.83_{.01}$    &$0.84_{.01}$   & $0.84_{.01}$    \\
LightGBM                      & ---   & $0.50_{.00}$    & $0.71_{.01}$    & $0.76_{.02}$    & $0.80_{.01}$    & $0.82_{.01}$    & $0.83_{.01}$    & $0.82_{\phantom{..}}$*   \\
\midrule
TabLLM (T0 + List Template) unbalanced & $0.70_{\phantom{.00}}$    & $0.64_{\phantom{.00}}$    & $0.69_{\phantom{.00}}$    & $0.74_{\phantom{.00}}$    & $0.74_{\phantom{.00}}$    & $0.77_{\phantom{.00}}$    & $0.79_{\phantom{.00}}$    & ---  \\
Logistic Regression unbalanced& ---   & $0.44_{.04}$    & $0.53_{.12}$    & $0.75_{.03}$    & $0.77_{.03}$    & $0.80_{.02}$    &$0.82_{.02}$   & $0.84_{.01}$    \\
\midrule
\midrule
\multicolumn{8}{l}{\textbf{Surgical Procedure (Surgery)}}\\
\midrule
TabLLM (T0 + List Template)            & $0.67_{\phantom{.00}}$    & $0.73_{\phantom{.00}}$    & $0.72_{\phantom{.00}}$    & $0.73_{\phantom{.00}}$    & $0.75_{\phantom{.00}}$    & $0.78_{\phantom{.00}}$    & $0.79_{\phantom{.00}}$    & ---  \\
TabLLM (T0 + Text Template)    & $0.62_{\phantom{.00}}$    & $0.71_{\phantom{.00}}$    & $0.69_{\phantom{.00}}$    & $0.72_{\phantom{.00}}$    & $0.74_{\phantom{.00}}$    & $0.77_{\phantom{.00}}$    & $0.78_{\phantom{.00}}$    & ---  \\
TabLLM (T0 + List Short)      & $0.66_{\phantom{.00}}$    & $0.70_{\phantom{.00}}$    & $0.69_{\phantom{.00}}$    & $0.72_{\phantom{.00}}$    & $0.73_{\phantom{.00}}$    & $0.76_{\phantom{.00}}$    & $0.78_{\phantom{.00}}$    & ---  \\
TabLLM (T0 + List Perm. Names)& $0.60_{\phantom{.00}}$    & $0.68_{\phantom{.00}}$    & $0.70_{\phantom{.00}}$    & $0.72_{\phantom{.00}}$    & $0.74_{\phantom{.00}}$    & $0.77_{\phantom{.00}}$    & $_{\phantom{.00}}$    & ---  \\
\midrule
Logistic Regression           & ---   & $0.72_{.04}$    & $0.75_{.05}$    & $0.77_{.01}$    & $0.79_{.01}$    & $0.80_{.01}$    & $0.80_{.00}$    & $0.81_{.00}$   \\
LightGBM                      & ---   & $0.50_{.00}$    & $0.73_{.02}$    & $0.77_{.01}$    & $0.79_{.01}$    & $0.80_{.00}$    & $0.81_{.01}$    & $0.82_{\phantom{..}}$*   \\
\midrule
TabLLM (T0 + List Template) unbalanced & $0.67_{\phantom{.00}}$    & $0.68_{\phantom{.00}}$    & $0.73_{\phantom{.00}}$    & $0.74_{\phantom{.00}}$    & $0.75_{\phantom{.00}}$    & $0.77_{\phantom{.00}}$    & $0.79_{\phantom{.00}}$    & ---  \\
Logistic Regression unbalanced& ---   & $0.61_{.15}$    & $0.77_{.01}$    & $0.77_{.02}$    & $0.78_{.01}$    & $0.80_{.01}$    &$0.80_{.00}$   & $0.81_{.00}$    \\
\midrule
\midrule
\multicolumn{8}{l}{\textbf{Likelihood of Hospitalization (LoH)}}\\
\midrule
TabLLM (T0 + List Template)            & $0.71_{\phantom{.00}}$    & $0.73_{\phantom{.00}}$    & $0.73_{\phantom{.00}}$    & $0.76_{\phantom{.00}}$    & $0.78_{\phantom{.00}}$    & $0.81_{\phantom{.00}}$    & $0.82_{\phantom{.00}}$    & ---  \\
TabLLM (T0 + Text Template)    & $0.65_{\phantom{.00}}$    & $0.74_{\phantom{.00}}$    & $0.72_{\phantom{.00}}$    & $0.74_{\phantom{.00}}$    & $0.78_{\phantom{.00}}$    & $0.80_{\phantom{.00}}$    & $0.81_{\phantom{.00}}$    & ---  \\
TabLLM (T0 + List Short)      & $0.70_{\phantom{.00}}$    & $0.73_{\phantom{.00}}$    & $0.75_{\phantom{.00}}$    & $0.78_{\phantom{.00}}$    & $0.79_{\phantom{.00}}$    & $0.80_{\phantom{.00}}$    & $0.82_{\phantom{.00}}$    & ---  \\
TabLLM (T0 + List Perm. Names)& $0.62_{\phantom{.00}}$    & $0.71_{\phantom{.00}}$    & $0.72_{\phantom{.00}}$    & $0.75_{\phantom{.00}}$    & $0.75_{\phantom{.00}}$    & $0.78_{\phantom{.00}}$    & $0.80_{\phantom{.00}}$    & ---  \\
\midrule
Logistic Regression           & ---   & $0.72_{.04}$    & $0.76_{.03}$    & $0.80_{.01}$    & $0.82_{.01}$    & $0.83_{.01}$    & $0.83_{.01}$    & $0.84_{.01}$   \\
LightGBM                      & ---   & $0.50_{.00}$    & $0.72_{.02}$    & $0.76_{.03}$    & $0.81_{.01}$    & $0.83_{.00}$    & $0.83_{.01}$    & $0.85_{\phantom{..}}$*   \\
\midrule
TabLLM (T0 + List Template) unbalanced & $0.71_{\phantom{.00}}$    & $0.66_{\phantom{.00}}$    & $0.72_{\phantom{.00}}$    & $0.75_{\phantom{.00}}$    & $0.75_{\phantom{.00}}$    & $0.78_{\phantom{.00}}$    & $0.80_{\phantom{.00}}$    & ---  \\
Logistic Regression unbalanced& ---   & $0.53_{.06}$    & $0.54_{.09}$    & $0.73_{.06}$    & $0.79_{.01}$    & $0.81_{.01}$    &$0.82_{.01}$   & $0.84_{.01}$    \\
\bottomrule

\end{tabular}}
\begin{tablenotes}[flushleft]
    \footnotesize 
    \item[\dag] * These experiments were only performed for a single run due to runtime limitations on the full dataset.
\end{tablenotes}
\label{table:results_ibc_dataset_full}
\end{table*}

\begin{table*}[!t]
\caption{Feature importance of zero-shot TabLLM and LR on all data for the \textbf{Income} dataset. To determine the feature importance of TabLLM, we fit a separate LR model to the predictions using the original feature values as covariates. For LR we simply use the feature coefficients. The features are ranked by their TabLLM importance score.}
\vspace{-0.4cm}
\setlength{\tabcolsep}{4.8pt}
\footnotesize
\begin{tabular}{lrlrl}
\toprule
\textbf{Feature}    & \multicolumn{2}{c}{\textbf{TabLLM}} & \multicolumn{2}{c}{\textbf{LR}} \\
                    &  rank & weight &  rank & weight \\
\midrule
                             capital\_gain &         1 &     5.310 &         2 &     2.393 \\
                        education\_Masters &         2 &     4.623 &         6 &     1.455 \\
                      education\_Doctorate &         3 &     3.410 &         4 &     2.066 \\
                      education\_Bachelors &         4 &     2.995 &         7 &     1.135 \\
                    education\_Prof-school &         5 &     2.949 &         5 &     1.900 \\
              occupation\_Machine-op-insp. &         6 &     2.589 &        75 &    -0.325 \\
                        workclass\_Private &         7 &     2.275 &        37 &     0.102 \\
                        relationship\_Wife &         8 &     2.109 &         8 &     0.955 \\
                     native\_country\_China &         9 &     2.086 &        94 &    -0.839 \\
             native\_country\_United-States &        10 &     2.045 &        38 &     0.087 \\
                    native\_country\_Taiwan &        11 &     1.965 &        54 &     0.000 \\
                    workclass\_Federal-gov &        12 &     1.784 &        14 &     0.574 \\
                               race\_White &        13 &     1.685 &        61 &     0.000 \\
                     education\_Assoc-acdm &        14 &     1.621 &        13 &     0.574 \\
                       native\_country\_nan &        15 &     1.565 &        63 &    -0.056 \\
           marital\_status\_Married-civ-sp. &        16 &     1.487 &         3 &     2.214 \\
               occupation\_Protective-serv &        17 &     1.434 &        17 &     0.535 \\
                                 sex\_Male &        18 &     1.335 &        42 &     0.000 \\
                  occupation\_Armed-Forces &        19 &     1.290 &        60 &     0.000 \\
                  occupation\_Adm-clerical &        20 &     1.245 &        52 &     0.000 \\
                           hours\_per\_week &        21 &     1.240 &        20 &     0.424 \\
                      native\_country\_Hong &        22 &     1.227 &        86 &    -0.749 \\
                  occupation\_Tech-support &        23 &     1.164 &        18 &     0.526 \\
                     relationship\_Husband &        24 &     1.087 &        72 &    -0.212 \\
                         occupation\_Sales &        25 &     0.857 &        28 &     0.298 \\
                   native\_country\_Vietnam &        26 &     0.803 &        95 &    -0.898 \\
            marital\_status\_Married-AF-sp. &        27 &     0.792 &         1 &     2.571 \\
               native\_country\_Philippines &        28 &     0.711 &        40 &     0.011 \\
                                      age &        29 &     0.710 &        22 &     0.411 \\
                    native\_country\_Poland &        30 &     0.698 &        53 &     0.000 \\
                occupation\_Prof-specialty &        31 &     0.684 &        12 &     0.620 \\
                  race\_Asian-Pac-Islander &        32 &     0.651 &        32 &     0.254 \\
native\_country\_Outlying-US &        33 &     0.591 &        92 &    -0.836 \\
               workclass\_Self-emp-not-inc &        34 &     0.582 &        76 &    -0.344 \\
                     native\_country\_Italy &        35 &     0.534 &        24 &     0.400 \\
                 marital\_status\_Separated &        36 &     0.523 &        70 &    -0.181 \\
                            workclass\_nan &        37 &     0.515 &        59 &     0.000 \\
               occupation\_Exec-managerial &        38 &     0.503 &        10 &     0.773 \\
                  native\_country\_Scotland &        39 &     0.491 &        81 &    -0.626 \\
                      native\_country\_Laos &        40 &     0.475 &        44 &     0.000 \\
                  native\_country\_Cambodia &        41 &     0.328 &        11 &     0.642 \\
                 native\_country\_Guatemala &        42 &     0.276 &        55 &     0.000 \\
                      workclass\_State-gov &        43 &     0.267 &        73 &    -0.223 \\
                   native\_country\_Germany &        44 &     0.262 &        39 &     0.043 \\
               native\_country\_Puerto-Rico &        45 &     0.241 &        67 &    -0.128 \\
                   native\_country\_Hungary &        46 &     0.177 &        34 &     0.191 \\
                    native\_country\_Mexico &        47 &     0.123 &        80 &    -0.579 \\
                   native\_country\_Ireland &        48 &     0.116 &         9 &     0.954 \\
                        education\_HS-grad &        49 &     0.092 &        43 &     0.000 \\
              occupation\_Transport-moving &        50 &     0.090 &        62 &    -0.048 \\
               native\_country\_El-Salvador &        51 &     0.027 &        90 &    -0.803 \\
                    native\_country\_Canada &        52 &     0.027 &        23 &     0.407 \\
                   workclass\_Self-emp-inc &        53 &     0.001 &        30 &     0.255 \\
\bottomrule
\end{tabular}
\begin{tabular}{lrlrl}
\toprule
\textbf{Feature}    & \multicolumn{2}{c}{\textbf{TabLLM}} & \multicolumn{2}{c}{\textbf{LR}} \\
                    &  rank & weight &  rank & weight \\
\midrule
              relationship\_Other-relative &        54 &    -0.010 &        88 &    -0.759 \\
           native\_country\_Trinadad\&Tob. &        55 &    -0.028 &        66 &    -0.097 \\
                               race\_Black &        56 &    -0.044 &        74 &    -0.291 \\
                   native\_country\_England &        57 &    -0.088 &        16 &     0.551 \\
                  native\_country\_Honduras &        58 &    -0.105 &        58 &     0.000 \\
               relationship\_Not-in-family &        59 &    -0.153 &        29 &     0.257 \\
        native\_country\_Holand-Neth. &        60 &    -0.154 &        57 &     0.000 \\
                  occupation\_Craft-repair &        61 &    -0.161 &        36 &     0.108 \\
                             capital\_loss &        62 &    -0.182 &        31 &     0.255 \\
                               race\_Other &        63 &    -0.202 &        65 &    -0.085 \\
                native\_country\_Yugoslavia &        64 &    -0.204 &        27 &     0.357 \\
                      workclass\_Local-gov &        65 &    -0.230 &        47 &     0.000 \\
                           occupation\_nan &        66 &    -0.248 &        82 &    -0.653 \\
             marital\_status\_Never-married &        67 &    -0.292 &        77 &    -0.443 \\
                      native\_country\_Iran &        68 &    -0.330 &        41 &     0.000 \\
        native\_country\_Dominican-Rep. &        69 &    -0.332 &        85 &    -0.731 \\
          marital\_status\_Married-sp.-abs. &        70 &    -0.379 &        51 &     0.000 \\
                   native\_country\_Jamaica &        71 &    -0.416 &        25 &     0.392 \\
                 native\_country\_Nicaragua &        72 &    -0.425 &        45 &     0.000 \\
                  native\_country\_Thailand &        73 &    -0.451 &       100 &    -1.116 \\
                      native\_country\_Peru &        74 &    -0.522 &        93 &    -0.837 \\
                     native\_country\_Japan &        75 &    -0.617 &        56 &     0.000 \\
                   relationship\_Unmarried &        76 &    -0.620 &        48 &     0.000 \\
                    native\_country\_France &        77 &    -0.754 &        21 &     0.416 \\
                 occupation\_Other-service &        78 &    -0.754 &        96 &    -0.903 \\
                   workclass\_Never-worked &        79 &    -0.763 &        50 &     0.000 \\
                        education\_1st-4th &        80 &    -0.763 &       101 &    -1.172 \\
                  native\_country\_Columbia &        81 &    -0.836 &       104 &    -1.855 \\
                        education\_5th-6th &        82 &    -0.843 &        97 &    -0.961 \\
                  marital\_status\_Divorced &        83 &    -0.870 &        46 &     0.000 \\
                            education\_9th &        84 &    -0.904 &       102 &    -1.222 \\
                   native\_country\_Ecuador &        85 &    -0.952 &        49 &     0.000 \\
                           education\_11th &        86 &    -0.993 &        91 &    -0.825 \\
                     native\_country\_Haiti &        87 &    -1.062 &        35 &     0.137 \\
                      education\_Assoc-voc &        88 &    -1.074 &        19 &     0.514 \\
                     native\_country\_India &        89 &    -1.074 &        71 &    -0.183 \\
                        education\_7th-8th &        90 &    -1.151 &       103 &    -1.303 \\
                   marital\_status\_Widowed &        91 &    -1.253 &        64 &    -0.071 \\
                           education\_10th &        92 &    -1.306 &        89 &    -0.797 \\
                    native\_country\_Greece &        93 &    -1.319 &        68 &    -0.140 \\
                               sex\_Female &        94 &    -1.327 &        84 &    -0.710 \\
                     native\_country\_South &        95 &    -1.466 &        99 &    -1.101 \\
                      native\_country\_Cuba &        96 &    -1.575 &        33 &     0.230 \\
                   education\_Some-college &        97 &    -1.950 &        26 &     0.363 \\
             occupation\_Handlers-cleaners &        98 &    -1.992 &        83 &    -0.681 \\
                  native\_country\_Portugal &        99 &    -2.049 &        15 &     0.572 \\
                  race\_Amer-Indian-Eskimo &       100 &    -2.081 &        78 &    -0.465 \\
                   relationship\_Own-child &       101 &    -2.404 &        87 &    -0.755 \\
               occupation\_Priv-house-serv &       102 &    -2.840 &       105 &    -1.909 \\
                           education\_12th &       103 &    -3.178 &        79 &    -0.480 \\
                      education\_Preschool &       104 &    -3.520 &       106 &    -2.385 \\
               occupation\_Farming-fishing &       105 &    -3.853 &        98 &    -0.982 \\
                    workclass\_Without-pay &       106 &    -4.423 &        69 &    -0.174 \\
\bottomrule
\end{tabular}
\label{table:feature_importance_full}
\end{table*}

\clearpage

\begin{table*}[!t]
\caption{Feature importance of zero-shot TabLLM and relative risk (RR) with 95\% confidence interval (CI) for \textbf{EoL} task on the healthcare claims dataset. For TabLLM we fit a separate LR model to the predictions using the original feature values as covariates. We determine the relative risk treating the respective feature as an intervention, i.e. the ratio of the label in the group that has a concept divided by the ratio in the group without it. We selected 50 features with the highest and the lowest importance.}
\vspace{-0.4cm}
\setlength{\tabcolsep}{2.2pt}
\footnotesize
\begin{tabular}{lrll}
\toprule
\textbf{Feature}    & \multicolumn{2}{c}{\textbf{TabLLM}} & \multicolumn{1}{c}{\textbf{RR (95\% CI)}} \\
                    &  rank & weight &   \\
\midrule
               atrial fibrillation &         1 &     0.633 & 2.72 (2.51-2.95) \\
atherosclerosis of coronary art... &         2 &     0.530 & 2.10 (1.94-2.27) \\
          atherosclerosis of aorta &         3 &     0.473 & 1.99 (1.81-2.19) \\
exudative age-related macular d... &         4 &     0.452 & 2.38 (2.06-2.75) \\
                          sex\_male &         5 &     0.442 & 1.23 (1.14-1.33) \\
 non-hodgkin's lymphoma (clinical) &         6 &     0.440 & 1.36 (0.94-1.96) \\
       chronic atrial fibrillation &         7 &     0.436 & 3.36 (3.05-3.70) \\
    chronic kidney disease stage 3 &         8 &     0.430 & 2.75 (2.53-2.98) \\
atherosclerosis of arteries of ... &         9 &     0.404 & 2.76 (2.42-3.15) \\
               barrett's esophagus &        10 &     0.402 & 1.07 (0.84-1.37) \\
  chronic obstructive lung disease &        11 &     0.401 & 2.39 (2.19-2.60) \\
    paroxysmal atrial fibrillation &        12 &     0.395 & 2.58 (2.37-2.81) \\
      systemic lupus erythematosus &        13 &     0.395 & 1.51 (0.99-2.29) \\
atherosclerosis of artery of lo... &        14 &     0.394 & 2.45 (2.20-2.72) \\
          coronary atherosclerosis &        15 &     0.381 & 2.15 (1.95-2.36) \\
nonexudative age-related macula... &        16 &     0.377 & 2.15 (1.95-2.37) \\
  age related macular degeneration &        17 &     0.371 & 2.18 (1.76-2.71) \\
        pseudoexfoliation glaucoma &        18 &     0.360 & 1.13 (0.72-1.76) \\
degenerative joint disease invo... &        19 &     0.359 & 1.77 (1.52-2.06) \\
         coronary arteriosclerosis &        20 &     0.357 & 2.00 (1.82-2.20) \\
     coronary artery graft present &        21 &     0.346 & 1.64 (1.41-1.91) \\
aortocoronary bypass graft present &        22 &     0.335 & 2.24 (1.98-2.54) \\
                       dehydration &        23 &     0.332 & 2.94 (2.68-3.22) \\
primary malignant neoplasm of f... &        24 &     0.327 & 1.19 (1.01-1.40) \\
                malignant lymphoma &        25 &     0.322 & 1.54 (0.96-2.46) \\
cerebral infarction due to thro... &        26 &     0.316 & 2.86 (2.46-3.32) \\
          congestive heart failure &        27 &     0.313 & 3.67 (3.38-3.99) \\
         old myocardial infarction &        28 &     0.299 & 2.04 (1.81-2.30) \\
                       sleep apnea &        29 &     0.294 & 1.16 (0.98-1.37) \\
acute hypoxemic respiratory fai... &        30 &     0.292 & 4.02 (3.62-4.46) \\
  obstructive sleep apnea syndrome &        31 &     0.287 & 1.09 (0.96-1.24) \\
primary malignant neoplasm of e... &        32 &     0.284 & 0.92 (0.56-1.53) \\
        sensorineural hearing loss &        33 &     0.281 & 1.26 (1.09-1.47) \\
                retention of urine &        34 &     0.280 & 2.19 (1.97-2.44) \\
                    atrial flutter &        35 &     0.280 & 2.14 (1.85-2.47) \\
abdominal aortic aneurysm witho... &        36 &     0.275 & 1.85 (1.58-2.18) \\
chronic kidney disease due to h... &        37 &     0.274 & 2.65 (2.42-2.90) \\
    non-rheumatic aortic sclerosis &        38 &     0.271 & 2.64 (2.38-2.93) \\
          type 2 diabetes mellitus &        39 &     0.267 & 2.14 (1.96-2.33) \\
intraductal carcinoma in situ o... &        40 &     0.265 & 0.62 (0.30-1.29) \\
    chronic kidney disease stage 2 &        41 &     0.264 & 1.77 (1.55-2.03) \\
   degenerative disorder of macula &        42 &     0.263 & 2.23 (1.88-2.65) \\
sensorineural hearing loss, bil... &        43 &     0.262 & 1.30 (1.17-1.43) \\
                        race\_white &        44 &     0.262 & 1.25 (1.14-1.37) \\
          metabolic encephalopathy &        45 &     0.259 & 4.42 (3.86-5.07) \\
               alzheimer's disease &        46 &     0.256 & 5.03 (4.45-5.69) \\
               sick sinus syndrome &        47 &     0.256 & 2.37 (2.08-2.71) \\
           ventricular tachycardia &        48 &     0.255 & 2.33 (2.00-2.70) \\
      acute posthemorrhagic anemia &        49 &     0.255 & 2.15 (1.92-2.41) \\
         impaired fasting glycemia &        50 &     0.254 & 0.97 (0.85-1.09) \\
\bottomrule
\end{tabular}
\begin{tabular}{lrlrl}
\toprule
\textbf{Feature}    & \multicolumn{2}{c}{\textbf{TabLLM}} & \multicolumn{1}{c}{\textbf{RR (95\% CI)}} \\
                    &  rank & weight &   \\
\midrule
open wound of forehead without ... &    14056  &    -0.152 & 1.80 (1.18-2.74) \\
                       prediabetes &    14057  &    -0.157 & 0.81 (0.68-0.96) \\
             primary iridocyclitis &    14058  &    -0.157 & 1.63 (1.03-2.56) \\
             discoloration of skin &    14059  &    -0.157 & 0.87 (0.73-1.04) \\
basal cell carcinoma of truncal... &    14060  &    -0.158 & 1.14 (0.94-1.40) \\
                     lumbar sprain &    14061  &    -0.158 & 1.14 (0.91-1.42) \\
                             spasm &    14062  &    -0.160 & 0.98 (0.82-1.16) \\
                  chronic rhinitis &    14063  &    -0.161 & 1.22 (1.06-1.42) \\
            primary cardiomyopathy &    14064  &    -0.161 & 2.50 (2.11-2.97) \\
         benign neoplastic disease &    14065  &    -0.162 & 1.04 (0.63-1.72) \\
                      palpitations &    14066  &    -0.166 & 1.12 (1.01-1.25) \\
localized, primary osteoarthrit... &    14067  &    -0.167 & 1.50 (1.33-1.70) \\
benign neoplasm of skin of lowe... &    14068  &    -0.167 & 0.68 (0.53-0.89) \\
                     cyst of ovary &    14069  &    -0.171 & 0.90 (0.64-1.26) \\
             microscopic hematuria &    14070  &    -0.171 & 1.18 (1.01-1.37) \\
      problem related to lifestyle &    14071  &    -0.172 & 0.96 (0.48-1.91) \\
           acquired hypothyroidism &    14072  &    -0.172 & 1.47 (1.34-1.62) \\
abnormal findings on diagnostic... &    14073  &    -0.176 & 0.63 (0.54-0.73) \\
  increased frequency of urination &    14074  &    -0.177 & 1.41 (1.22-1.64) \\
                  disorder of skin &    14075  &    -0.178 & 1.18 (0.95-1.48) \\
                       thyroiditis &    14076  &    -0.180 & 0.87 (0.49-1.57) \\
        race\_hispanic\_or\_latino &    14077  &    -0.186 & 0.96 (0.60-1.51) \\
herpes zoster without complication &    14078  &    -0.187 & 1.14 (0.96-1.35) \\
         altered sensation of skin &    14079  &    -0.191 & 1.00 (0.82-1.22) \\
         generalized hyperhidrosis &    14080  &    -0.194 & 1.37 (1.07-1.76) \\
       primary open angle glaucoma &    14081  &    -0.194 & 1.35 (1.20-1.52) \\
                     stool finding &    14082  &    -0.195 & 1.48 (1.26-1.73) \\
                      primary gout &    14083  &    -0.196 & 1.80 (1.51-2.15) \\
localized, primary osteoarthrit... &    14084  &    -0.199 & 1.10 (0.92-1.30) \\
                          diarrhea &    14085  &    -0.200 & 1.73 (1.57-1.90) \\
benign neoplasm of skin of uppe... &    14086  &    -0.204 & 0.78 (0.58-1.03) \\
                       prostatitis &    14087  &    -0.204 & 1.20 (0.89-1.62) \\
                          eruption &    14088  &    -0.205 & 1.25 (1.11-1.41) \\
scar conditions and fibrosis of... &    14089  &    -0.206 & 1.00 (0.86-1.15) \\
             hashimoto thyroiditis &    14090  &    -0.215 & 0.91 (0.49-1.68) \\
         acquired deformity of toe &    14091  &    -0.227 & 1.25 (0.94-1.65) \\
                       race\_asian &    14092  &    -0.228 & 0.70 (0.50-0.99) \\
localized swelling, mass and lu... &    14093  &    -0.242 & 1.48 (1.15-1.91) \\
  benign neoplasm of skin of trunk &    14094  &    -0.245 & 0.91 (0.79-1.05) \\
     benign essential hypertension &    14095  &    -0.245 & 1.86 (1.72-2.01) \\
 finding of frequency of urination &    14096  &    -0.255 & 1.48 (1.34-1.64) \\
benign essential microscopic he... &    14097  &    -0.258 & 1.10 (0.76-1.59) \\
localized swelling, mass and lu... &    14098  &    -0.262 & 1.93 (1.67-2.23) \\
                 digestive symptom &    14099  &    -0.267 & 0.91 (0.68-1.21) \\
type 1 diabetes mellitus withou... &    14100  &    -0.298 & 2.34 (2.03-2.70) \\
open angle with borderline intr... &    14101  &    -0.338 & 1.20 (1.03-1.40) \\
primary localized osteoarthrosi... &    14102  &    -0.366 & 1.08 (0.82-1.43) \\
 localized, primary osteoarthritis &    14103  &    -0.393 & 1.23 (1.07-1.40) \\
                       sex\_female &    14104  &    -0.441 & 0.81 (0.75-0.88) \\
  open-angle glaucoma - borderline &    14105  &    -0.495 & 0.97 (0.85-1.10) \\
\bottomrule
\end{tabular}
\label{table:feature_importance_ibc_full}
\end{table*}

\clearpage

\twocolumn

\section{TASK TEMPLATES}
\label{sec:task_templates}

\newcommand{\exbox}[2]{
{\begin{samepage}
\noindent
#1
\vspace{-0.15cm}

\nopagebreak
\noindent
\fbox{
\begin{minipage}{0.45\textwidth}{
\begin{flushleft}
\footnotesize
\texttt{\noindent #2
}
\end{flushleft}}
\end{minipage}}
\end{samepage}
}}

\exbox{Bank Dataset:}
{answer\_choices: 'No ||| Yes'\newline
jinja: '\{\{serialization\}\}\newline\newline
Does this client subscribe to a term deposit? Yes or no?\newline
Answer: \newline
|||\newline
\{\{ answer\_choices[label] \}\}'}

\exbox{Blood Dataset:}
{answer\_choices: 'No ||| Yes'\newline
jinja: '\{\{serialization\}\}\newline\newline
Did the person donate blood? Yes or no?\newline
Answer: \newline
|||\newline
\{\{ answer\_choices[label] \}\}'}

\exbox{California Dataset:}
{answer\_choices: 'No ||| Yes'\newline
jinja: '\{\{serialization\}\}\newline\newline
Is this house block valuable? Yes or no?\newline
Answer: \newline
|||\newline
\{\{ answer\_choices[label] \}\}'}

\exbox{Car Dataset:}
{answer\_choices: 'Unacceptable ||| Acceptable ||| Good ||| Very good'\newline
jinja: '\{\{serialization\}\}\newline\newline
How would you rate the decision to buy this car? Unacceptable, acceptable, good or very good?\newline
Answer: \newline
|||\newline
\{\{ answer\_choices[label] \}\}'}

\exbox{Credit-g Dataset:}
{answer\_choices: 'No ||| Yes'\newline
jinja: '\{\{serialization\}\}\newline\newline
Does this person receive a credit? Yes or no?\newline
Answer: \newline
|||\newline
\{\{ answer\_choices[label] \}\}'}

\exbox{Diabetes Dataset:}
{answer\_choices: 'No ||| Yes'\newline
jinja: '\{\{serialization\}\}\newline\newline
Does this patient have diabetes? Yes or no?\newline
Answer: \newline
|||\newline
\{\{ answer\_choices[label] \}\}'}

\newpage

\exbox{Heart Dataset:}
{answer\_choices: 'No ||| Yes'\newline
jinja: '\{\{serialization\}\}\newline\newline
Does the coronary angiography of this patient show a heart disease? Yes or no?\newline
Answer: \newline
|||\newline
\{\{ answer\_choices[label] \}\}'}

\exbox{Income Dataset:}
{answer\_choices: 'No ||| Yes'\newline
jinja: '\{\{serialization\}\}\newline\newline
Does this person earn more than 50000 dollars per year? Yes or no?\newline
Answer: \newline
|||\newline
\{\{ answer\_choices[label] \}\}'}

\exbox{Jungle Dataset:}
{answer\_choices: 'No ||| Yes'\newline
jinja: '\{\{serialization\}\}\newline\newline
Does the white player win this two pieces endgame of Jungle Chess? Yes or no?\newline
Answer: \newline
|||\newline
\{\{ answer\_choices[label] \}\}'}

\exbox{End Of Life Task:}{answer\_choices: 'No ||| Yes'\newline
jinja: '\{\{serialization\}\}\newline\newline
Does this patient die in the next nine months? Yes or no?\newline
Answer: \newline
|||\newline
\{\{ answer\_choices[label] \}\}'}

\exbox{Surgical Procedure Task:}
{answer\_choices: 'No ||| Yes'\newline
jinja: '\{\{serialization\}\}\newline\newline
Does this patient need a surgery in the next nine months? Yes or no?\newline
Answer: \newline
|||\newline
\{\{ answer\_choices[label] \}\}'}

\exbox{Likelihood of Hospitalization Task:}
{answer\_choices: 'No ||| Yes'\newline
jinja: '\{\{serialization\}\}\newline\newline
Is this patient admitted to the hospital in the next nine months? Yes or no?\newline
Answer: \newline
|||\newline
\{\{ answer\_choices[label] \}\}'}

\section{EXAMPLE SERIALIZATIONS}
\label{sec:example_serializations}

\exbox{Bank Dataset (List Template):}
{- age: 69 \newline
- type of job: retired \newline
- marital status: single \newline
- education: tertiary \newline
- has credit in default?: no \newline
- average yearly balance, in euros: 2144 \newline
- has housing loan?: no \newline
- has personal loan?: no \newline
- contact communication type: cellular \newline
- last contact day of the month: 29 \newline
- last contact month of year: jul \newline
- last contact duration, in seconds: 417 \newline
- number of contacts performed during this campaign and for this client: \newline
- number of days that passed by after the client was last contacted from a previous campaign: 184 \newline
- number of contacts performed before this campaign and for this client: 4 \newline
- outcome of the previous marketing campaign: success}

\exbox{Bank Dataset (Text Template):}
{The age is 69. The type of job is retired. The marital status is single. The education is tertiary. The has credit in default? is no. The average yearly balance, in euros is 2144. The has housing loan? is no. The has personal loan? is no. The contact communication type is cellular. The last contact day of the month is 29. The last contact month of year is jul. The last contact duration, in seconds is 417. The number of contacts performed during this campaign and for this client is. The number of days that passed by after the client was last contacted from a previous campaign is 184. The number of contacts performed before this campaign and for this client is 4. The outcome of the previous marketing campaign is success.}

\newpage

\exbox{Bank Dataset (Table-To-Text):}
{the age of 69 was 69 years. the retired retired. the marital status is single with the single name. the school has a school of four students. the has a credit of \$500,000. The average yearly balance in euros is 2144. the has a total of 2,000+ housing units. the has an official loan of \$500 million. the standard definition has been updated to the standard definition. the current record of the month is 29. the first contact month was on December 20, 2005, and then on March 22, 2006, the next month was on March 22, 2006. the first contact duration was 417 seconds. the DVB has a selection of DVB. The year, in which the client was first contacted by a former airline operator, was by a former airline operator, and by a former airline operator, he was the first to enter the post of the office. the 4 is a 4-purpose cycle. the first of the first 20 MB of the history history to use the 20 MB.}

\exbox{Bank Dataset (Text T0):}
{a retired soldier shows off his tattoos. a city is a city with a population of singles and tertiary education. no, the average yearly balance is 2144 euros. no he has no personal loan or housing loan a man is contacting a woman on her cell phone on the 29th day of the month. last contact month of year was july, last contact duration was 417 seconds. 184 days after the client was last contacted from a previous campaign. a previous marketing campaign for this client resulted in success with 4 contacts}

\exbox{Bank Dataset (Text GPT-3):}
{The person is 69 years old, retired, single, and has a tertiary education. They have no credit in default, and their average yearly balance is 2144 euros. They have no housing loan or personal loan. The contact communication type is cellular, and the last contact was on the 29th day of the month and lasted 417 seconds. They have been contacted 4 times before this campaign, and the outcome of the previous marketing campaign was success.}

\exbox{Blood Dataset (List Template):}
{- Recency - months since last donation: 23 \newline
- Frequency - total number of donation: 1 \newline
- Monetary - total blood donated in c.c.: 250 \newline
- Time - months since first donation: 23}

\exbox{Blood Dataset (Text Template):}
{The Recency - months since last donation is 23. The Frequency - total number of donation is 1. The Monetary - total blood donated in c.c. is 250. The Time - months since first donation is 23.}

\exbox{Blood Dataset (Table-To-Text):}
{the number of the public can be from the number of the public. The 1.2 has a maximum speed of 1.2.  The first set of the first set was in 1742 and was in 1742.}

\exbox{Blood Dataset (Text T0):}
{The donor has made 1 donation in the last 23 months. monetary - total blood donated in c.c. : 250, time - months since first donation : 23}

\exbox{Blood Dataset (Text GPT-3):}
{The blood donor is a 23-year-old male who has donated blood once, 250 c.c. of blood, 23 months ago.}

\exbox{California Dataset (List Template):}
{- median income: 3.2377 \newline
- median age: 32 \newline
- total rooms: 6597 \newline
- total bedrooms: 1579 \newline
- population: 3689 \newline
- households: 1459 \newline
- latitude: 34.15 \newline
- longitude: -118.01}

\exbox{California Dataset (Text Template):}
{The median income is 3.2377. The median age is 32. The total rooms is 6597. The total bedrooms is 1579. The population is 3689. The households is 1459. The latitude is 34.15. The longitude is -118.01.}

\exbox{California Dataset (Table-To-Text):}
{there were 3.2377 people residing in the city.  the total rooms have 6597 rooms. the total has a total of 1579. The population was 3689 at the time of the census. The households 1459 is a standard households. The value 34.15 is a value that has a value of 34.15. The longitude has a distance of 1.5 km and is approximately 1.5 km.}

\exbox{California Dataset (Text T0):}
{median age of 32 years old the hotel has a total of 6597 rooms and 1579 bedrooms. a city has a population of 3689 and households of 1459. a city is located in the southwestern part of the country at latitude 34.15 and longitude -118.01.}

\newpage

\exbox{California Dataset (Text GPT-3):}
{The house block is located in the city of Los Angeles, in the state of California. The median income in the area is \$3,237, the median age is 32 years old, the total number of rooms is 6,597, the total number of bedrooms is 1,579, the population is 3,689, and the number of households is 1,459. The latitude is 34.15, and the longitude is -118.01.}

\exbox{Car Dataset (List Template):}
{- Buying price: low \newline
- Doors: three \newline
- Maintenance costs: low \newline
- Persons: more than four \newline
- Safety score: medium \newline
- Trunk size: medium}

\exbox{Car Dataset (Text Template):}
{The Buying price is low. The Doors is three. The Maintenance costs is low. The Persons is more than four. The Safety score is medium. The Trunk size is medium.}

\exbox{Car Dataset (Table-To-Text):}
{The price of the price is €1,000. the three Doors were three.  The total number of people in the city is more than four. the Safety score was 17.5. the Trunk size is 20.5-inch.}

\exbox{Car Dataset (Text T0):
}{The refrigerator has three doors and is very cheap. The maintenance costs are low for a family of more than four. The car has a medium safety score and a medium trunk size.}

\exbox{Car Dataset (Text GPT-3):}
{This car a good choice for those who are looking for a low-priced vehicle with low maintenance costs. It is also a good choice for families or groups of friends who need a car with a bit more space than a smaller car. The safety score is medium, so it is not the best choice for those who are looking for a car with the highest safety rating.}

\newpage

\exbox{Credit-g Dataset (List Template):}
{- Status of existing checking account: 0 <= ... < 200 DM \newline
- Duration in month: 11 \newline
- Credit history : existing credits paid back duly till now \newline
- Purpose: furniture/equipment \newline
- Credit amount: 1577 \newline
- Savings account/bonds: ... >= 1000 DM \newline
- Present employment since: <1 \newline
- Installment rate in percentage of disposable income: 4 \newline
- Personal status and sex: female : divorced/separated/married \newline
- Other debtors / guarantors: none \newline
- Present residence since: 1 \newline
- Property: real estate \newline
- Age in years: 20 \newline
- Other installment plans: none \newline
- Housing: own \newline
- Number of existing credits at this bank: 1 \newline
- Job: skilled employee / official \newline
- Number of people being liable to provide maintenance for: 1.0 \newline
- Telephone: none \newline
- foreign worker: yes}

\exbox{Credit-g Dataset (Text Template):}
{The Status of existing checking account is 0 <= ... < 200 DM. The Duration in month is 11. The Credit history is existing credits paid back duly till now. The Purpose is furniture/equipment. The Credit amount is 1577. The Savings account/bonds is ... >= 1000 DM. The Present employment since is <1. The Installment rate in percentage of disposable income is 4. The Personal status and sex is female : divorced/separated/married. The Other debtors / guarantors is none. The Present residence since is 1. The Property is real estate. The Age in years is 20. The Other installment plans is none. The Housing is own. The Number of existing credits at this bank is 1. The Job is skilled employee / official. The Number of people being liable to provide maintenance for is 1.0. The Telephone is none. The foreign worker is yes.}

\newpage

\exbox{Credit-g Dataset (Table-To-Text):}
{the 0.2 (0.2) is a type of 00.2. The average annual precipitation is 11.5 millimetres (4.5 in). the Credit history has been paid back to a few years. the standard\_cell is a standard\_cell. the amount was 1577. the Savings account/bonds were from the Savings account/bonds to the Savings account/bonds. there were 1,000 employees. there were 4,000 people in the city. The male has a male score of the female. the debt was \$12.5 million (\$9.5 million in 2013). the current residence has a 1,000 feet (460 m) long. the standard estate is a standard estate. It has a age of 20 years. the first installment was the first installment in the year 2005. The Housing is a public transport system that is a network of the public. the company has a number of existing and existing works, and has a number of existing and existing works. the company's job is job with the job name as "Success".  the network has a network of over 800 MT/s. the foreign worker has no foreign worker.}

\exbox{Credit-g Dataset (Text T0):}
{The checking account has a balance of 0 DM. A man is paying for furniture and equipment with a credit card. The credit amount is 1577, the savings account/bonds are >= 1000 DM. The present employee has been in this job for a year, and the installment rate is 4. \% of disposable income. A female who is divorced/separated/married is requesting a loan. The property is located in a gated community and has been on the market since. The man is 20 years old and has no other installment plans. The number of existing credits at this bank is 1. A skilled employee is liable to provide maintenance for 1.0. A foreign worker is without a telephone.}

\exbox{Credit-g Dataset (Text GPT-3):}
{The person is a 20-year-old female with a checking account status of 0-200 DM. She has been employed for less than a year and her installment rate is 4\% of her disposable income. She is divorced/separated/married and has no other debtors or guarantors. She has been living in her current residence for 1 year and owns real estate. She has 1 credit at this bank and is a skilled employee/official. She is liable for maintenance for 1 person. She has no telephone. She is a foreign worker.}

\newpage

\exbox{Diabetes Dataset (List Template):}
{- Age: 30 years \newline
- Number of times pregnant: 1 \newline
- Diastolic blood pressure: 64 mmHg \newline
- Triceps skin fold thickness: 32 mm \newline
- Plasma glucose concentration at 2 hours in an oral glucose tolerance test (GTT): 122 mg/dl \newline
- 2-hour serum insulin: 156 µU/ml \newline
- Body mass index: 35.1 \newline
- Diabetes pedigree function: 0.692}

\exbox{Diabetes Dataset (Text Template):}
{The Age is 30. The Number of times pregnant is 1. The Diastolic blood pressure is 64. The Triceps skin fold thickness is 32. The Plasma glucose concentration at 2 hours in an oral glucose tolerance test (GTT) is 122. The 2-hour serum insulin is 156. The Body mass index is 35.1. The Diabetes pedigree function is 0.692.}

\exbox{Diabetes Dataset (Table-To-Text):}
{The age was 30 years, and was the youngest ever to enter the age. the number of children is 1. The Diastolic blood pressure is 64. the Triceps can run up to 32. the 2 hours of the glucose is 122. the 2-hour cycle peaked to 156. The mass index was 35.1. The 0.692 is a fast and pathos.}

\exbox{Diabetes Dataset (Text T0):
}{The woman is 30 years old and has been pregnant once. The doctor checks the blood pressure and triceps skin fold thickness of the patient. The glucose concentration at 2 hours in an oral glucose tolerance test (GTT) was 122 and the 2-hour serum insulin was 156. The pedigree function of this family is 0.692.}

\exbox{Diabetes Dataset (Text GPT-3):}
{This patient 30 years old, has been pregnant once, has a diastolic blood pressure of 64 mmHg, and has a triceps skin fold thickness of 32 mm. The patient's plasma glucose concentration at 2 hours in an oral glucose tolerance test (GTT) is 122 mg/dl, and the patient's 2-hour serum insulin is 156 µU/ml. The patient's body mass index is 35.1, and the patient's diabetes pedigree function is 0.692.}

\newpage

\exbox{Heart Dataset (List Template):}
{- Age of the patient: 43 years \newline
- Sex of the patient: male \newline
- Chest pain type: asymptomatic \newline
- Resting blood pressure: 132 \newline
- Serum cholesterol: 247 \newline
- Fasting blood sugar > 120 mg/dl: yes \newline
- Resting electrocardiogram results: probable or definite left ventricular hypertrophy \newline
- Maximum heart rate achieved: 143 \newline
- Exercise-induced angina: yes \newline
- ST depression induced by exercise relative to rest: 0.1 \newline
- Slope of the peak exercise ST segment: flat}

\exbox{Heart Dataset (Text Template):}
{The Age of the patient is 43. The Sex of the patient is male. The Chest pain type is asymptomatic. The Resting blood pressure is 132. The Serum cholesterol is 247. The Fasting blood sugar > 120 mg/dl is yes. The Resting electrocardiogram results is probable or definite left ventricular hypertrophy. The Maximum heart rate achieved is 143. The Exercise-induced angina is yes. The ST depression induced by exercise relative to rest is 0.1. The Slope of the peak exercise ST segment is flat.}

\exbox{Heart Dataset (Table-To-Text):}
{The male patient was the 43rd of the Age of the patient. The male is a male of the same class.  The blood pressure was 132. The Serum cave has a cave of 247. the sugar has a low of 120 mg/dl. the type of the group is the type of the group that has a group of the group. The highest heart rate achieved is 143. the Exercise angina has a yes value. The ST depression has ranged from 0.1 to 0.1. the first segment was a flat of the ST.}

\exbox{Heart Dataset (Text T0):
}{The patient is a 43-year-old male. The chest pain is asymptomatic and resting blood pressure is 132. The doctor checks the fasting blood sugar and finds it is above 120 mg/dl. The resting ECG results showed probable or definite left ventricular hypertrophy, with maximum heart rate of 143 beats per minute. The patient had exercise-induced angina, with ST depression induced by exercise relative to rest of 0.1. The slope of the peak exercise segment is flat.}

\newpage

\exbox{Heart Dataset (Text GPT-3):}
{This patient a 43-year-old male with asymptomatic chest pain. His resting blood pressure is 132 mmHg and his serum cholesterol is 247 mm/dl. He has fasting blood sugar > 120 mg/dl and his resting electrocardiogram results are probable or definite left ventricular hypertrophy. His maximum heart rate achieved is 143 and he has exercise-induced angina. His ST depression induced by exercise relative to rest is 0.1 and his slope of the peak exercise ST segment is flat.}

\exbox{Income Dataset (List Template):}
{- Age: 30 \newline
- Race: Asian-Pac-Islander \newline
- Sex: Female \newline
- Marital status: never married \newline
- Rel. to head of the household: Own \newline
- Native country: Taiwan \newline
- Occupation: execution and management \newline
- Work class: private sector employee \newline
- Capital gain last year: 0 \newline
- Capital loss last year: 0 \newline
- Education: bachelor's degree \newline
- Work hours per week: 52}

\exbox{Income Dataset (Text Template):}
{The Age is 30. The Race is Asian-Pac-Islander. The Sex is Female. The Marital status is never married. The Relation to head of the household is Own-child. The Native country is Taiwan. The Occupation is execution and management. The Work class is private sector employee. The Capital gain last year is 0. The Capital loss last year is 0. The Education is bachelor's degree. The Work hours per week is 52.}

\exbox{Income Dataset (Table-To-Text):}
{The age was 30 years, and was the youngest ever to enter the age. The race was held in the Asian-Pac-Islander, and was won by the race. The sex of the village was Female. The first female to be married is Marital status never reported. the family has the head of the household. The Chinese:  native region of Taiwan. He was the executioners of the execution and management of the city of New York City. the private sector employee is a private sector employee. The capital was Capital of the State of India. The capital loss of the state was 0.5\%. The bachelor's degree in Education was bachelor's degree. the week 52 was the 52-hour week.}

\newpage
\exbox{Income Dataset (Text T0):
}{Kim is a 30-year-old Asian-Pacific Islander. She is never married and has never had children. The man is the owner of the house and he is the only child. A woman is executing a contract as a private sector employee. The company had a capital loss of \$ 0 last year. The man has a bachelor's degree and works 52 hours a week.}

\exbox{Income Dataset (Text GPT-3):}
{The person is 30 years old, Asian-Pac-Islander, female, never married, and an own-child relation to the head of the household. The person is from Taiwan and is an execution and management occupation in the private sector employee work class. The person has 0 dollars in capital gain and 0 dollars in capital loss from the previous year. The person has a bachelor's degree and works 52 hours per week.}

\exbox{Jungle Dataset (List Template):}
{- white piece strength: 6 \newline
- white piece file: 4 \newline
- white piece rank: 7 \newline
- black piece strength: 0 \newline
- black piece file: 5 \newline
- black piece rank: 2}

\exbox{Jungle Dataset (Text Template):}
{The white piece strength is 6. The white piece file is 4. The white piece rank is 7. The black piece strength is 0. The black piece file is 5. The black piece rank is 2.}

\exbox{Jungle Dataset (Table-To-Text):}
{the piece has a value of 6. the 4 file file has a 4-polytopic file. the piece has a cross point of the right side. the black piece strength is 0. The black piece file has a 5.0.}

\exbox{Jungle Dataset (Text T0):}
{The white piece has a strength of 6 and a file of 4. The white piece is ranked 7, the black piece is ranked 0. The black piece is ranked number two.}

\exbox{Jungle Dataset (Text GPT-3):}
{The white piece is stronger than the black piece. The white piece is on file 4 and rank 7. The black piece is on file 5 and rank 2.}

\newpage

\subsection{Large Healthcare Claims Dataset}
\label{subsec:example_serializations_ibc}

\exbox{End Of Life Task anonymized (List Template):}
{Summary: The patient is a 73 year old hispanic or latino man. \newline \newline
May 30, 2014: saw a doctor for dermatology \newline
Conditions: \newline
- chronic cholecystitis \newline
- aplastic anemia due to drugs \newline \newline
April 21, 2017: visited the hospital for 12 days \newline
Conditions: \newline
- chronic cholecystitis [...]}

\exbox{End Of Life Task anonymized (Text Template):}
{Summary: The patient is a 73 year old hispanic or latino man. \newline \newline
On May 30, 2014 the patient saw a doctor for dermatology with a primary complaint of chronic cholecystitis. He was also treated for aplastic anemia due to drugs. \newline \newline
On April 21, 2017 the patient visited the hospital for 12 days with a primary complaint of chronic cholecystitis. [...]}

\exbox{End Of Life Task anonymized (List Permuted Names):}
{Summary: The patient is a 73 year old hispanic or latino man. \newline \newline
May 30, 2014: saw a doctor for dermatology \newline
Conditions: \newline
- onychomycosis due to dermatophyte \newline
- chronic kidney disease \newline \newline
April 21, 2017: visited the hospital for 12 days \newline
Conditions: \newline
- onychomycosis due to dermatophyte [...]}

\clearpage

\onecolumn

\bibliographyonline{references.bib}

\end{document}